\documentclass{article}
\usepackage[preprint]{neurips_2026}
\PassOptionsToPackage{numbers,square,sort&compress}{natbib}
\usepackage[preprint]{neurips_2026}
\usepackage[utf8]{inputenc}
\usepackage[T1]{fontenc}
\usepackage{hyperref}
\usepackage{url}
\usepackage{booktabs}
\usepackage{amsfonts}
\usepackage{nicefrac}
\usepackage{microtype}
\usepackage{xcolor}
\usepackage{amsmath}
\usepackage{subcaption}

\newcommand{\vx}{\mathbf{x}}
\newcommand{\vth}{\boldsymbol{\theta}}
\newcommand{\vect}[1]{\boldsymbol{#1}}

\usepackage{graphicx}
\usepackage{amsmath}
\usepackage{amssymb}
\usepackage[ruled, noend, vlined]{algorithm2e}
\usepackage{xcolor}
\usepackage{enumitem}
\usepackage{xspace}
\usepackage{booktabs}
\usepackage{colortbl}
\usepackage{multirow}
\usepackage{makecell}
\usepackage{lipsum} % Just for generating dummy text, can be removed
\usepackage{gensymb} % for \degree
\usepackage{tikz}
\usepackage{wrapfig}

\definecolor{gen}{RGB}{181,80,65}

\definecolor{disc}{RGB}{62,91,134}

\definecolor{perceptual}{RGB}{46,139,87}
\definecolor{classif}{RGB}{212,160,23}
\definecolor{generalization}{RGB}{122,79,163}
\definecolor{clickme}{RGB}{140,90,60}

\newcommand{\dotgen}{\tikz[baseline=-0.6ex]\fill[gen] (0,0) circle (0.8ex);}

\newcommand{\dotdisc}{\tikz[baseline=-0.6ex]\fill[disc] (0,0) circle (0.8ex);}

\usepackage[labelsep=period]{caption}
\renewcommand{\paragraph}[1]{\vspace{0.2em}\noindent \textbf{#1 \hspace{0.2em}}}

\definecolor{MyDarkRed}{rgb}{0.46, 0.16, 0.16}
\definecolor{MyDarkBlue}{rgb}{0.16, 0.16, 0.66}

\title{%
Not Too Generative, Not Too Discriminative:\\ The Human Alignment Sweet Spot
}

\author{%
  Jorge Chang Ortega \\
  ANITI \\
  \And
  Bastien Le Lan \\
  ANITI \\
  \And
  Thomas Serre\thanks{These authors jointly supervised this work.}\\
  Brown University, ANITI \\
  \And
  Victor Boutin\footnotemark[1] \\
  CNRS, ANITI \\
}

\begin{document}

\maketitle

%%%%%%%%% ABSTRACT
\begin{abstract}

A central question in computational vision is whether human-like visual representations are better explained by discriminative or generative learning. 
Existing comparisons, however, often confound the learning objective with architecture, scale, and training data, leaving open whether the objective itself drives alignment.

We address this confound using Joint Energy-Based Models (JEMs), which interpolate continuously between discriminative and generative training within a fixed architecture. 
By varying a single mixing coefficient, we isolate the effect of the learning objective and evaluate the resulting models across six human-alignment benchmarks spanning perceptual similarity, gloss perception, human response uncertainty, robustness, shape–texture cue conflict, and diagnostic feature attribution. Across this diverse suite, human alignment is consistently maximized at intermediate points of the generative–discriminative continuum, rather than at either endpoint.
Hybrid JEMs combine the categorical structure induced by discriminative learning with the sensitivity to input structure induced by generative learning, yielding more human-like behavior across multiple levels of vision. 

These results suggest that the generative–discriminative dichotomy is the wrong axis for understanding human-aligned vision: alignment emerges not from choosing one objective over the other, but from balancing both.

\end{abstract}

\section{Introduction}
\label{sec:intro}

%P1 : Are human-like representations better explained by discriminative or generative objectives?

What computational principle underlies human-like visual representations---discriminative or generative learning?
Discriminative models cast vision as a feedforward mapping from inputs to labels~\cite{riesenhuber1999hierarchical, dicarlo2007untangling}. 
Generative models take the opposite view---vision as inference under an internal world model, shaped by top-down priors~\cite{rao1999predictive, lee2003hierarchical, yuille2006vision, peters2024does}.
In representation learning, this tension runs just as deep: discriminative training pushes representations toward category boundaries, while generative training pushes them toward the structure of the input distribution~\cite{ng2001discriminative}. 
Both traditions have claimed the crown at different levels of vision: discriminative training long dominated the modeling of high-level visual cortex~\cite{riesenhuber1999hierarchical, dicarlo2007untangling}, while generative training has been argued to better capture mid-level perception, gloss, and shape-based behavior~\cite{rao1999predictive, lee2003hierarchical, yuille2006vision, peters2024does}.

%Which computational principle best explains human-like visual representations: discriminative or generative learning?
%
%Discriminative accounts cast vision as a feedforward, bottom-up process that maps inputs to category labels~\cite{riesenhuber1999hierarchical, dicarlo2007untangling}, whereas generative accounts cast it as inference under an internal model of the world, in which top-down feedback supplies priors over the causes of sensory input~\cite{rao1999predictive, lee2003hierarchical, yuille2006vision, peters2024does}.
%
%The same divide carries over to representation learning: discriminative training optimizes for category separation, while generative training optimizes for modeling the distribution of sensory inputs~\cite{ng2001discriminative}.
%
%Each tradition has, in turn, been advanced as the right account of human visual representations: discriminative training on object categorization~\cite{yamins2014performance, schrimpf2018brain} long defined the state of the art for explaining high-level visual cortex, whereas generative training has been argued to better capture mid-level perception, including gloss and shape-based behavior~\cite{storrs2021unsupervised}, with recent work extending such claims to high-level recognition by reporting human-like shape biases in large diffusion models~\cite{jaini2024intriguing}.
%
%Yet, which principle better captures human-like representations remains unclear.

%
%P2 : This question is hard to answer because generative and discriminative models usually differ in more than just their objective (architecture).

The debate, however, rests on a shaky comparison.
Discriminative and generative models rarely differ only in their objective: they also differ in architecture, scale, and training data. 
Recent work makes this confound concrete: apparent advantages of generative models in shape bias and human alignment can be partly reproduced by simply low-pass filtering the inputs to a discriminative model~\cite{wolff2026lowpass}; training objective and dataset matter more than architecture for human similarity judgments~\cite {muttenthaler2022human}; and latent diffusion models align with humans no better than standard ImageNet classifiers on odd-one-out tasks~\cite{linhardt2024analysis}. What is still missing is a comparison in which the objective is varied in isolation. %The objective has rarely been varied in isolation. 
Until it is, the debate cannot be resolved.

%Comparing these two paradigms is challenging because generative and discriminative models rarely differ solely in their objective.
%
%Existing comparisons across object recognition~\cite{khaligh2014deep,cadieu2014deep}, gloss perception~\cite{storrs2021unsupervised}, perceptual similarity~\cite{zhang2018unreasonable}, and shape--texture bias~\cite{jaini2024intriguing,geirhos2018imagenet} also vary architecture, scale, and training data, leaving the role of the objective itself unclear.
%
%Recent evidence sharpens this concern: \citet{wolff2026lowpass} show that much of the apparent shape bias and human alignment of diffusion-based classifiers can be reproduced by simply low-pass filtering the inputs to a discriminative model, suggesting input resolution, not the generative objective, drives much of the reported gap; \citet{muttenthaler2022human} find that, for human similarity judgments, training objective and dataset matter substantially more than architecture or scale; and \citet{linhardt2024analysis} report that latent diffusion models align with humans no better than ImageNet-trained discriminative models on triplet odd-one-out tasks.
%
%To disentangle the objective from these confounds, it must be varied within a fixed architecture and training regime.

% P3 : Energy-based models offer a way around this problem.
\begin{wrapfigure}{r}{0.5\textwidth}
%\vspace{-4mm}
    \centering
    \includegraphics[width=\linewidth]{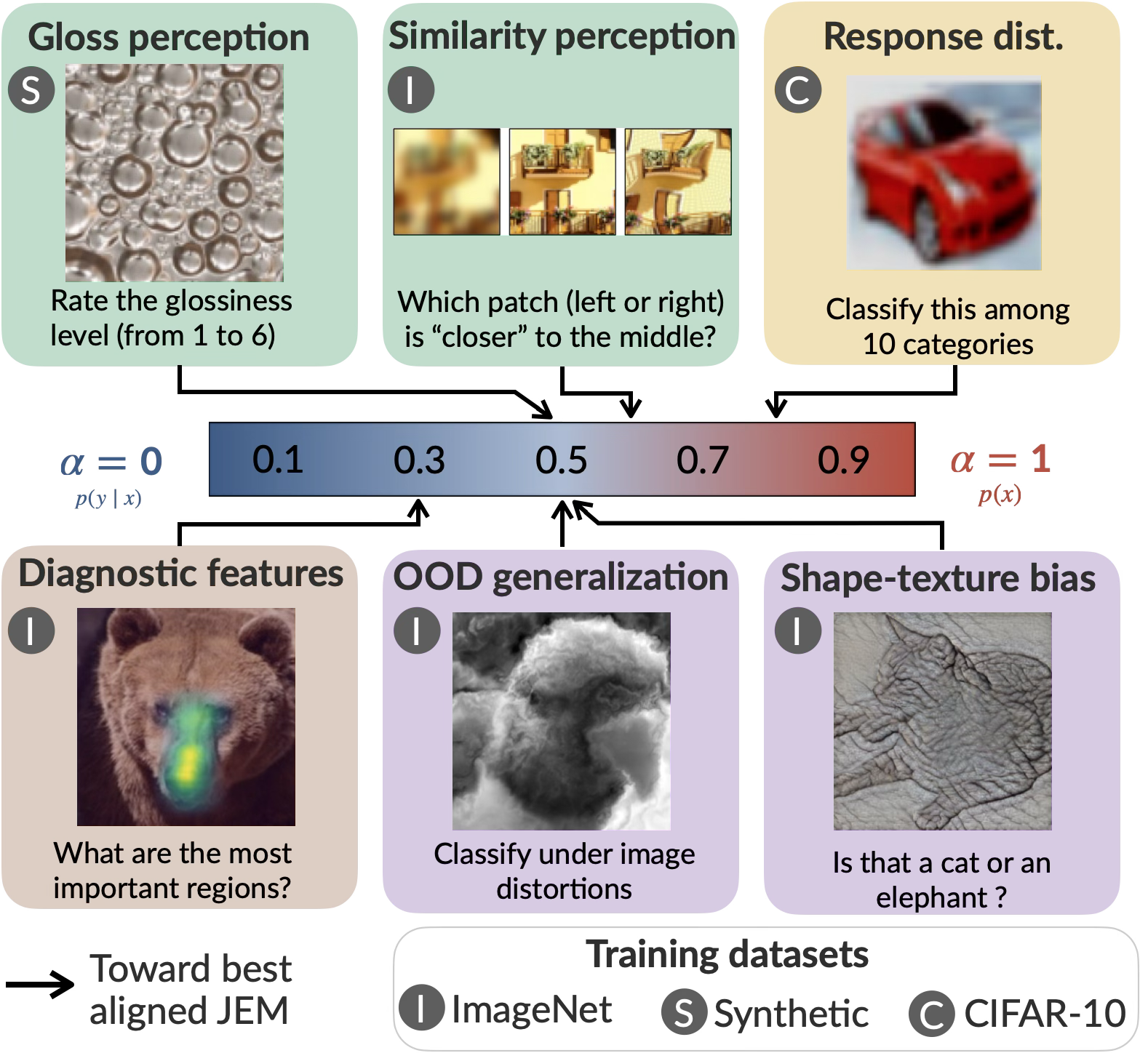}
    \vspace{-4mm}
\caption{\textbf{Human alignment peaks in the hybrid regime.} JEMs are trained across the \textcolor{gen}{generative ($p(x)$)}--\textcolor{disc}{discriminative ($p(y|x)$)} continuum by varying $\alpha\in[0,1]$ and evaluated on six human--machine comparison benchmarks. Arrows indicate the best-aligned $\alpha$ for each benchmark.}
    \label{fig:intro_fig}
    \vspace{-4mm}
\end{wrapfigure}

Joint Energy-Based Models (JEMs)~\cite{grathwohl2020your}
%Energy-based models (EBMs)~\cite{lecun2006tutorial, du2019implicit} 
offer a principled way to resolve this debate.
A JEM assigns an energy to each input–label pair, with lower energy meaning higher compatibility.
The same network simultaneously supports discriminative prediction---the conditional \textcolor{disc}{$p(y|x)$}--- and generative density modeling---the marginal \textcolor{gen}{$p(x)$}.
A single mixing coefficient $\alpha\in[0, 1]$ controls the balance between the two objectives, while architecture, scale, and data stay fixed. 
This gives us a clean axis along which to ask: where does human alignment live?

%
%The same scoring function admits two readings: discriminatively, it selects the lowest-energy label for a given input; generatively, it assigns low energy to inputs that are likely under the data distribution.
%
%Joint Energy-based Models (JEMs)~\cite{grathwohl2020your} make this duality practical: a single classifier-like architecture supports both discriminative and generative training, with the relative weight of the two objectives controlled by a single mixing coefficient.
%
%JEMs therefore let us vary the learning objective in isolation, holding architecture, scale, and data fixed.

We train JEMs across this continuum and evaluate them on a battery of human–machine benchmarks.  
The benchmarks cover low-, mid-, and high-level vision: perceptual similarity, gloss perception, human uncertainty on ambiguous images, generalization under distribution shift, shape–texture cue conflict, and diagnostic feature attribution.
Across all benchmarks, the answer is consistent: maximum human alignment is never achieved at the extreme values $\alpha$ = 0.0 or $\alpha$ = 1.0 (see Fig.~\ref{fig:intro_fig}). 
Hybrid JEMs match or exceed discriminative baselines on perceptual similarity, surpass all baselines on gloss judgments, better reproduce human uncertainty and shape bias, and retain strong generalization under distribution shift.
We further show that the energy landscape itself can be exploited at test time — refining inputs toward higher-probability images amplifies shape-consistent responses without any retraining.
Taken together, these results reframe the long-standing debate. The question shifts from ``which objective is best?'' to ``how should discriminative and generative objectives be balanced?''
%Taken together, these results reframe the long-standing debate. Rather than asking which objective best captures human vision, the right question is how the two are balanced. 
%
Human-aligned representations emerge not from either extreme, but from the regime in between.

% P4 : Using this framework, we show that human alignment is not maximized at either extreme, but at an intermediate regime.
%Building on this framework, we train JEMs across the discriminative--generative continuum and evaluate them on a battery of human--machine comparisons that span low-, mid-, and high-level vision: perceptual similarity, gloss perception, human uncertainty on ambiguous images, generalization under distribution shift, shape--texture cue conflict, and diagnostic feature attribution.
%
%Across the entire battery, we find a consistent intermediate optimum: human-aligned representations emerge neither from purely discriminative nor from purely generative training, but from a balance between the two.
%
%We further show that the energy landscape learned by hybrid JEMs can be exploited at test time, by iteratively refining inputs toward higher-probability images, to amplify shape-consistent responses without retraining---revealing that the same mechanism that shapes human-aligned representations during training can also be recruited at inference.
%
%Taken together, our results recast the long-standing discriminative--generative debate: rather than asking which paradigm best models human vision, the field should ask how the two are balanced. The architectures, evaluations, and analyses that follow argue that this balance---not either extreme---is the regime in which human-aligned vision emerges.

%-------------------------------------------------------------------------
\section{Related Work}
\label{sec:related_works}

\paragraph{Human alignment of generative and discriminative models.} 

Deep networks have become influential models of human and biological vision, from hierarchical recognition models inspired by the ventral visual stream~\cite{riesenhuber1999hierarchical,kriegeskorte2015deep} to supervised networks that predict primate neural responses and aspects of human behavior~\cite{khaligh2014deep,cadieu2014deep,yamins2014performance,yamins2016using,schrimpf2018brain}.
However, high classification accuracy does not guarantee human-like perception: standard discriminative models can be fooled by imperceptible or unrecognizable patterns~\cite{szegedy2014intriguing,nguyen2015deep}, rely on local or texture-based cues~\cite{baker2018deep,geirhos2018imagenet}, and diverge from humans under cue conflict and distribution shift~\cite{geirhos2020shortcut,geirhos2021partial,bowers2023deep}. 
This has motivated benchmarks that compare models to humans beyond accuracy, including perceptual similarity~\cite{zhang2018unreasonable}, human response distributions~\cite{peterson2019human,battleday2020capturing,collins2022behavioral}, shape--texture bias~\cite{geirhos2018imagenet,hermann2020origins}, robustness to distribution shift~\cite{hendrycks2019benchmarking,geirhos2021partial}, gloss perception~\cite{storrs2021unsupervised}, and perceptual illusions~\cite{jaini2024intriguing}. 
Together, this work suggests a mixed picture. 
Discriminative models capture important aspects of recognition and neural predictability, whereas recent generative models have been reported to better capture some human-like perceptual biases, including gloss judgments, shape bias, human-like errors, and illusion sensitivity~\cite{storrs2021unsupervised,jaini2024intriguing}. 
Yet these comparisons typically contrast models that differ in objective, architecture, training data, scale, and inductive bias. 
As a result, it remains unclear whether human alignment stems from the learning objective itself or from correlated model properties.

%A growing body of work has compared generative and discriminative models on human-relevant perceptual and behavioral tasks.
%Supervised discriminative deep networks capture important aspects of primate IT representations~\cite{cadieu2014deep,khaligh2014deep,yamins2014performance,yamins2016using,schrimpf2018brain}, whereas generative models have been reported to show advantages on several human-alignment benchmarks, including gloss perception, shape--texture bias, human classification errors, perceptual illusions, and one-shot drawing behavior~\cite{storrs2021unsupervised,jaini2024intriguing,boutin2022diversity}.
%Recently, there has been some work stating that generative classifiers can outperform standard discriminative models on several human-alignment benchmarks~\cite{jaini2024intriguing}.
%However, these comparisons are typically confounded by simultaneous differences in architecture, training pipeline, and inductive bias in models.

\paragraph{Hybrid generative--discriminative models.}

Hybrid generative--discriminative learning aims to combine density modeling with category prediction~\cite{lasserre2006principled,kingma2014semi,maaloe2016auxiliary}. 
However, many hybrid approaches introduce auxiliary inference networks, modify the architecture, or combine separately parameterized generative and discriminative components, making it difficult to isolate the effect of the learning objective. 
Energy-based models provide a cleaner testbed by defining a single energy function over input--label pairs~\cite{lecun2006tutorial,du2019implicit}. 
In particular, Joint Energy-based Models (JEMs) reinterpret a standard classifier as a joint energy model, enabling both conditional classification and density modeling within the same architecture~\cite{grathwohl2020your}. 
We exploit this property to treat the generative--discriminative balance as an experimental variable, interpolating between the two objectives while holding architecture, scale, and data fixed. 
Additional related work is discussed in Appendix.~\ref{supp:extended_related_works}.

%To address this limitation, a related line of work explores hybrid generative--discriminative models~\cite{ng2001discriminative,lasserre2006principled,druck2007semi,kuleshov2017hybrid,gordon2020combining,grathwohl2020your}.
%For our purposes, Joint Energy-based Models (JEMs) represent an attractive alternative because they preserve a standard classifier-like architecture while enabling a controlled interpolation between discriminative and generative training within a single scoring function~\cite{grathwohl2020your,yang2021jempp,yang2022towards}.
%This allows us to ask a cleaner question than prior human--model comparisons: where on the generative--discriminative spectrum do the most human-aligned representations emerge?
%

\section{Method}

\subsection{Joint Energy-Based Models (JEMs)}
\label{sub:jem}

Let $p_{\mathcal D}(\vx,y)$ denote the unknown data distribution over input images $\vx \in \mathbb{R}^N$ and class labels $y \in \{1,\dots,K\}$. 

Joint Energy-Based Models (JEMs)~\cite{grathwohl2020your} define a parametric model $p_{\vth}(\vx,y)$ over the joint distribution of inputs and labels, where $\vth$ denotes the model parameters. 
They can therefore be viewed as generative classifiers: rather than modeling only $p_{\vth}(y\mid\vx)$, they model the joint distribution $p_{\vth}(\vx,y)$. Because the model is joint, its log-probability decomposes as
\begin{equation}
\label{eq:decompo}
\log p_{\vth}(\vx,y) = \textcolor{disc}{\log p_{\vth}(y \mid \vx)} + \textcolor{gen}{\log p_{\vth}(\vx)}
\end{equation}
Eq.~\ref{eq:decompo} reveals two complementary components of the joint log-likelihood: \textcolor{disc}{a discriminative term, $\log p_{\vth}(y \mid \vx)$}, that drives correct classification, and a \textcolor{gen}{generative term, $\log p_{\vth}(\vx)$}, that captures the structure of the input distribution. 

A key property of JEMs is that both terms are implemented using the same classifier-like network. Let $\vect{f}_{\vth}(\vx) \in \mathbb{R}^K$ denote the logits produced by a neural network with parameters $\vth$ for an input $\vx$, and let $\vect{f}_{\vth}(\vx)[y]$ denote the logit associated with class $y$. JEMs reinterpret these logits by defining an energy function $E_{\vth}(\vx, y)$ over input--label pairs
\vspace{-1.5mm}\begin{equation}
\label{eq:joint_energy}
p_{\vth}(\vx, y) = \frac{\exp\!\left(-E_{\vth}(\vx, y)\right)}{Z(\vth)}\qquad\text{with} \qquad E_{\vth}(\vx, y) = -\vect{f}_{\vth}(\vx)[y] 
\end{equation}
where 
$Z(\vth)=\sum_{y'} \int \exp(-E_{\vth}(\vx',y'))\,d\vx'$ 
is the partition function. Because it requires summing over labels and integrating over all possible inputs $\vx$ in a high-dimensional space, $Z(\vth)$ is computationally intractable and cannot be evaluated exactly. Intuitively, lower energy indicates greater compatibility between input $\vx$ and label $y$. From this joint distribution, we recover both the conditional distribution over labels and the marginal distribution over inputs:

\vspace{-1mm}\begin{minipage}{.5\linewidth}
\begin{equation} \label{eq:conditional_jem}
 \textcolor{disc}{p_{\vth}(y \mid \vx)} = \frac{\exp\!\left(\vect{f}_{\vth}(\vx)[y]\right)} {\sum_{y'} \exp\!\left(\vect{f}_{\vth}(\vx)[y']\right)}
\end{equation}
\end{minipage}%
\begin{minipage}{.5\linewidth}
\begin{equation}
\label{eq:marginal_jem}
\displaystyle \textcolor{gen}{p_{\vth}(\vx)} = \frac{\sum_y \exp\!\left(\vect{f}_{\vth}(\vx)[y]\right)}{Z(\vth)}
\end{equation}
\end{minipage}

Eq.~\ref{eq:conditional_jem} is exactly the usual softmax classifier, while Eq.~\ref{eq:marginal_jem} defines a density over inputs using the same logits. 
This is the key advantage of JEMs for our purpose: the same network, and therefore the same learned representation, supports both discriminative prediction and generative modeling. %
This lets us probe a shared representation shaped by both objectives, rather than comparing separately parameterized generative and discriminative models.

We train $p_{\vth}(\vx,y)$ by maximizing the expected joint log-likelihood under the data distribution:
\vspace{-1mm}\begin{equation}
\label{eq:loss_1}
\mathbb{E}_{(\vx,y)\sim p_{\mathcal D}}
\left[\log p_{\vth}(\vx,y)\right]
=
\textcolor{disc}{\mathcal{L}_{D}} + \textcolor{gen}{\mathcal{L}_{G}},
\qquad
\left\{
\begin{aligned}
\textcolor{disc}{\mathcal{L}_{D}}
&=
\textcolor{disc}{\mathbb{E}_{(\vx,y)\sim p_{\mathcal D}}
\left[\log p_{\vth}(y \mid \vx)\right]}
\\
\textcolor{gen}{\mathcal{L}_{G}}
&=
\textcolor{gen}{\mathbb{E}_{\vx\sim p_{\mathcal D}}
\left[\log p_{\vth}(\vx)\right]}
\end{aligned}
\right.
%\vspace{-3mm}
\end{equation}

\vspace{-3mm}In Eq.~\ref{eq:loss_1}, $\mathcal{L}_{D}$ is the discriminative log-likelihood, corresponding to the negative of the standard cross-entropy loss, while $\mathcal{L}_{G}$ is the generative log-likelihood over inputs. In practice, $\mathcal{L}_{G}$ is approximated via contrastive divergence~\cite{hinton2002training,woodford2006notes,du2019implicit} (see Appendix~\ref{supp:CD}), which requires drawing samples from the marginal $p_{\vth}(\vx)$. Because the partition function $Z(\vth)$ is intractable, direct sampling is infeasible. However, Eq.~\ref{eq:marginal_jem} reveals that $p_{\vth}(\vx) \propto \sum_y \exp\!\left(\vect{f}_{\vth}(\vx)[y]\right)$, so the corresponding unnormalized marginal energy takes the simple form of a log-sum-exp over the logits:
\vspace{-1mm}\begin{equation}
\label{eq:marginal_energy}
    E_{\vth}(\vx) = -\log \sum_y \exp\!\left(\vect{f}_{\vth}(\vx)[y]\right)
%\vspace{-3mm}
\end{equation}
This quantity is cheap to compute, differentiable w.r.t.\ both $\vx$ and $\vth$, and requires no architectural change: the same logits that define the softmax classifier also define the marginal energy. We draw approximate samples from $p_{\vth}(\vx)$ by running Stochastic Gradient Langevin Dynamics (SGLD)~\cite{welling2011bayesian} on $E_{\vth}(\vx)$. Starting from an initial sample $\vx_0$, SGLD iterates:
\begin{equation}
\label{eq:sgld}
    \vx_{t+1} = \vx_{t} - \frac{\eta}{2}\,\nabla_{\!\vx}\,E_{\vth}(\vx_{t}) + \omega,
    \qquad \omega \sim \mathcal{N}(0,\,\eta\,\mathbf{I})
\end{equation}
where $\eta$ is the step size and $\omega$ is Gaussian noise, which turns gradient descent on the energy into an approximate sampling procedure rather than pure mode-seeking optimization. After $L$ steps, the output $\vx_L$ serves as the negative sample for the contrastive-divergence update of $\mathcal{L}_{G}$~\cite{du2019implicit}. Beyond training, SGLD also provides a
natural refinement mechanism: since the energy landscape encodes the model's learned image prior, iterating SGLD from an arbitrary input tends to move it toward lower-energy, higher-probability regions under the model. % 
We exploit this property at test time in Section~\ref{sec:shape-texture}, where we use it to amplify shape-consistent responses without any retraining. 

Finally, the joint decomposition in Eq.~\ref{eq:loss_1} provides the basis for our main experimental manipulation: we use the mixing coefficient $\alpha$ to interpolate between discriminative and generative training.

\subsection{Interpolating the Generative–Discriminative Trade-off}

To control the balance between discriminative and generative learning, we maximize the following interpolated objective:
\begin{equation}
\label{eq:real_loss}
\mathcal{L} = \alpha \cdot \frac{\|\nabla_{\vth}\mathcal{L}_{D}\|}{\|\nabla_{\vth}\mathcal{L}_{G}\|}\cdot\textcolor{gen}{\mathcal{L}_{G}} + (1-\alpha)\cdot\textcolor{disc}{\mathcal{L}_{D}}
\end{equation}
Here, $\alpha \in [0,1]$ controls the balance between the two terms: $\alpha=0$ recovers purely \textcolor{disc}{discriminative} training, while $\alpha=1$ recovers purely \textcolor{gen}{generative} training. The generative term is rescaled by the ratio of gradient norms, treated as a stop-gradient scalar, so that both objectives contribute at a comparable scale during training. Without this correction, the raw magnitudes of $\|\nabla_{\vth}\mathcal{L}_{D}\|$ and $\|\nabla_{\vth}\mathcal{L}_{G}\|$ can differ by orders of magnitude---particularly at intermediate $\alpha$---causing $\alpha$ to behave non-linearly as an interpolation parameter and destabilizing hybrid training. Throughout this paper, JEMs trained with different values of $\alpha$ are represented by a continuous color gradient, ranging from \dotdisc\ for fully \textcolor{disc}{discriminative} JEMs to \dotgen\ for fully \textcolor{gen}{generative} JEMs. We use a Wide ResNet architecture~\cite{zagoruyko2016wide}, following the original JEM implementation~\cite{grathwohl2020your}. Training 11 JEMs across the full continuum is computationally intensive, particularly at ImageNet scale where we operate in the latent space of a pretrained autoencoder to keep hybrid training feasible. Full training details, compute requirements, and hyperparameter choices are provided in Appendix~\ref{supp:ebm_training_global}, and reproducible code is available at \url{https://anonymous.4open.science/r/JEM-0B78/}.

\subsection{Human--Machine Comparison Tasks}

To evaluate how the generative--discriminative trade-off affects human alignment, we use a diverse set of benchmarks spanning four complementary aspects of human visual behavior: low- and mid-level perceptual judgments, response distributions under uncertainty, generalization and cue use, and diagnostic feature attribution. 
This breadth is important because human alignment is a multidimensional property: models may match human perception, uncertainty, and diagnostic features to different degrees, and we ask whether the same generative--discriminative balance improves alignment across all of them. We selected benchmarks with established, publicly available protocols, well-defined human reference points, and deep-learning baselines against which JEMs can be quantitatively compared.

\paragraph{\textcolor{perceptual}{Low- and mid-level perceptual judgments.}}

For low-level perception, we use the perceptual similarity benchmark of~\citet{zhang2018unreasonable}, based on the BAPPS dataset of distorted image patches. It includes two tasks. In the two-alternative forced choice (2AFC) task, observers choose which of two distorted patches is more similar to a reference. In the just noticeable difference (JND) task, they judge whether a distorted patch is perceptually distinguishable from the reference. 
Alignment is measured by 2AFC Test ($\%$) and JND Test (mAP), respectively, with higher values indicating better alignment. Human consistency on the 2AFC task provides an empirical ceiling of 83\%~\cite{zhang2018unreasonable}; the metric maximum of 100\% would imply perfect agreement, which is not observed in practice.

For mid-level perception, we use the gloss perception benchmark of~\citet{storrs2021unsupervised}, based on controlled synthetic images in which illumination and surface geometry are systematically varied as nuisance factors, so that surface reflectance is the only reliable cue for a correct gloss judgment. Alignment is measured by the Pearson correlation between human and model responses. Higher values indicate better agreement, with $r = 1$ as the theoretical upper bound of the metric rather than an empirically measured human ceiling. See Appendix~\ref{supp:bench_perceptual} and Appendix~\ref{supp:bench_gloss_depth} for more details on the low- and mid-level perceptual benchmarks, respectively.

\paragraph{\textcolor{classif}{Human classification response distributions.}}

Unlike benchmarks that reduce alignment to a single accuracy score, CIFAR-10H~\cite{peterson2019human} captures graded uncertainty across observers, allowing us to test whether models reproduce the full distribution of human responses rather than just the modal one. Alignment is measured by cross-entropy between model and human response distributions, where lower values indicate better alignment. The cross-entropy between independent groups of human observers ($0.55$ nats) serves as an empirical ceiling, reflecting irreducible inter-observer disagreement. See Appendix~\ref{supp:cifar10h} for more details.

\paragraph{\textcolor{generalization}{Generalization and representational bias.}}

We next evaluate whether models generalize and rely on visual cues in a human-like way. 
For generalization, we use the ImageNet-based benchmark of~\citet{geirhos2018generalisation}, which compares human and model classification accuracy under a range of out-of-distribution image transformations. We report error consistency with human responses (Cohen's $\kappa$, chance-corrected), with higher values indicating better alignment and a human inter-rater ceiling of approximately $\kappa = 0.39$~\cite{geirhos2021partial}. 

For cue use, we use the shape--texture conflict benchmark of~\citet{geirhos2018imagenet}, in which each image combines the global shape of one object category with the local texture of another. A model's classification reveals which cue dominates its decision. 
We report shape bias, defined as the proportion of cue-conflict images classified according to shape rather than texture, with values closer to the human reference of approximately $96\%$ indicating more human-like cue use~\cite{geirhos2018imagenet}. Together, these benchmarks test whether the generative--discriminative trade-off affects the model's learned invariances and visual features. See Appendix~\ref{supp:bench_generalization} and Appendix~\ref{supp:bench_shape_texture} for details.

\paragraph{\textcolor{clickme}{Human diagnostic features.}}

Finally, we use the ClickMe dataset~\cite{linsley2018learning} to test whether models rely on the same diagnostic features as humans during ImageNet object recognition. Participants clicked on image regions they found informative for recognizing the object, yielding spatial importance maps that reflect human visual strategies. We compare model attribution maps against these human maps using the mean Spearman correlation normalized by human inter-rater agreement~\cite{fel2022harmonizing}, with higher values indicating better alignment and a normalized score of 1 corresponding to the human ceiling. This benchmark is uniquely diagnostic: rather than measuring what the model predicts, it measures which visual evidence it uses, directly probing the spatial structure of learned representations. See Appendix~\ref{supp:click_me} for more details.

The JEM family evaluated on each benchmark matches its underlying visual domain. Specifically, we use ImageNet-trained JEMs for perceptual similarity (BAPPS), human-like generalization, shape--texture cue-conflict, and ClickMe; CIFAR-10-trained JEMs for CIFAR-10H; and synthetic-data-trained JEMs for gloss perception. This ensures that comparisons probe the effect of the generative--discriminative trade-off within an appropriate training domain, rather than across mismatched datasets. See Appendix~\ref{supp:ebm_training} for full training details.

\section{Results}
\label{sec:Results}

\begin{figure}[ht!]
\begin{tikzpicture}
\draw [anchor=north west] (0.0\linewidth, 0.98\linewidth) node {\includegraphics[width=0.48\linewidth]{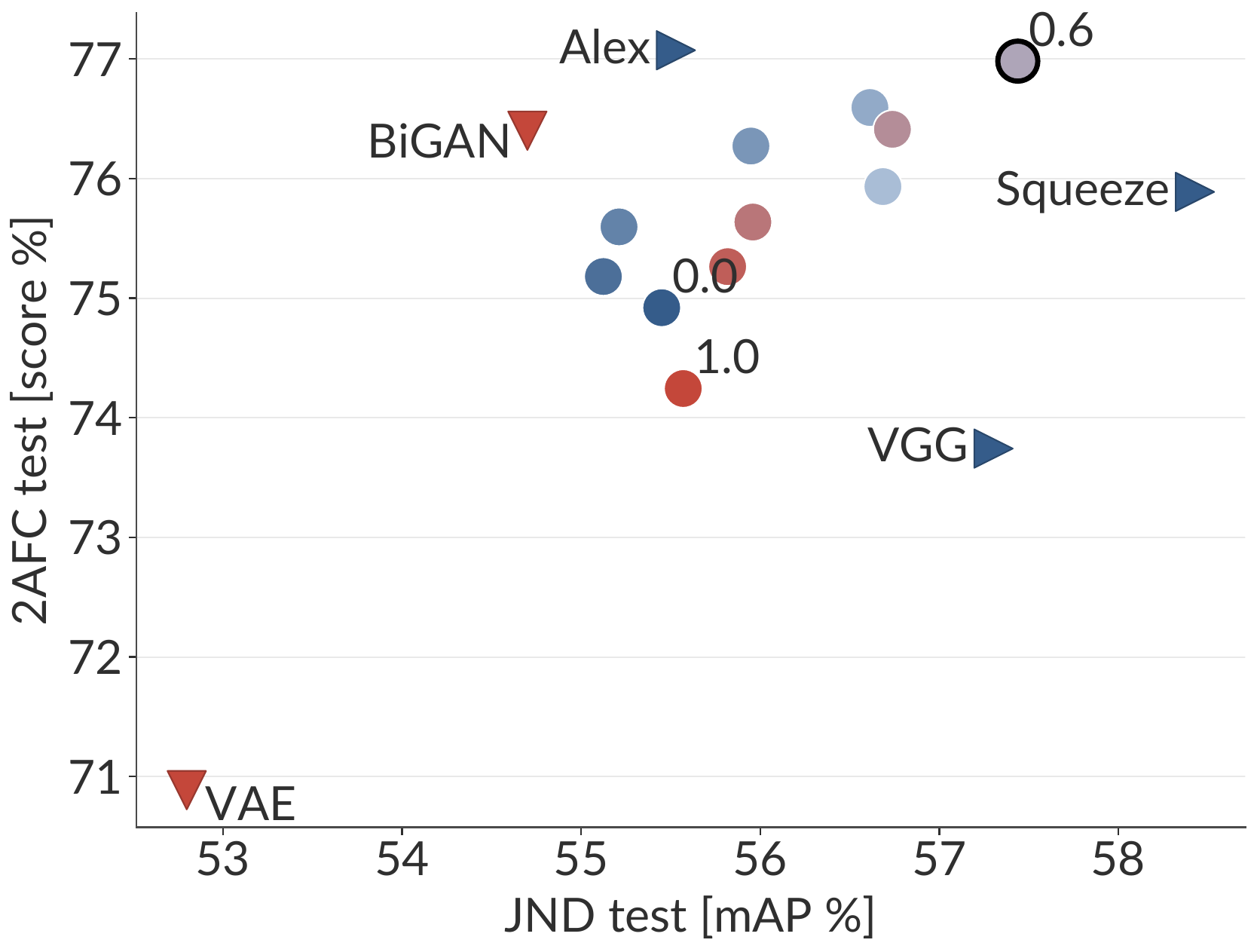}};
\draw [anchor=north west] (0.5\linewidth, 0.98\linewidth) node {\includegraphics[width=0.48\linewidth]{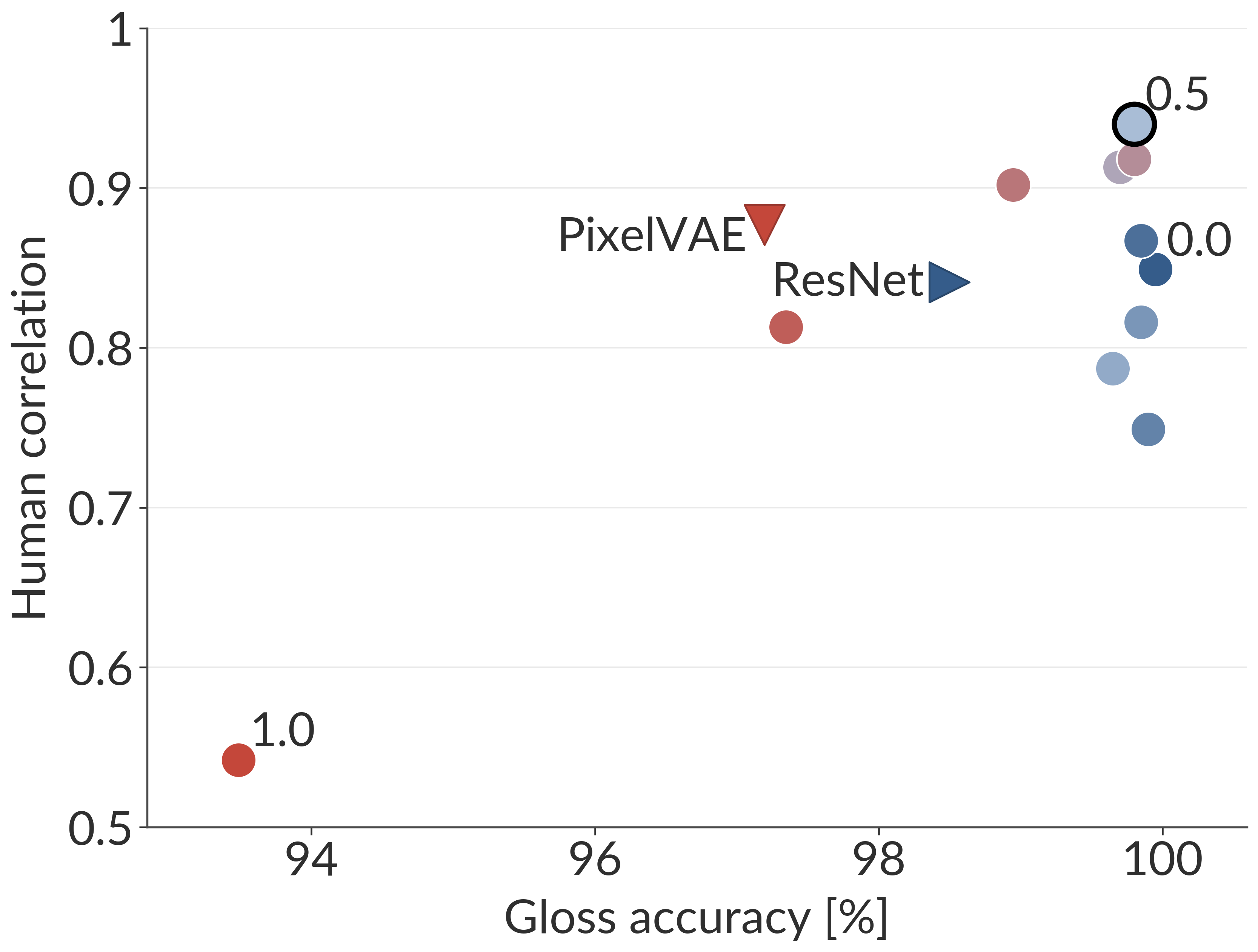}};
\draw [anchor=north west] (0.0\linewidth, 0.58\linewidth) node {\includegraphics[width=0.48\linewidth]{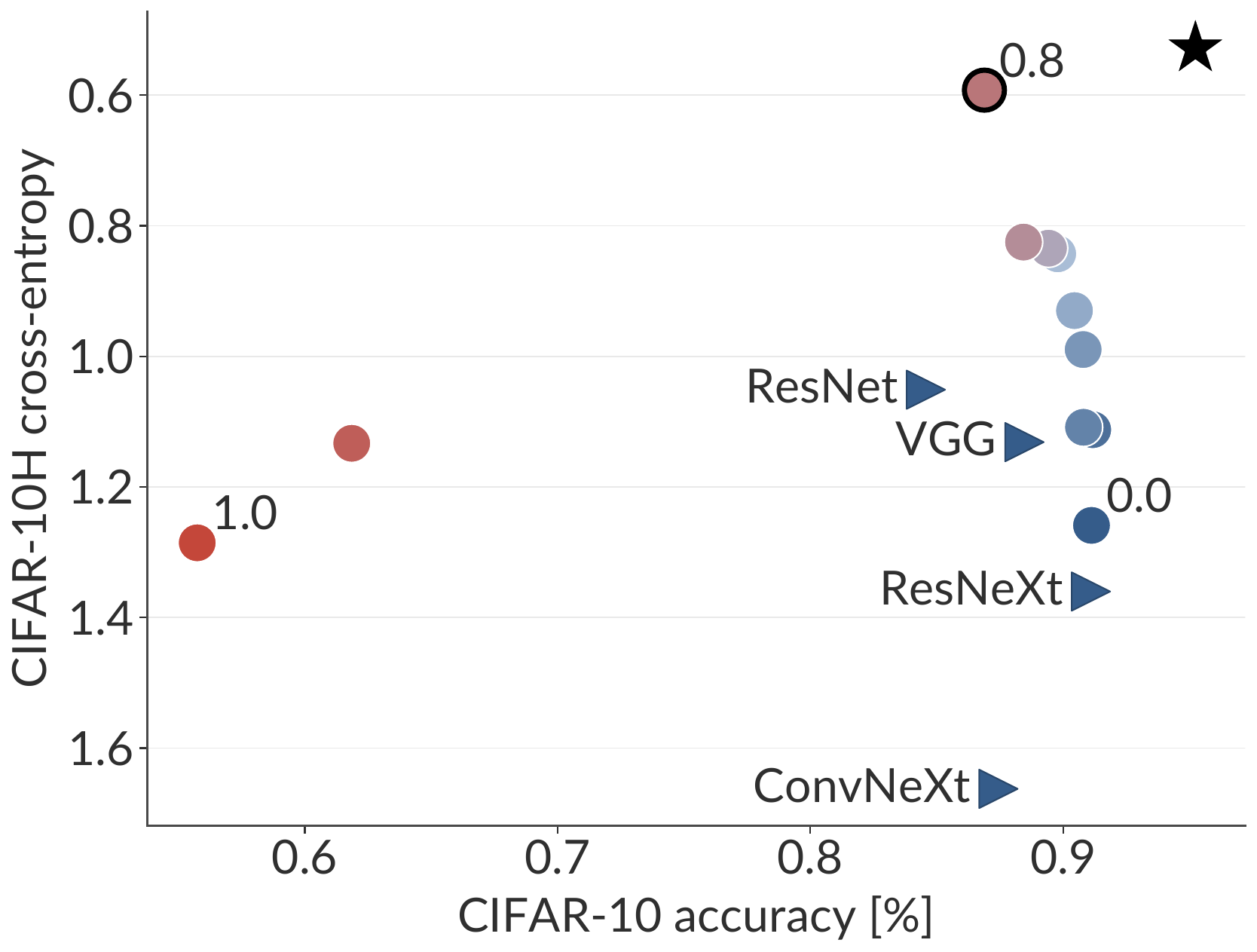}};
\draw [anchor=north west] (0.5\linewidth, 0.58\linewidth) node {\includegraphics[width=0.48\linewidth]{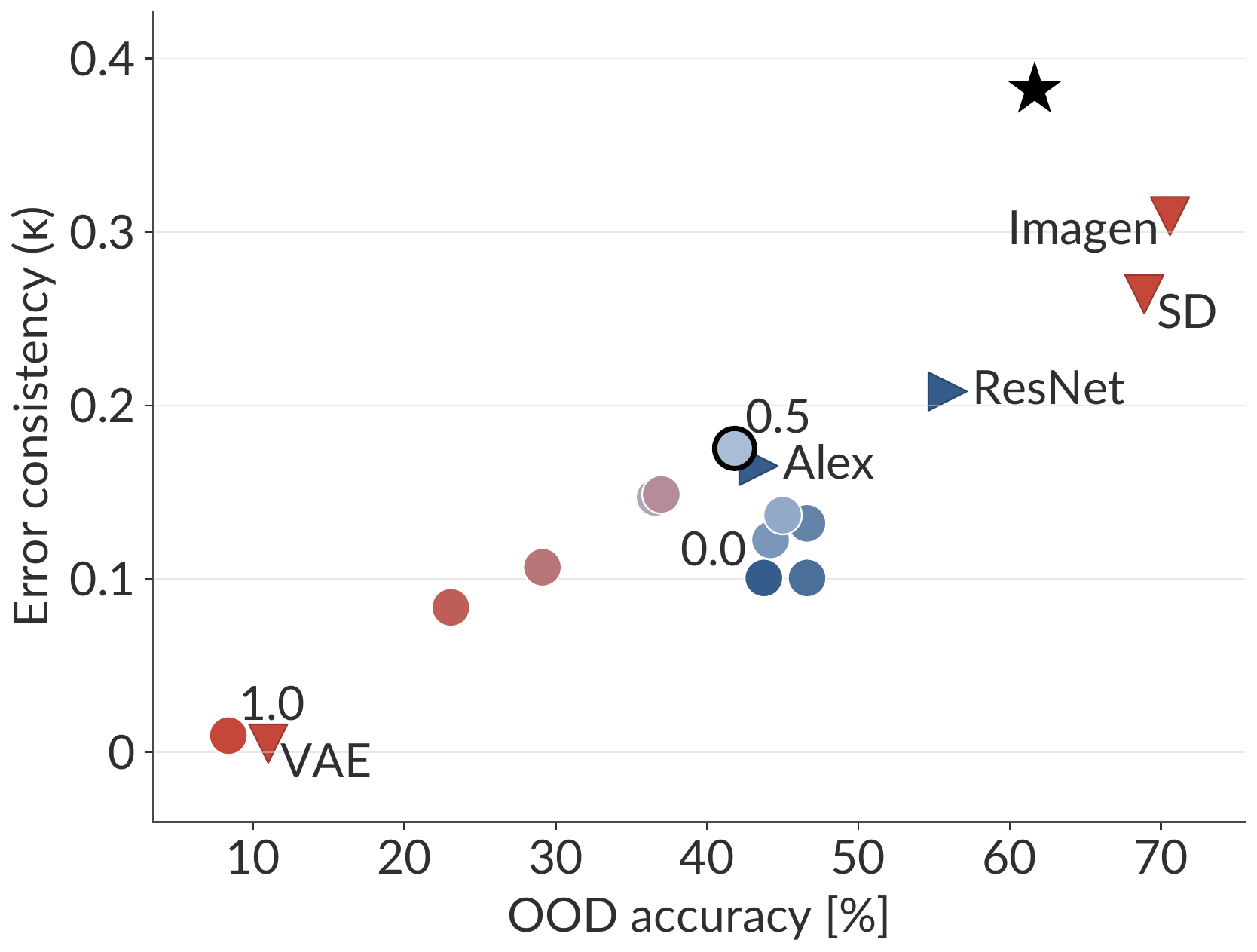}};
\draw [anchor=north west] (0.0\linewidth, 0.18\linewidth) node {\includegraphics[width=0.48\linewidth]{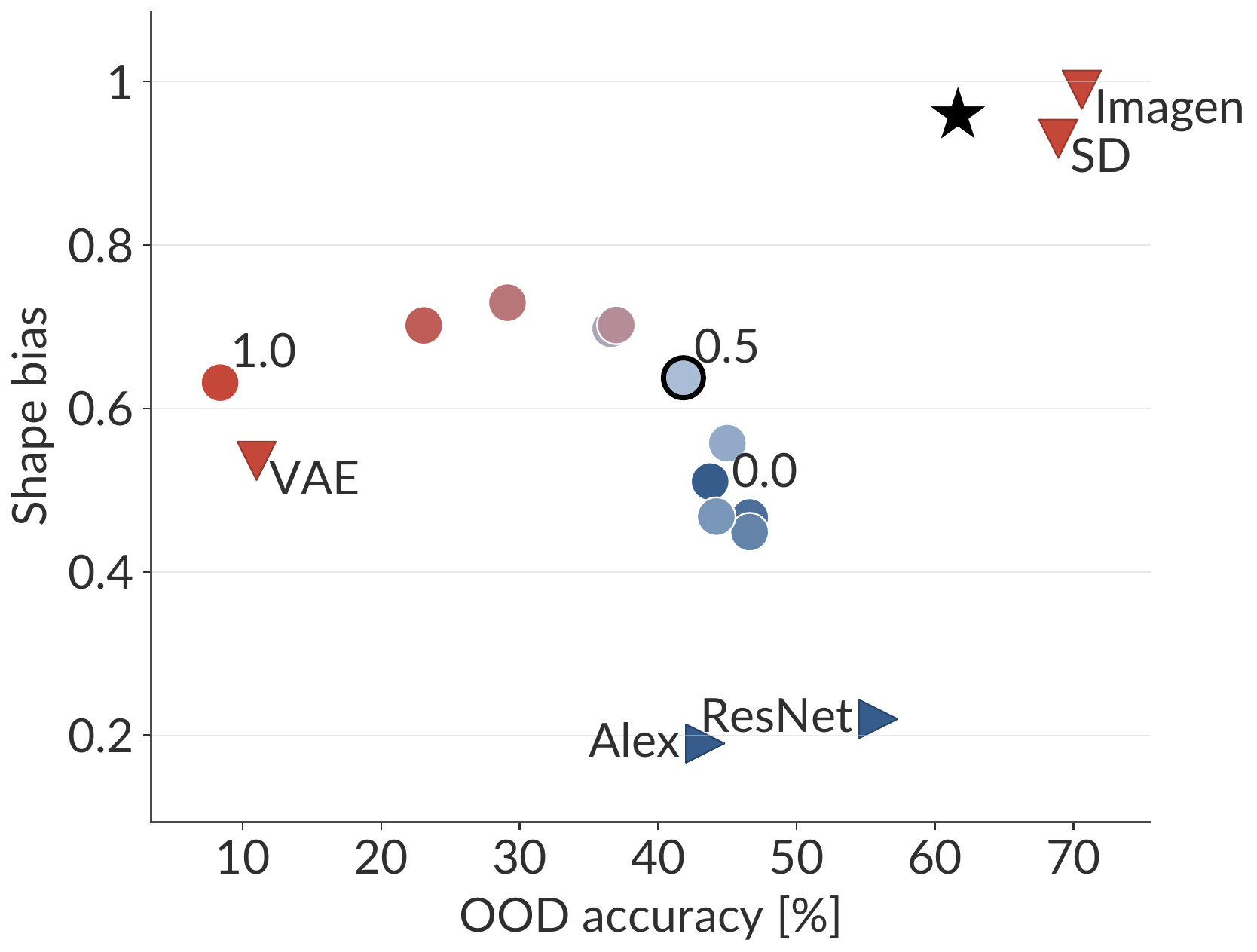}};
\draw [anchor=north west] (0.5\linewidth, 0.18\linewidth) node {\includegraphics[width=0.48\linewidth]{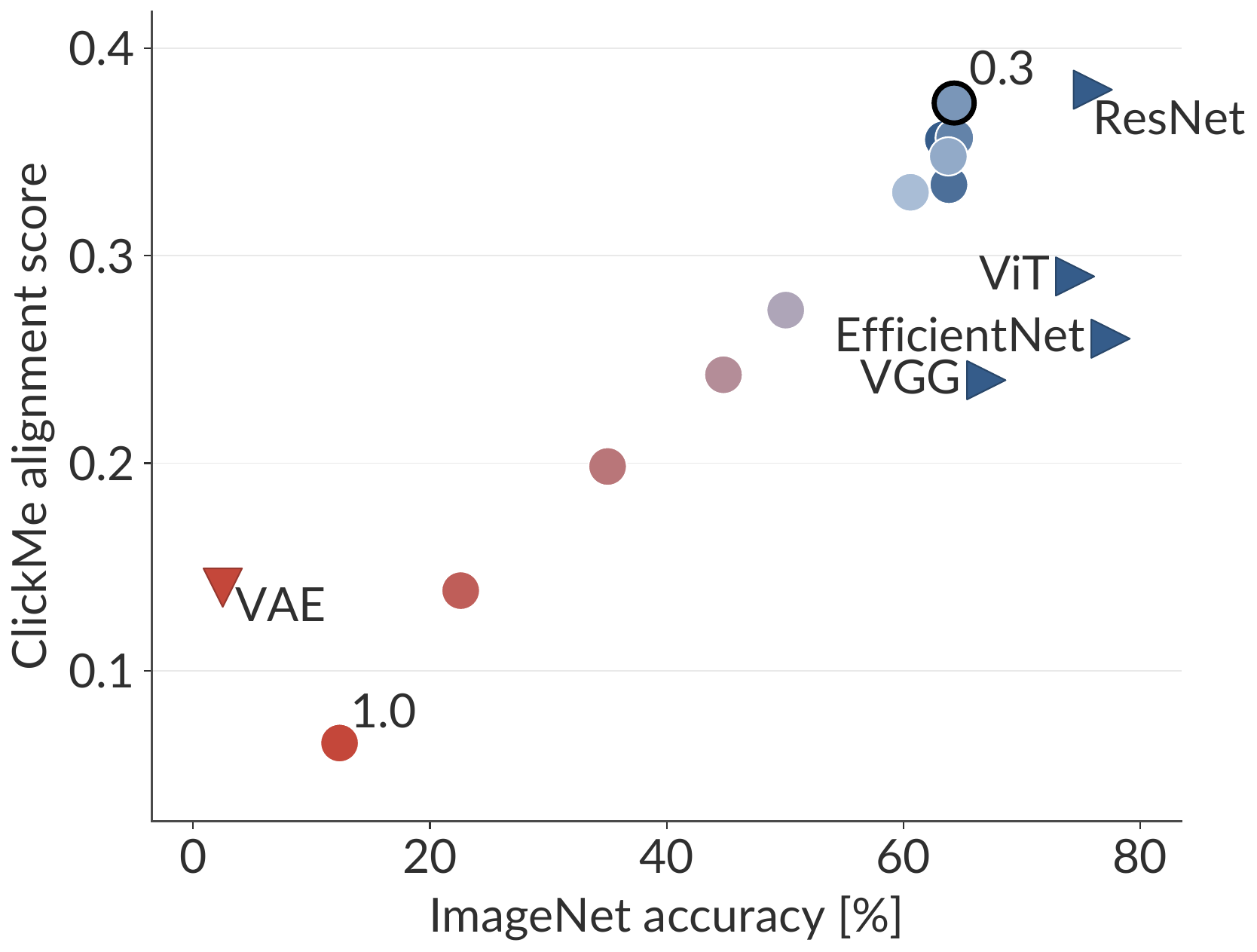}};
\node[anchor=north west, inner sep=0pt] at (0.00\linewidth, -0.2\linewidth)
{\includegraphics[width=1.0\linewidth]{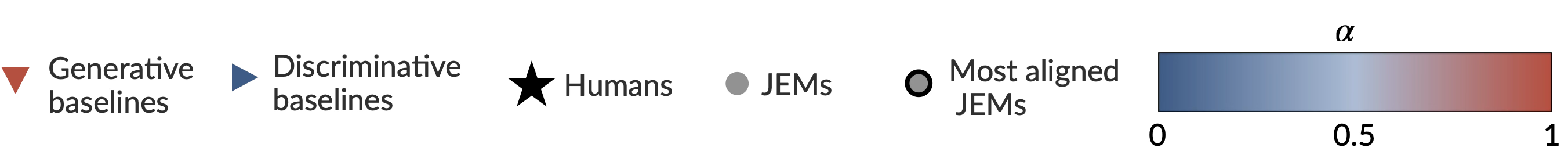}};
\begin{scope}
    \draw [anchor=north west,fill=white, align=left] (0\linewidth, 1\linewidth) node {{\bf a)} \textcolor{perceptual}{Low-level perceptual similarity}} ;
    
    \draw [anchor=north west,fill=white, align=left] (0.5\linewidth, 1\linewidth) node {{\bf b)} \textcolor{perceptual}{Mid-level gloss perception}};

    \draw [anchor=north west,fill=white, align=left] (0.0\linewidth, 0.6\linewidth) node {{\bf c)}  \textcolor{classif}{Response uncertainty in classification}};
    
    \draw [anchor=north west,fill=white, align=left] (0.5\linewidth, 0.6\linewidth) node {{\bf d)} \textcolor{generalization}{Robustness to transformation}};

    \draw [anchor=north west,fill=white, align=left] (0.0\linewidth, 0.2\linewidth) node {{\bf e)} \textcolor{generalization}{Shape-texture cue-conflict}};

    \draw [anchor=north west,fill=white, align=left] (0.5\linewidth, 0.2\linewidth) node {{\bf f)} \textcolor{clickme}{Human diagnostic features}};
\end{scope}
\end{tikzpicture}
%\caption{
%Human alignment across the generative--discriminative continuum. {\bf (a)} Low-level perceptual similarity on BAPPS, measured by JND and 2AFC scores. {\bf (b)} Mid-level gloss perception, measured by gloss classification accuracy and correlation with human gloss judgments. {\bf (c)} CIFAR-10H response uncertainty, measured by accuracy and human-label cross-entropy (lower is better). {\bf (d)} Human-like generalization under transformations, measured by OOD accuracy and observed consistency with humans. {\bf (e)} Shape--texture cue conflict, measured by OOD accuracy and shape bias. {\bf (f)} Diagnostic feature alignment. Across benchmarks, the most human-aligned JEMs generally occur at intermediate values of $\alpha$, suggesting that hybrid objectives better match human visual behavior than either endpoint.
\vspace{-5mm}
\caption{\textbf{Human alignment across the generative--discriminative continuum.} JEMs are evaluated across $\alpha \in [0, 1]$, from purely \textcolor{disc}{discriminative ($\alpha = 0$)} to purely \textcolor{gen}{generative ($\alpha = 1$)}. \textbf{(a)} Low-level perceptual similarity on BAPPS (JND mAP and 2AFC; human ceiling: $83\%$~\cite{zhang2018unreasonable}). \textbf{(b)} Mid-level gloss perception (gloss accuracy and Pearson correlation with human judgments; theoretical upper bound: $r = 1$). \textbf{(c)} CIFAR-10H response uncertainty (CIFAR-10 accuracy and cross-entropy with human distributions, lower is better; human ceiling: $0.55$ nats). \textbf{(d)} Generalization under out-of-distribution transformations (OOD accuracy and error consistency with humans, Cohen's $\kappa$; human ceiling: $\kappa = 0.39$~\cite{geirhos2021partial}). \textbf{(e)} Shape--texture cue conflict (OOD accuracy and shape bias; human reference: ${\approx}\,96\%$~\cite{geirhos2018imagenet}). \textbf{(f)} Diagnostic feature alignment on ClickMe (Spearman correlation normalized by human inter-rater agreement~\cite{fel2022harmonizing}; human ceiling: $1$ by construction). Across all benchmarks, purely discriminative and purely generative JEMs are never the most human-aligned, confirming that hybrid objectives better capture human visual behavior.
}
\label{fig:main_alpha_curves}
\end{figure}
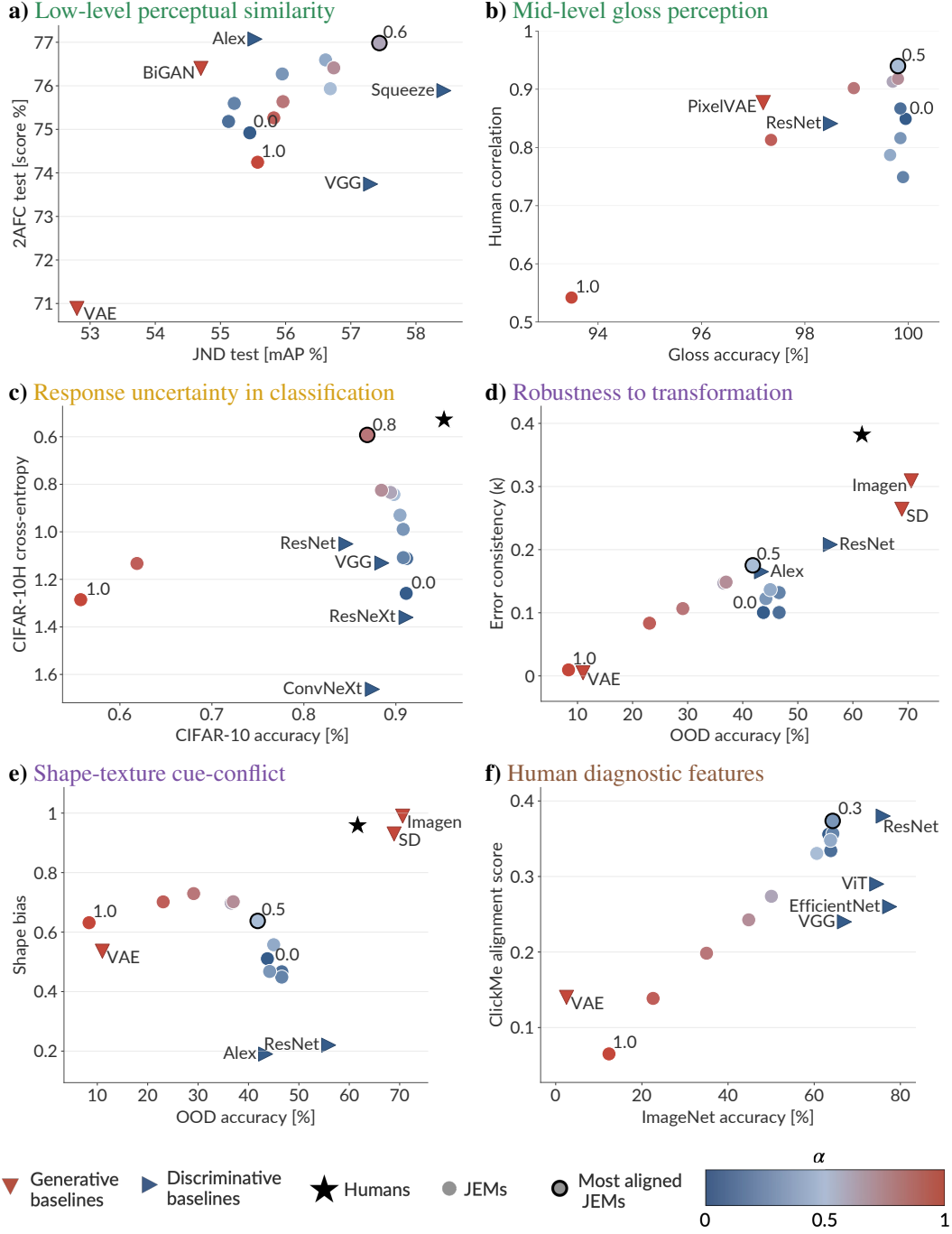

\subsection{A consistent sweet spot across six benchmarks (Fig.~\ref{fig:main_alpha_curves})}

Fig.~\ref{fig:main_alpha_curves} plots JEMs across the full continuum of $\alpha$ (ranging from \dotgen\ \textcolor{gen}{generative} to \dotdisc\ \textcolor{disc}{discriminative}) against a distinct alignment benchmark. Despite the diversity of tasks — spanning low-level image perception, mid-level material perception, classification uncertainty, robustness, cue use, and feature attribution---a single pattern recurs throughout: the most human-aligned models consistently cluster at intermediate values (often around~$\alpha=0.5$), away from both endpoints.

For each benchmark, we also report the established discriminative and generative baselines from the original evaluation protocol, so that JEMs are compared against the relevant reference models rather than a single arbitrary baseline family.
Across benchmarks, the best-aligned JEM consistently falls at intermediate $\alpha$---where alignment is measured as the distance to the human reference in each benchmark's metric space. The optimal value ranges from $\alpha \approx 0.3$ (diagnostic features, see Fig.~\ref{fig:main_alpha_curves}f) to $\alpha \approx 0.8$ (response uncertainty, see Fig.~\ref{fig:main_alpha_curves}c), with most tasks clustering near $\alpha \in [0.5, 0.6]$ (see Fig~\ref{fig:main_alpha_curves}a-b-d-e). The remainder of this section examines each benchmark in detail. To rule out the VAE encoder itself as a confound in Imagenet-based benchmarks (Fig~\ref{fig:main_alpha_curves}a-d-e-f), we include a baseline that uses the same frozen pretrained VAE as our JEMs---but without any JEM training on top. This baseline consistently underperforms hybrid JEMs, confirming that the alignment gains reflect the learned objective rather than the latent representation.

\subsection{\textcolor{perceptual}{Perceptual alignment peaks in hybrid regimes (Fig.~\ref{fig:main_alpha_curves}a-b)}}
We first examine whether the intermediate optimum holds at the perceptual level, where existing comparisons between generative and discriminative models have produced conflicting conclusions. For \textcolor{perceptual}{low-level perceptual similarity} on BAPPS (Fig.~\ref{fig:main_alpha_curves}a), prior work suggested that discriminative models tend to outperform generative ones, especially on the JND task (e.g., SqueezeNet~\cite{iandola2016squeezenet} and AlexNet~\cite{krizhevsky2012imagenet} vs. BiGAN~\cite{brock2019biggan} and VAE~\cite{kingma2014vae}). However, the JEM continuum reveals a richer picture — a hybrid model at $\alpha=0.6$ nearly matches the best discriminative baseline on JND ($57.5\%$ vs. $58.4\%$) while simultaneously achieving the highest 2AFC score, a combination no purely discriminative or generative model achieves. This indicates that the joint pressure to both classify and model the input distribution produces low-level features that are more perceptually calibrated than either objective alone.

For \textcolor{perceptual}{mid-level gloss perception} (Fig.~\ref{fig:main_alpha_curves}b), the pattern is even more informative because it directly revisits a published conclusion.~\citet{storrs2021unsupervised} found that an unsupervised PixelVAE~\cite{zhao2017towards} outperformed a supervised ResNet~\cite{he2016deep} in predicting human gloss judgments, taken as evidence for the superiority of generative learning at this level. Using JEMs to control for architecture, we obtain a different result: the purely discriminative JEM ($\alpha=0$) already outperforms the purely generative JEM ($\alpha=1$), while adding a small generative contribution improves alignment further. Crucially, the best hybrid JEM ($\alpha=0.5$) surpasses both the PixelVAE and ResNet baselines from~\citet{storrs2021unsupervised}. It suggests that the originally reported advantage of generative models was confounded by architectural differences rather than the learning objective per se. Within a controlled architecture, even a small dose of generative pressure is beneficial---but the optimum remains consistently in hybrid territory.

\subsection{\textcolor{classif}{Hybrid objectives better capture human response uncertainty (Fig.~\ref{fig:main_alpha_curves}c)}} 
Having established the intermediate optimum at low and mid levels, we ask whether it extends to the full distribution of human responses on ambiguous images. Fig.~\ref{fig:main_alpha_curves}c shows CIFAR-10H results, where alignment is measured by cross-entropy between the model's output distribution and the empirical distribution of human labels. The lowest cross-entropy---reflecting the closest match to human uncertainty---is obtained at $\alpha=0.8$. Purely discriminative models capture the dominant label but under-represent graded uncertainty, while purely generative models recover some distributional spread at the cost of weakened category structure. Hybrid JEMs strike a balance between these failure modes: they preserve enough categorical structure to agree with the modal human response while remaining sensitive to the ambiguity that leads to disagreement across observers. Notably, this alignment emerges without any direct supervision on human soft labels---unlike~\citet{peterson2019human}, who fine-tune on CIFAR-10H soft labels directly---suggesting it is a natural consequence of the right generative–discriminative balance.

\begin{figure}[ht!]
\vspace{-4mm}
\begin{tikzpicture}
\draw [anchor=north west] (0.0\linewidth, 1\linewidth) node {\includegraphics[width=0.58\linewidth]{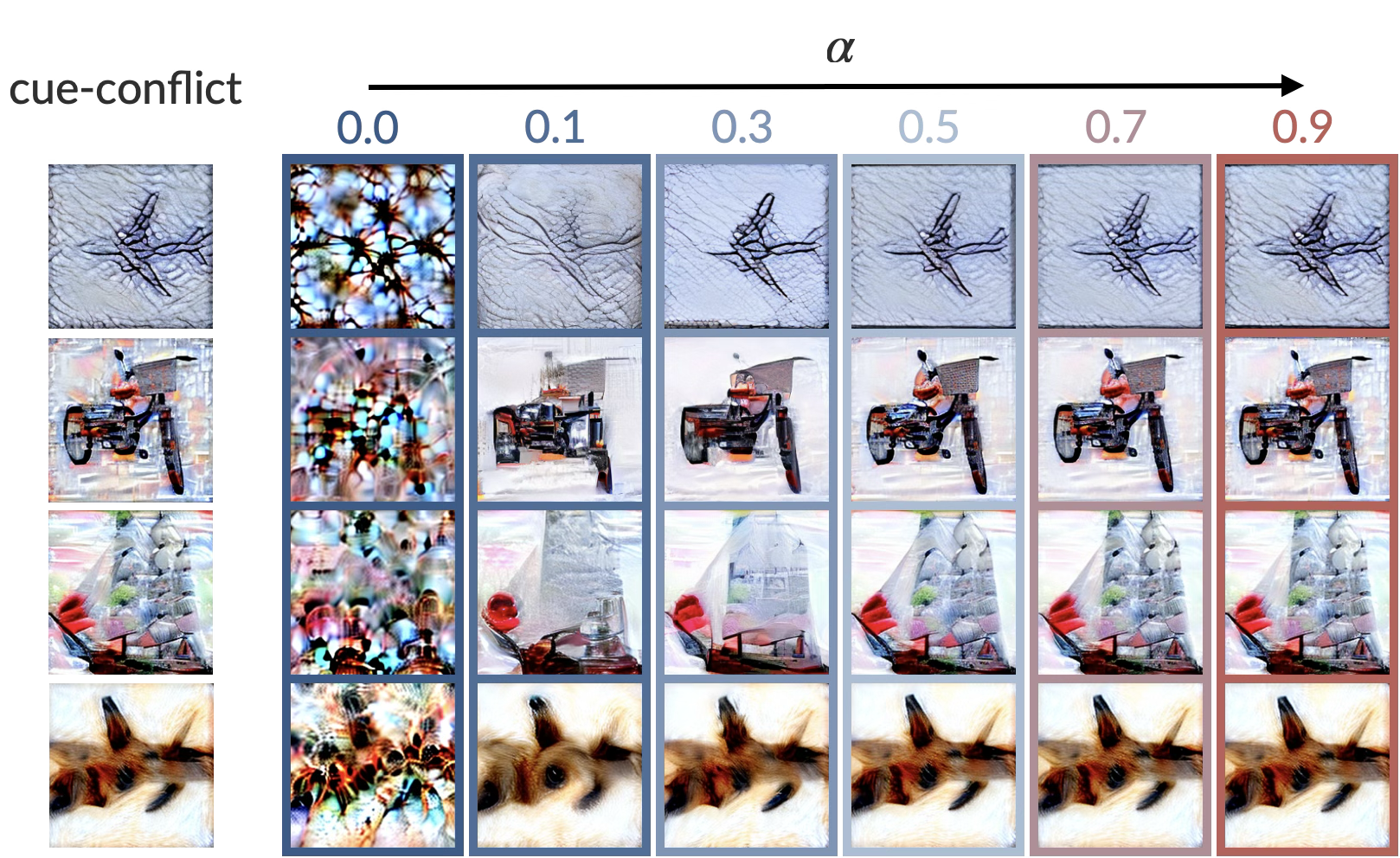}};
\draw [anchor=north west] (0.6\linewidth, 0.98\linewidth) node {\includegraphics[width=0.38\linewidth]{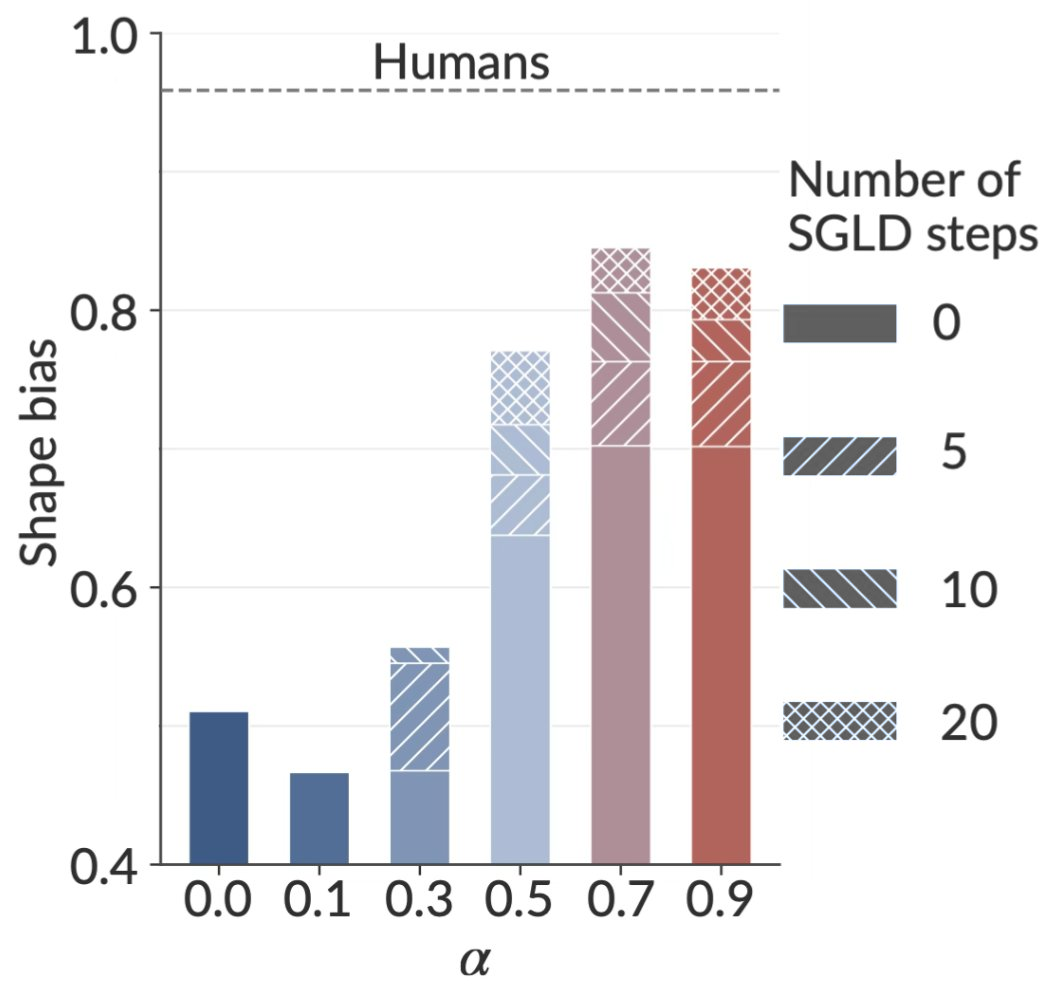}};
\begin{scope}
    \draw [anchor=north west,fill=white, align=left] (0\linewidth, 1\linewidth) node {{\bf a)}};
    \draw [anchor=north west,fill=white, align=left] (0.6\linewidth, 1\linewidth) node {{\bf b)}};
\end{scope}
\end{tikzpicture}
\vspace{-7mm}
\caption{{\bf Generative pressure reveals shape bias.}
{\bf a)} Visualization of a cue-conflict image under the generative component of JEMs trained with different $\alpha$ values; increasing $\alpha$ shifts the visualization from texture-consistent toward shape-consistent.
{\bf b)} Shape bias across SGLD steps for each $\alpha$. SGLD increases shape bias in hybrid/generative JEMs, indicating shape-favoring energy landscapes. The $\alpha=1$ endpoint is omitted because the purely generative model is not trained for categorization, making shape-bias scores unreliable.}
\label{fig:shape-bias}
\end{figure}

\subsection{\textcolor{generalization}{Hybrid objectives promote shape-based generalization (Fig.~\ref{fig:main_alpha_curves}d-e and Fig.~\ref{fig:shape-bias})}}
\label{sec:shape-texture}

We next turn to higher-level visual behavior, where the generative–discriminative trade-off has a mechanistically interpretable effect. Fig.~\ref{fig:main_alpha_curves}d shows that hybrid JEMs improve error consistency with humans on out-of-distribution (OOD) transformations relative to both endpoints of the continuum. However, they remain below large-scale generative baselines such as Imagen~\cite{saharia2022photorealistic} and Stable Diffusion~\cite{rombach2022high}, which benefit from broader training data and language conditioning. Within the JEM family, the improvement is consistent and monotonically ordered around the intermediate regime.

The shape--texture cue-conflict benchmark (Fig.~\ref{fig:main_alpha_curves}e) reveals a particularly clean effect. As $\alpha$ increases, shape bias increases monotonically across the JEM continuum: the best hybrid JEMs reach a shape bias close to $0.7$, well above AlexNet and ResNet-50, and approach the level of shape use seen in large generative models. We attribute this to an asymmetry between the two objectives: a discriminative model can reduce its loss by relying on locally predictive texture cues, whereas a generative model is pressured to assign high probability to globally coherent images, for which shape structure is often a dominant organizing cue. As $\alpha$ increases, the energy landscape increasingly penalizes globally incoherent inputs, gradually shifting learned representations toward global shape.

This mechanism can be exploited at test time via SGLD refinement (Fig.~\ref{fig:shape-bias}). 
Applying Langevin dynamics to a cue-conflict image moves it toward lower-energy regions of the model's energy landscape. 
For high-$\alpha$ models, this favors the shape-consistent interpretation: local, high-frequency texture cues are attenuated, while globally coherent shape structure is preserved. 
Fig.~\ref{fig:shape-bias} shows this effect both qualitatively and quantitatively: visualizations of the generative component shift toward the shape-consistent interpretation as $\alpha$ increases, and SGLD refinement amplifies shape bias for hybrid and generative JEMs ($\alpha \geq 0.3$), with negligible effect at $\alpha=0$. 
Thus, SGLD exposes at inference the representational preference induced by the generative objective, without any retraining.
%This mechanism can be directly exploited at test time via SGLD refinement (Fig.~\ref{fig:shape-bias}). Applying L steps of Langevin dynamics to a cue-conflict image at inference moves it toward the nearest high-probability basin in the model's energy landscape. For high-$\alpha$ models, this basin corresponds to the shape-consistent interpretation: the texture signal, which is local and high-frequency, is progressively attenuated along the chain, while the globally coherent shape structure is preserved. Fig.~\ref{fig:shape-bias}a illustrates this qualitatively--- the low alpha refined images towards texture, and increasing $\alpha$ up to $0.7$ progressively shifts them toward the shape-consistent category. Fig.~\ref{fig:shape-bias}b shows that this refinement systematically amplifies shape bias for hybrid and generative JEMs ($\alpha \geq 0.3$), with the effect negligible for $\alpha=0$, where the flat energy landscape provides no directional signal. SGLD thus makes explicit the representational preference that the generative objective encodes during training, and makes it available at inference without any retraining.

\begin{figure}[h]
\begin{tikzpicture}
\centering
\draw [anchor=north west] (0.0\linewidth, 0.99\linewidth) node {\includegraphics[width=0.95\linewidth]{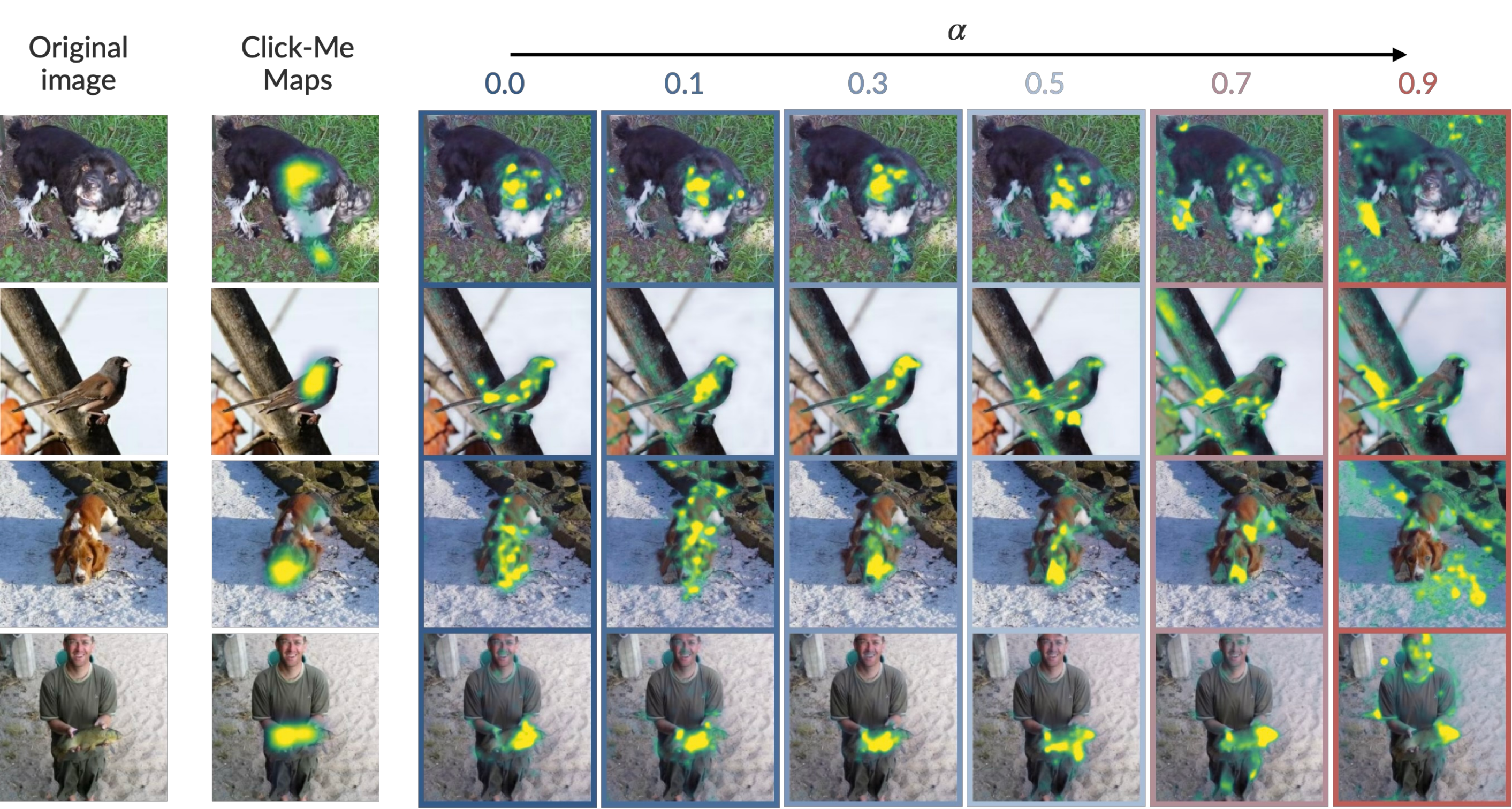}};
\end{tikzpicture}
\vspace{-3mm}
\caption{{\bf Hybrid JEMs align with human saliency.} Original images are shown on the left, followed by human ClickMe maps and model attribution maps for JEMs trained across the generative–discriminative continuum. As the generative contribution increases up to intermediate values, attribution maps become more concentrated on object-relevant regions and better resemble human diagnostic regions. Beyond this hybrid regime, maps become less focused, suggesting that excessive generative weighting weakens diagnostic feature selectivity.}
\label{fig:click-me}
\vspace{-5mm}
\end{figure}

\subsection{\textcolor{clickme}{Hybrid objectives align model attention with human diagnostic regions (Fig.~\ref{fig:main_alpha_curves}f and Fig.~\ref{fig:click-me})}}
%\vspace{-4mm}

Finally, we ask whether the generative–discriminative balance affects not just what the model predicts, but which visual evidence it uses to do so. Using the ClickMe benchmark~\cite{linsley2019,fel2022b}, we compare model attribution maps against human importance maps collected during object recognition. The best alignment is obtained at $\alpha = 0.3$, with both purely generative and purely discriminative models performing worse (see Fig.~\ref{fig:main_alpha_curves}f). The progression across $\alpha$ in Fig.~\ref{fig:click-me} is visually interpretable. 
At $\alpha=0$, attribution maps are diffuse and fragmented; at $\alpha=0.3$, they sharpen around object-relevant regions that humans also mark as diagnostic. 
Beyond this hybrid regime, maps lose selectivity, consistent with weakened discriminative pressure in overly generative models. 
This non-monotonic pattern in Fig.~\ref{fig:main_alpha_curves}f mirrors the shape--texture benchmark: a moderate generative contribution improves alignment, but too much erodes the categorical structure that makes features diagnostic. 
Thus, the generative objective shapes not only model predictions, but also the spatial structure of the representations that support them.

%The progression across $\alpha$ in Fig.~\ref{fig:click-me} is visually interpretable. At $\alpha=0$, attribution maps are diffuse or fragmented, spreading saliency broadly across the image rather than concentrating it on the object. As the generative component increases to $\alpha=0.3$, attribution progressively sharpens onto the object-relevant regions that humans also identify as diagnostic---shapes, contours, and salient object parts. Beyond this hybrid regime, maps begin to lose selectivity: overly generative models become less focused on category-specific features, consistent with the weakening of discriminative pressure. The non-monotonic shape of the ClickMe alignment curve  mirrors the pattern seen in the shape–texture benchmark: a moderate generative contribution is beneficial, but too much erodes the categorical organization that makes features diagnostic. Together, these results suggest that the generative objective shapes not only the model's predictions but also the spatial structure of the representations that underlie them — and that this effect, too, peaks in the hybrid regime.

Taken together, no benchmark favors either endpoint, suggesting that human-like visual representations depend not on choosing between generative and discriminative learning, but on balancing them.
\vspace{-3mm}

\section{Conclusion \& Discussion}
\label{sec:Conclusion}

This paper asked whether human-like visual representations are better explained by generative or discriminative learning. Using JEMs as a controlled interpolation framework, we found that this framing is incomplete: across six qualitatively distinct benchmarks, human alignment is consistently maximized at intermediate values of $\alpha$.

This result reframes the usual opposition between discriminative and generative accounts of vision. Discriminative learning provides categorical structure, whereas generative learning provides sensitivity to the input distribution. Human-like vision appears to require both. This is consistent with neuroscientific theories proposing that the brain combines bottom-up discriminative recognition with top-down generative prediction---as in predictive coding~\cite{rao1999predictive} and 
analysis-by-synthesis~\cite{yuille2006vision}. If this hybrid computation is central to human vision~\cite{peters2024does}, our results offer a machine-learning analog: models become more human-like when trained to balance discriminative recognition with generative modeling. A natural next step is to test whether the same hybrid regimes better predict neural responses and dynamics 
across the visual hierarchy.

Our study has several limitations. JEMs provide a clean testbed, but energy-based training remains computationally demanding and requires careful stabilization. Extending this analysis to modern architectures, larger datasets, and more stable large-scale generative classifiers will be important. Moreover, the optimal value of $\alpha$ varies across benchmarks, suggesting that human alignment is not a single scalar property but a family of constraints spanning perception, uncertainty, generalization, and feature use.

The systematic shift in optimal $\alpha$ across benchmarks is itself informative. We propose that it reflects a gradient from category-level to input-level processing demands: diagnostic feature alignment peaks in a more discriminative regime ($\alpha \approx 0.3$) because humans selectively attend to features that maximally separate categories, a pressure best captured by the conditional $\textcolor{disc}{p(y \mid x)}$; response uncertainty peaks in a more generative regime ($\alpha \approx 0.8$) because human ambiguity tracks the statistical atypicality of images relative to natural image manifolds, a property encoded by $\textcolor{gen}{p(x)}$; and mid-level tasks such as gloss perception and perceptual similarity reside near the center, requiring both input-structure preservation and categorical abstraction. These interpretations are post-hoc and await direct empirical test, but they suggest a principled basis for the variation: the relative weight of generative world-modeling versus discriminative labeling should scale with the degree to which a task depends on input-level rather than category-level information.

Finally, these results have broader implications for human-aligned AI. Hybrid objectives may improve alignment with human perception, uncertainty, and diagnostic feature use, but human-likeness can also reproduce human errors, biases, and overconfident interpretations. Human alignment should therefore be treated as a measurable design objective, not as a guarantee of fairness, safety, or robustness. Progress may require moving beyond the generative-versus-discriminative dichotomy entirely, toward understanding how perceptual systems structure the interaction between recognition and density modeling.

\newpage
{\small

\bibliographystyle{unsrtnat}
\bibliography{ref, ref_tserre}
}

\appendix
\clearpage
\section*{Supplementary Material}
\setcounter{section}{0}
\renewcommand{\thesection}{\Alph{section}}

\section{Extended Related Work}
\label{supp:extended_related_works}

\paragraph{Generative and discriminative theories of vision.}
A longstanding question in vision science is whether human-like visual representations are better explained by discriminative or generative learning principles.
Recent work frames this debate as a contrast between two influential views of vision: one emphasizing direct discriminative inference from sensory data, the other emphasizing perception as inference through the inversion of a generative model of the world~\cite{peters2024does,kriegeskorte2018cognitive}.
On one side, rapid object recognition has often been modeled as a largely feedforward and discriminative process~\cite{riesenhuber1999hierarchical,serre2007feedforward,vanrullen2001timecourse,dicarlo2007untangling,khaligh2014deep}.
On the other, generative accounts emphasize recurrent inference under an internal model, as in predictive coding~\cite{rao1999predictive,friston2005theory,boutin2022pooling} and Bayesian frameworks~\cite{lee2003hierarchical,yuille2006vision,kersten2004object}.
An emerging view is that biological vision recruits both, with their relative contribution depending on task demands and sensory conditions~\cite{kriegeskorte2018cognitive,kreiman2020beyond,kar2021fast}.
Feedforward discriminative processing may dominate rapid, simple recognition, whereas recurrent generative inference may become more important under ambiguity, degradation, or occlusion.
Rather than resolving this debate at the algorithmic level, we ask a more operational question: which balance between discriminative and generative learning yields the most human-aligned representations?

\paragraph{Human alignment of generative and discriminative models.}
A growing body of work has compared generative and discriminative models on human-relevant perceptual and behavioral tasks.
In recognition, supervised discriminative deep networks capture important aspects of primate IT representations~\cite{cadieu2014deep,khaligh2014deep,yamins2014performance,yamins2016using,schrimpf2018brain}, whereas generative classifiers have been reported to show more human-like shape--texture bias, stronger alignment with human classification errors, and greater sensitivity to some perceptual illusions than standard discriminative models~\cite{jaini2024intriguing}.
In perceptual judgment tasks, unsupervised generative models predict human judgments and misperceptions of gloss~\cite{storrs2021unsupervised}, while deep feature spaces align well with human perceptual similarity judgments~\cite{zhang2018unreasonable}.
Related comparisons have also been made in one-shot generation, where generative models differ in how well they reproduce human-like drawing behavior~\cite{boutin2022diversity,boutin2023diffusion,boutin2024latent}.
More generally, comparisons between humans and models reveal substantial gaps in generalization, including robustness to image degradations, shape versus texture reliance, and image-level error patterns~\cite{geirhos2018generalisation,geirhos2018imagenet,geirhos2020beyond}, while controversial stimuli provide a sharper way to test which models best capture human judgments~\cite{golan2020controversial}.
Together, these results suggest that both generative and discriminative objectives can support aspects of human alignment.
However, they do not isolate the role of the learning objective itself, because the compared models often differ simultaneously in architecture, training pipeline, and inductive bias.

\paragraph{Hybrid generative--discriminative models.}
A related line of work explores models that combine generative and discriminative learning rather than treating them as mutually exclusive~\cite{ng2001discriminative,raina2003classification,bouchard2004tradeoff,lasserre2006principled,boutin2020iterative}.
This idea has been developed in several directions, including semi-supervised hybrid training~\cite{druck2007semi,kuleshov2017hybrid,gordon2020combining} and hybrid energy-based classifiers~\cite{larochelle2008classification,grathwohl2020your}.
These approaches are motivated by the complementary strengths of generative and discriminative objectives, but many introduce additional latent variables, separate modules, or partially distinct parameterizations~\cite{kuleshov2017hybrid,gordon2020combining}.
For our purposes, Joint Energy-based Models (JEMs) are especially attractive because they preserve a standard classifier-like architecture while enabling a controlled interpolation between discriminative and generative training within a single scoring function~\cite{grathwohl2020your}.
Subsequent work has further improved their training stability, speed, and empirical performance, making them more practical on modern image benchmarks~\cite{yang2021jempp,yang2022towards}.
This makes JEMs a particularly clean testbed for isolating the representational effect of the learning objective while minimizing architectural confounds.

\paragraph{Positioning of the present work.}
Closest to ours, recent work showed that generative classifiers can outperform standard discriminative models on several human-alignment benchmarks~\cite{jaini2024intriguing}.
Our goal is complementary: rather than comparing distinct model families, we use JEMs to interpolate continuously between generative and discriminative objectives while keeping the architecture fixed.
This lets us test how human alignment varies across that spectrum.
We find that it peaks at an intermediate regime: the most human-like representations emerge neither from purely generative nor purely discriminative training, but from a balance between the two.

\section{EBM training}\label{supp:ebm_training_global}

\subsection{Contrastive divergence}\label{supp:CD}

The demonstration below is adapted from~\cite{woodford2006notes} to fit our notation. Even though this mathematical derivation is not crucial for a good understanding of our work, we include
it to make sure our article is self-contained and complete.

We consider an Energy-Based Model (EBM) defining a probability distribution via the Boltzmann form:
\begin{equation}
p_{\vth}(\vx) = \frac{\exp(-E_{\vth}(\vx))}{Z(\vth)} \quad \text{with} \quad Z(\vth) = \int \exp(-E_{\vth}(\vx)) \, d\vx. \nonumber
\end{equation}

Our goal is to minimize the negative log-likelihood with respect to the empirical data distribution $p_{\mathcal{D}}$:
\begin{equation}
\mathcal{L}_{\text{ML}}(\vth) = \mathbb{E}_{\vx \sim p_{\mathcal{D}}}[-\log p_{\vth}(\vx)]. \nonumber
\end{equation}

We first expand the log-probability:
\begin{equation}
-\log p_{\vth}(\vx) = E_{\vth}(\vx) + \log Z(\vth). \nonumber
\end{equation}

Taking the gradient with respect to $\vth$:
\begin{equation}
\nabla_{\vth} \mathcal{L}_{\text{ML}} = \mathbb{E}_{\vx \sim p_{\mathcal{D}}} \left[ \nabla_{\vth} E_{\vth}(\vx) + \nabla_{\vth} \log Z(\vth) \right]. \nonumber
\end{equation}

The derivative of the log-partition function could be simplified:
\begin{align}
\nabla_{\vth} \log Z(\vth) 
&= \frac{1}{Z(\vth)} \nabla_{\vth} Z(\vth) \nonumber \\
&= \frac{1}{Z(\vth)} \nabla_{\vth} \int \exp(-E_{\vth}(\vx)) \, d\vx \nonumber \\
&= - \frac{1}{Z(\theta)} \int \exp(-E_{\theta}(\vx)) \nabla_{\vth} E_{\vth}(\vx) \, d\vx \nonumber \\
&= - \int p_{\vth}(\vx) \nabla_{\vth} E_{\vth}(\vx) \, d\vx \nonumber \\
&= - \mathbb{E}_{\vx \sim p_{\vth}} \left[ \nabla_{\vth} E_{\vth}(\vx) \right]. \nonumber
\end{align}

Substituting this back into the gradient of the loss:
\begin{equation}
\nabla_{\vth} \mathcal{L}_{\text{ML}} = \mathbb{E}_{\vx \sim p_{\mathcal{D}}} \left[ \nabla_{\vth} E_{\vth}(\vx) \right] - \mathbb{E}_{\vx \sim p_{\vth}} \left[ \nabla_{\vth} E_{\vth}(\vx) \right]. \nonumber
\end{equation}

In practice, we denote the $\vx^{+}$ the "positive" samples from the empirical data distribution $p_{D}$, and $\vx^{-}$ the "negative" samples from the model:
\begin{equation}
\nabla_{\vth} \mathcal{L}_{\text{ML}} \approx \mathbb{E}_{\vx^+ \sim p_{\mathcal{D}}} \left[ \nabla_{\vth} E_{\vth}(\vx^+) \right] - \mathbb{E}_{\vx^- \sim p_{\vth}} \left[ \nabla_{\vth} E_{\vth}(\vx^-) \right]. \nonumber
\label{eq:ml_ebm_derivation}
\end{equation}

\newpage
\subsection{JEM training}\label{supp:ebm_training}

\paragraph{Algorithm.} To train our Joint Energy-Based Models (JEMs), we follow the general approach of~\cite{grathwohl2020}. Algorithm~\ref{algo:train-jem} summarizes the training procedure used in our experiments.

\begin{algorithm}[H]
\SetAlgoLined
\KwIn{Training dataset $\mathcal{D}$, learning rate $\eta$, replay buffer $\mathcal{B}$, initial Langevin step size $\lambda$, noise scale $\sigma$, number of Langevin steps $L$, interpolation weight $\alpha$}
\While{\textnormal{Training}}{
    $\vx^+ \sim \mathcal{D}$ \,\,\,\textcolor{gray}{\# sample from the dataset}\\
    $(\vx, y) \sim \mathcal{D}$ \,\,\,\textcolor{gray}{\# sample a batch for the discriminative term}\\
    $\vx^{0} \sim \mathcal{B}$ \,\,\,\textcolor{gray}{\# sample from a replay buffer}\\
    
    \textcolor{gray}{\# Refine negative samples using Langevin dynamics}\\
    \For{$t \gets 0$ \KwTo $L-1$} {
        $\lambda_t \leftarrow \lambda \left(1-\frac{t}{L}\right)$ \,\,\,\textcolor{gray}{\# linearly decayed step size}\\
        $\vx^{t+1} \leftarrow \vx^{t} + \lambda_t \nabla_{\vx^{t}} \log p_{\vth}(\vx^{t}) + \omega
        \quad \text{with } \omega \sim \mathcal{N}(0,\sigma)$ \\
        $\vx^{t+1} \leftarrow \mathrm{clip}(\vx^{t+1},-1,1)$ \textcolor{gray}{\# optional clipping for stabilization}
    }
    $\vx^{-} = \vx^{L}\textnormal{.detach()}$ \\
    
    $\displaystyle \ell_{G}
    =
    -\left(
    \frac{1}{B}\sum_i f_{\vth}(\vx_i^+) - \frac{1}{B}\sum_i f_{\vth}(\vx_i^-)
    \right)$
    \,\,\,\textcolor{gray}{\# generative loss}\\
    
    $\displaystyle \ell_{D}
    =
    -\frac{1}{B}\sum_i \log p_{\vth}(y_i \mid \vx_i)$
    \,\,\,\textcolor{gray}{\# discriminative loss}\\
    
    $\displaystyle g_G \leftarrow \left\|\nabla_{\vth}\ell_{G}\right\|,\qquad
    g_D \leftarrow \left\|\nabla_{\vth}\ell_{D}\right\|$ \\
    
    $\displaystyle c \leftarrow \frac{g_D}{g_G+\varepsilon}$
    \,\,\,\textcolor{gray}{\# gradient-norm correction}\\
    
    $\displaystyle \ell
    = \alpha \, c \, \ell_{G}
    + (1-\alpha)\ell_{D}$ \\
    
    $\vth \leftarrow \vth - \eta \nabla_{\vth}\ell$ \,\,\,\textcolor{gray}{\# update parameters by gradient descent}\\
    $\mathcal{B} \leftarrow \mathcal{B} \cup \vx^{-}$ \,\,\,\textcolor{gray}{\# update replay buffer}
}
\caption{Training Joint Energy Model using Langevin dynamics and gradient-balanced interpolation between generative and discriminative objectives.}
\label{algo:train-jem}
\end{algorithm}

In the reported experiments, we sweep \(\alpha\) from \(0\) to \(1\) in increments of \(0.1\), yielding 11 models spanning the range from discriminative to generative learning. This provides a controlled continuum that allows us to test whether human alignment and scene-property-related metrics arise from the balance between both, rather than from differences in model class.

%In all experiments, we use \(L=20\) Langevin steps. Within each Langevin chain, we linearly decay the step size from its initial value to zero, which we found helpful for stabilizing training. The initial Langevin step size \(\lambda\) is typically set between \(0.8\) and \(1.0\), and the Langevin noise scale \(\sigma\) between \(9\times 10^{-3}\) and \(2\times 10^{-2}\). We found input noise augmentation with standard deviation between \(0.05\) and \(0.07\) to be helpful for stabilizing training, and use dropout rates of at most \(0.05\), or no dropout at all. Model parameters are optimized with Adam, using learning rates between \(1\times10^{-4}\) and \(5\times10^{-5}\), since larger learning rates tended to destabilize hybrid objectives. We use a cosine learning-rate schedule with 1000 warmup iterations. When both generative and discriminative objectives are active, we rescale the generative term using the ratio of gradient norms, as described in Eq.~\ref{eq:real_loss}, to keep its contribution comparable to that of the discriminative term during training. We do not use batch normalization in the energy model, as in our experiments it tended to destabilize generative training and often prevented convergence.

Unless otherwise noted, the main training settings are as follows:

\begin{itemize}
    \item We use \(L=20\) Langevin steps with a multistep decay schedule for the SGLD step size, which improved stability in all experiments. The initial Langevin step size \(\lambda\) is typically set between \(0.8\) and \(1.0\), and the Langevin noise scale \(\sigma\) between \(9\times 10^{-3}\) and \(2\times 10^{-2}\).
    \item Within each Langevin chain, we linearly decay the step size from its initial value to zero, which we found helpful for stabilizing training.
    \item Model parameters are optimized with Adam, using learning rates between \(1\times10^{-4}\) and \(5\times10^{-5}\), since larger learning rates tended to destabilize hybrid objectives.
    \item We use a cosine learning-rate schedule with 1000 warmup iterations.
    \item We found input noise augmentation with standard deviation around \(0.05\) to be helpful for stabilizing training, and use dropout rates of at most \(0.04\), or no dropout at all.
    \item When both generative and discriminative objectives are active, we rescale the generative term using the ratio of gradient norms, as described in Eq.~\ref{eq:real_loss}.
    \item We do not use batch normalization in the energy model, as in our experiments it tended to destabilize generative training and often prevented convergence.
    \item All JEMs were trained using mixed precision (via PyTorch AMP) and \texttt{torch.compile} to improve training efficiency.
    \item For Gloss and CIFAR-10H, the generative model ($\alpha=1.0$) was selected at the best-FID epoch using Clean-FID~\cite{parmar2021cleanfid}.

\end{itemize}

\paragraph{Compute resources.}
Experiments were run on an internal Slurm cluster. CIFAR-10 experiments used single-GPU jobs on NVIDIA L40S GPUs, with 4 CPU cores and 26\,GB RAM per run. Based on the training logs, a typical CIFAR-10 epoch took about 5.5 minutes, and runs were allocated up to 36 hours of wall-clock time.

Gloss and ImageNet experiments used single-GPU jobs on NVIDIA B200 GPUs, with 6 CPU cores and 80\,GB RAM per run. For gloss, a typical epoch took about 2.1 minutes, and the slowest-converging settings were allocated up to 13 hours of wall-clock time. For ImageNet, a typical epoch took about 17.9 minutes. ImageNet runs were typically allocated up to 4 days, while the highest-\(\alpha\) settings (\(\alpha=0.7,0.8,0.9,1.0\)) were allocated up to 5 days because they took longer to converge. 

Full \(\alpha\)-sweeps consisted of 11 runs launched in parallel, so the elapsed wall-clock time for a sweep was determined by the slowest job, while total compute scaled with the number of parallel runs. These values are reported as wall-clock budgets for the main runs. The full research project required additional preliminary and failed runs beyond the final experiments reported in the paper.

\paragraph{Existing datasets, codebases, and pretrained models.}
Our experiments use existing public datasets, codebases, and pretrained models, including CIFAR-10, CIFAR-10H, the gloss datset, the Model-vs-Human benchmark suite, ClickMe-related evaluation assets, texture--shape cue-conflict benchmarks, the JEM codebase, \texttt{pytorch\_image\_classification} models, LPIPS/PerceptualSimilarity, the Harmonization repository, and the \texttt{stabilityai/sd-vae-ft-mse} AutoencoderKL checkpoint. We cite the original papers and repositories for all of these assets throughout the paper and appendix, and report licenses when clearly stated in the original source, including Apache-2.0 for JEM, MIT for \texttt{pytorch\_image\_classification}, Harmonization, and \texttt{stabilityai/sd-vae-ft-mse}, CC BY-NC-SA 4.0 for CIFAR-10H, and CC BY 4.0 for the gloss and texture--shape cue-conflict datasets. For assets whose official public source did not clearly expose a license, we cite the original source and follow stated access and usage conditions.

\subsection{Architecture of the Energy Function}\label{supp:ebm_arch}

Our implementation is based on the public code released by Grathwohl et al.~\cite{grathwohl2020your}, which parameterizes the energy function with a Wide Residual Network (WRN) backbone~\cite{zagoruyko2016wide}. Relative to the standard WRN formulation, this implementation uses LeakyReLU activations, configurable normalization layers that can be disabled entirely, and optional dropout, all of which we found useful for stabilizing hybrid generative--discriminative training. For smaller-scale datasets such as CIFAR-10 and gloss, we use WRN-28-10 and WRN-22-10, respectively, where the first number denotes network depth and the second the widening factor.

%Interestingly, mildly hybrid settings (\(\alpha=0.1, 0.2, 0.3\)) achieved about a \(1\%\) improvement in top-1 accuracy, suggesting that the generative objective may provide a modest regularizing effect.

\subsection{ImageNet training}\label{supp:ebm_arch_imagenet}

For ImageNet, we use WRN-22-8 to obtain a more computationally manageable model while retaining the same architectural family. Rather than training directly in pixel space, we train the model in the latent space of a frozen pretrained AutoencoderKL~\cite{rombach2022high}, using the \texttt{stabilityai/sd-vae-ft-mse} checkpoint. Operating in latent space substantially reduces the dimensionality of the input, which makes hybrid and fully generative training at ImageNet scale computationally feasible. Concretely, images are first encoded into latent representations, and these latents are then used as inputs to the energy model (see~\cite{bethune2025}). We use the posterior mean of the encoder, rather than posterior sampling, yielding a deterministic latent representation; in our experiments, this option was more stable for training than sampling from the posterior.

As a sanity check, we trained a purely discriminative WRN with the same hyperparameters directly in pixel space and found that it achieved essentially the same top-1 accuracy as the \(\alpha=0\) JEM trained in VAE latent space. This suggests that operating in the VAE latent space preserves discriminative performance at a level comparable to the pixel-space baseline.

\begin{figure}[h]
    \centering
    \begin{subfigure}{\linewidth}
        \centering
        \includegraphics[width=\linewidth]{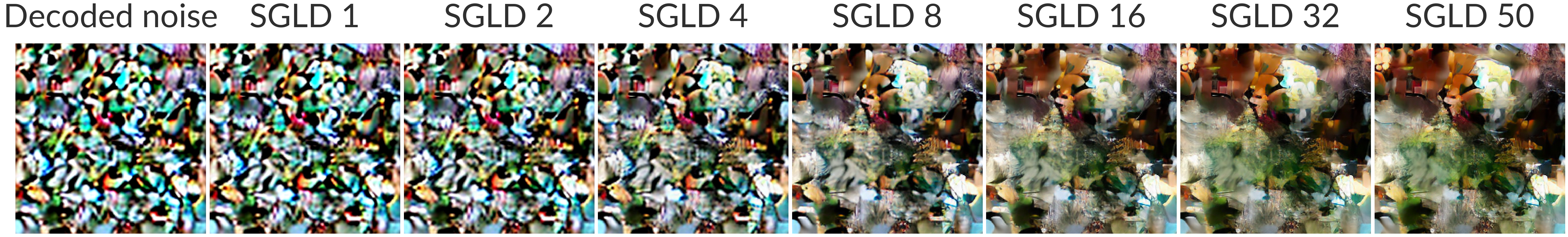}
        \caption{First training epoch.}
        \label{fig:imagenet_sgld_first_epoch}
    \end{subfigure}

    \vspace{1em}

    \begin{subfigure}{\linewidth}
        \centering
        \includegraphics[width=\linewidth]{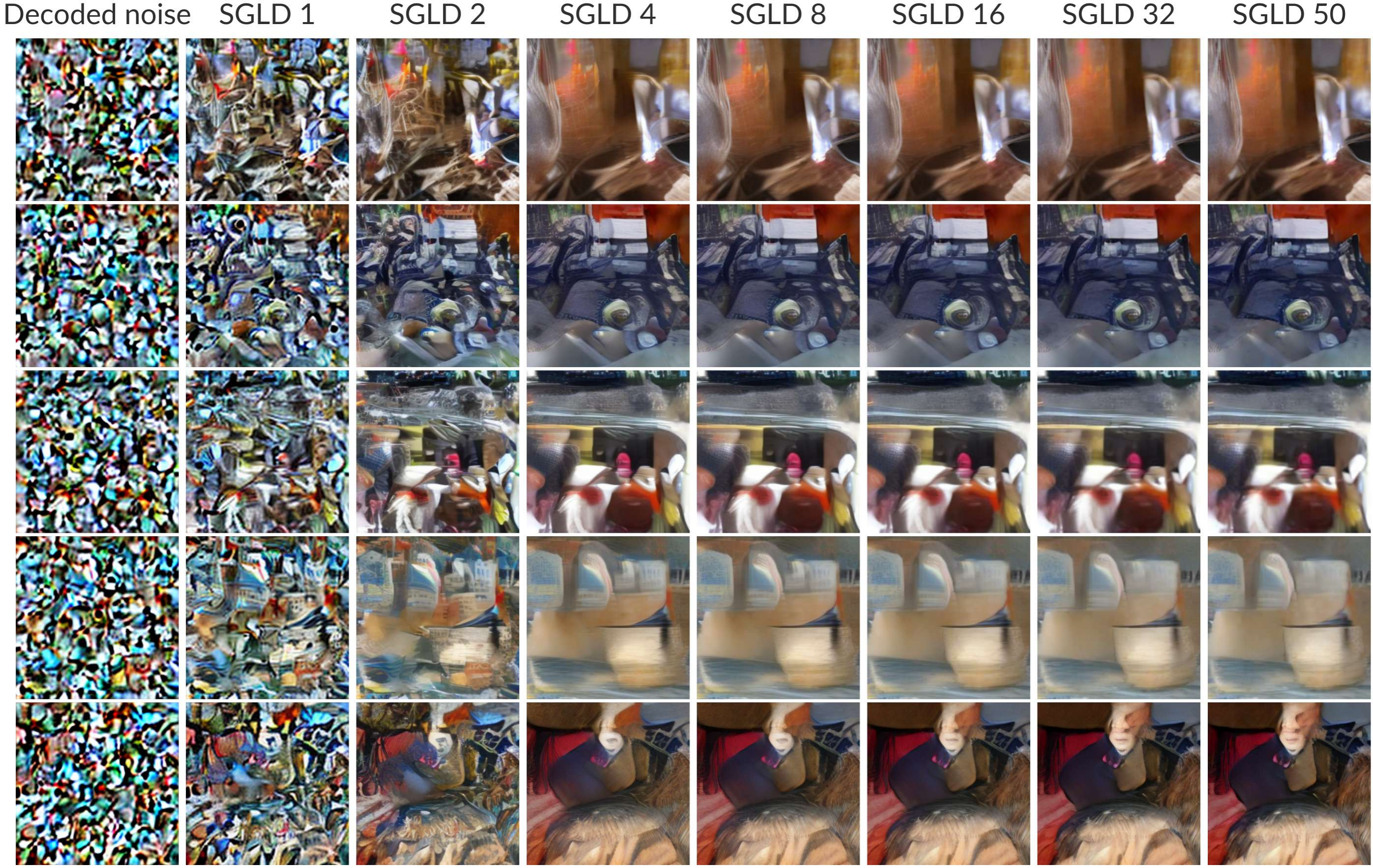}
        \caption{Best validation epoch.}
        \label{fig:imagenet_sgld_best_epoch}
    \end{subfigure}

    \caption{Latent-space sampling trajectories for an ImageNet JEM with $\alpha=0.5$. Each row starts from the VAE latent of a random noise image, and successive columns show the decoded sample after increasing numbers of SGLD refinement steps. (a) shows trajectories at the first training epoch; (b) shows trajectories at the best validation epoch.}
    \label{fig:imagenet_sgld}
\end{figure}

\begin{figure}[h]
    \centering
    \includegraphics[width=1\textwidth]{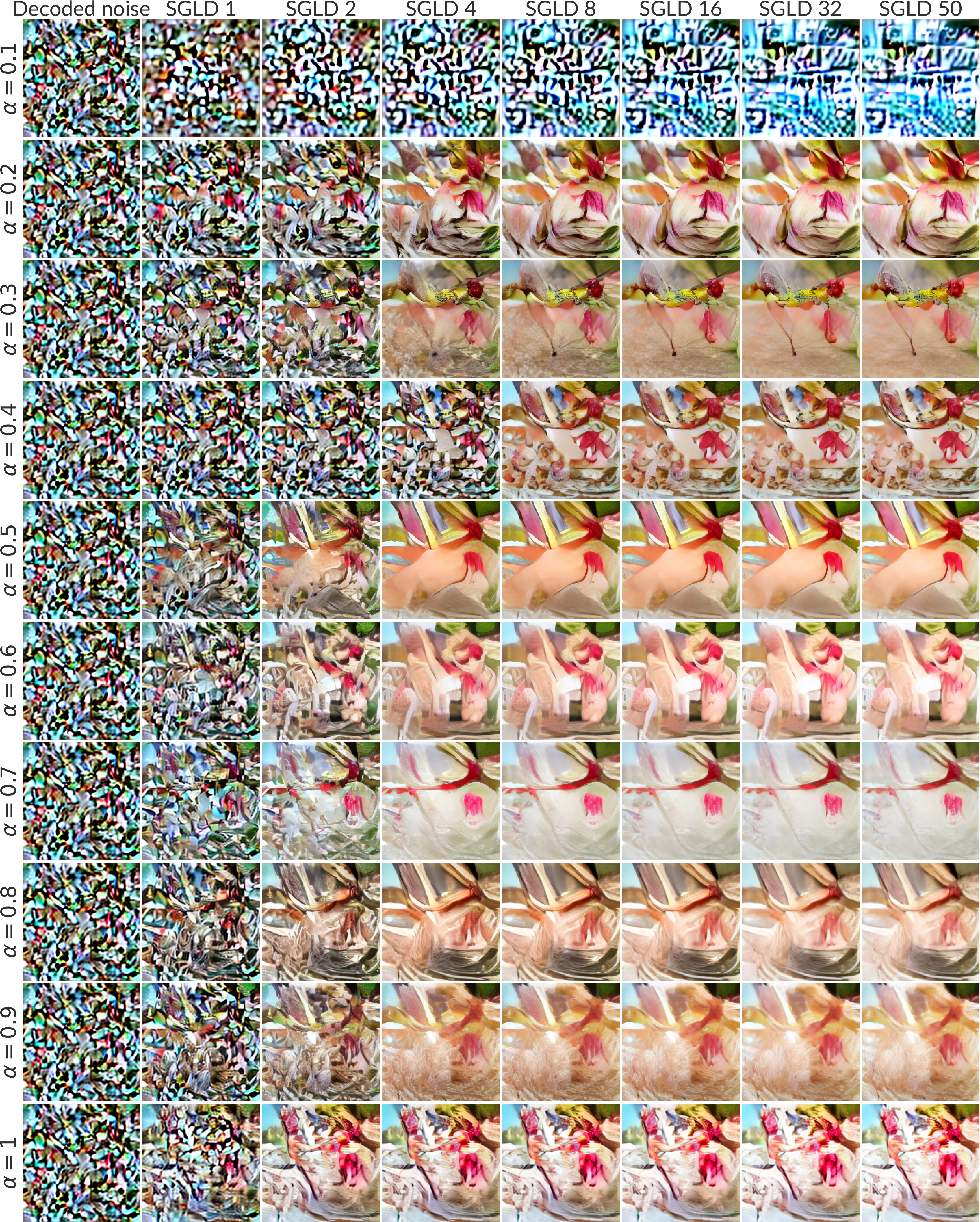}
    \caption{Qualitative generations obtained after 50 SGLD steps from the same shared latent initialization across different values of \(\alpha\). The same shared initial latent code and the same sampling seed are used for all models, so that visual differences more directly reflect the effect of \(\alpha\). The first column shows the VAE decoding of the shared initial latent, and the remaining columns show the decoded samples after refinement under each model.}
    \label{fig:imagenet_alpha_rows}
\end{figure}

\paragraph{Qualitative sampling behavior.} Figures~\ref{fig:imagenet_sgld_first_epoch} and~\ref{fig:imagenet_sgld_best_epoch} illustrate the effect of latent-space SGLD refinement for an ImageNet JEM with \(\alpha=0.5\). We begin from a random noise image, encode it with the VAE, and use the resulting latent as the starting point for SGLD under the learned energy function. The first column shows the VAE decoding of this initial latent, while subsequent columns show the decoded latent after increasing numbers of SGLD refinement steps. Early in training, this refinement produces only weakly organized image structure, whereas by the best validation epoch, it yields substantially more coherent generations. Most of the qualitative improvement occurs within the first several refinement steps, after which the samples largely stabilize. Figure~\ref{fig:imagenet_alpha_rows} compares generations obtained from the same initialization across values of \(\alpha\), illustrating how the balance between discriminative and generative training affects the resulting samples. The figures are intended as qualitative illustrations of the sampling dynamics rather than quantitative evaluation metrics.

\newpage
\newpage
\section{Low-level Perceptual Similarity Benchmark}\label{supp:bench_perceptual}

\begin{figure}[h]
    \centering
    \includegraphics[width=.44\textwidth]{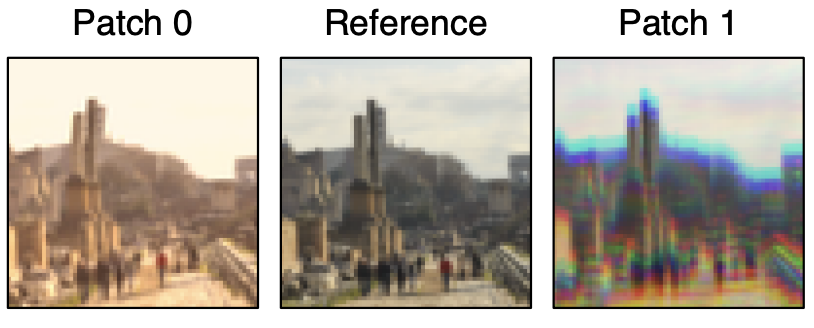}
    \caption{Example of a reference image and its distorted patches.}
    \label{fig:lpips_example}
\end{figure}

To evaluate alignment with human judgments of \textbf{low-level perceptual similarity}, we used the BAPPS dataset, which contains approximately 484,000 human judgments. As mentioned, its main benchmark is a two-alternative forced choice (2AFC) task in which, given a reference patch and two distorted versions of it, observers indicate which distorted patch is more similar to the reference (Figure~\ref{fig:lpips_example}). The dataset includes a wide range of traditional distortions, CNN-based artifacts, and outputs from real image-processing algorithms, making it a broad test for determining whether learned representations preserve perceptual similarity in a way that agrees with human judgments. 

BAPPS also includes a just noticeable difference (JND) task to measure sensitivity to small perceptual changes. In the JND task, observers are required to judge whether a distorted path and the reference image appear perceptually the same or different. Models are then evaluated through similarity, by computing a distance measure for each image pair and ranking pairs from most to least similar; a good model assigns small distances to pairs that humans most often judge as the same. %Both tasks prove to be useful for positioning our method relative to supervised, self-supervised, and unsupervised models on a common perceptual task grounded in human judgments.

\subsection{Models and evaluation}\label{supp:bapps_models_eval}

\begin{figure}[!htbp]
    \centering
    \begin{tikzpicture}
        \node [anchor=south west, inner sep=0pt] (img) at (0,0)
            {\includegraphics[width=0.46\linewidth]{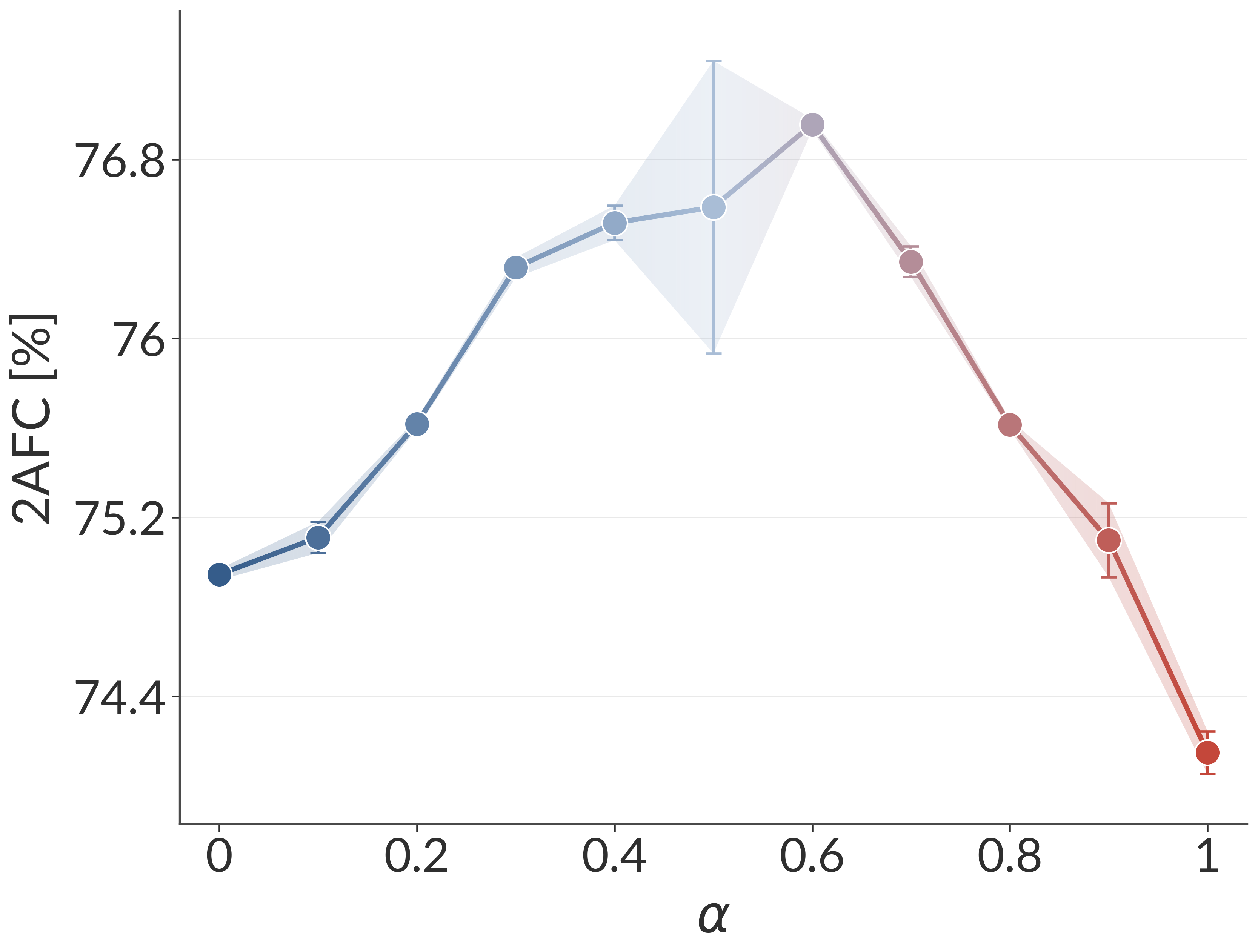}%
             \includegraphics[width=0.46\linewidth]{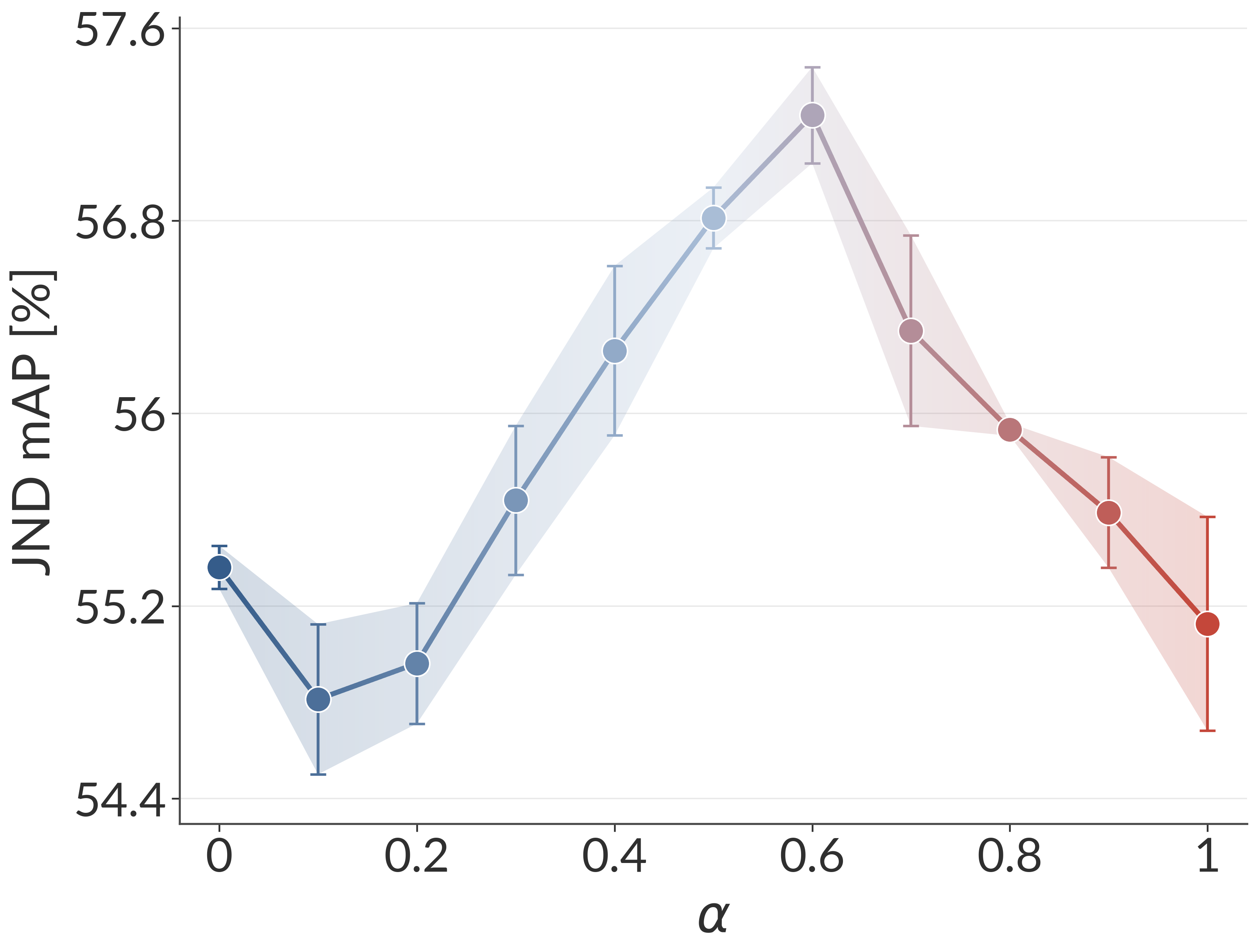}};

    \end{tikzpicture}
    \caption{2AFC accuracy and JND mAP on the BAPPS perceptual similarity benchmark, across JEM $\alpha$ values. Shaded regions indicate the standard error of the mean (SEM) across two seeds.}
    \label{fig:bapps_2afc_jnd}
\end{figure}

We evaluated JEMs on BAPPS using the standard 2AFC and JND protocols (Figure~\ref{fig:bapps_2afc_jnd}), following Zhang et al.~\cite{zhang2018unreasonable}. For 2AFC, performance is reported as agreement with human preference judgments; for JND, we rank image pairs by increasing perceptual distance and report mean average precision (mAP). %so that a good metric assigns smaller distances to pairs that humans judge as perceptually the same. 
%Both tasks prove to be useful for positioning our method relative to supervised, self-supervised, and unsupervised models on a common perceptual task grounded in human judgments.

To evaluate a JEM, we construct an LPIPS-style perceptual distance from intermediate activations of the WRN backbone, using features extracted from the three main residual stages. We evaluate the raw distance induced by the pretrained representation alone, using equal layer/channel weighting as in the uncalibrated setting of Zhang et al.~\cite{zhang2018unreasonable}. Thus, our BAPPS results test whether perceptual similarity emerges naturally, rather than from direct supervision on BAPPS itself.

For comparison, we include the pretrained supervised baselines released by Zhang et al.~\cite{zhang2018unreasonable}. For BiGAN, because the implementation and pretrained checkpoint were not publicly available, we report the values provided in the original paper. We additionally report a VAE baseline from the frozen AutoencoderKL used in our ImageNet experiments, using features from encoder down-blocks 0--3. This baseline provides a reference point for the perceptual quality of the pretrained latent representation itself to determine whether JEM training improves upon the VAE features. %For ImageNet-trained JEMs, reported BAPPS scores are averaged over two independently trained seeds.

\newpage
\section{Gloss and depth perceptual benchmark}\label{supp:bench_gloss_depth}

\begin{figure}[h]
    \centering
    \includegraphics[width=.9\textwidth]{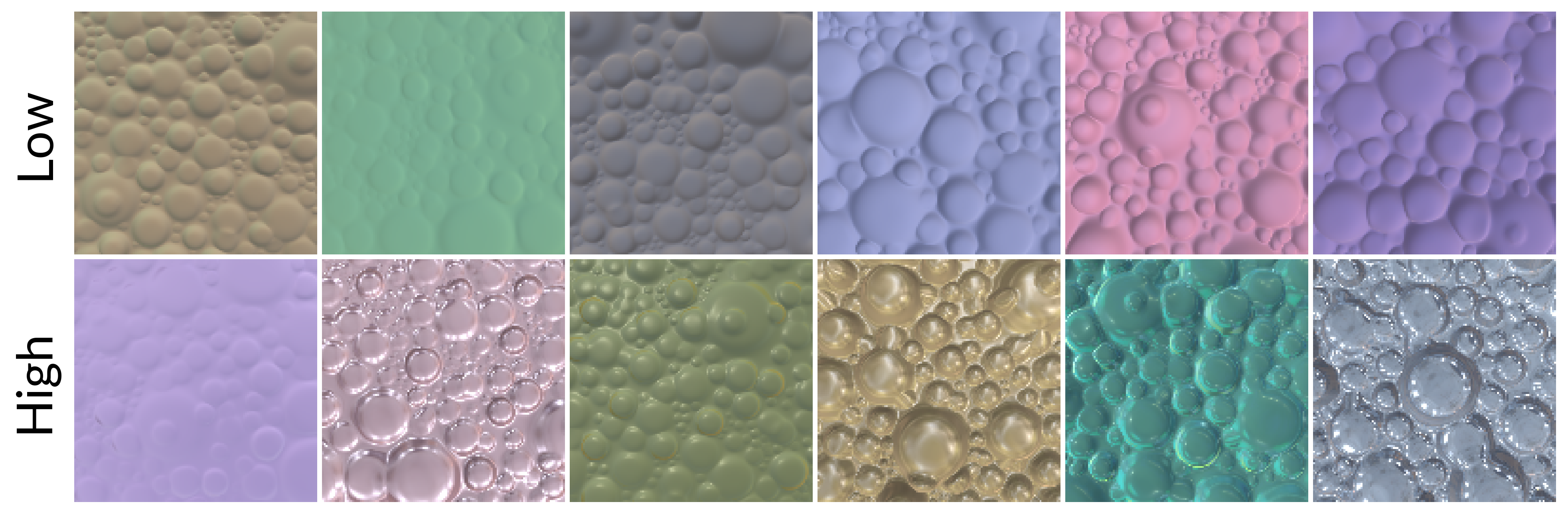}
    \caption{Examples of images labeled as Low or High gloss.}
    \label{fig:gloss_dataset}
\end{figure}

The gloss perception dataset probes \textbf{mid-level material perception}. This is a challenging task because it requires distinguishing surface reflectance properties while disentangling them from other physical factors, such as illumination and surface geometry, that also shape image appearance. We use the dataset introduced by Storrs et al.~\cite{storrs2021unsupervised}, which includes visual stimuli, class labels, and human ratings. One component of the dataset consists of 9,998 rendered images of bumpy surfaces with varying surface relief, diffuse color, lighting, and specular reflectance properties, together with binary labels indicating high versus low gloss (Figure~\ref{fig:gloss_dataset}). The authors also provide a second set of 50 rendered surfaces with randomly sampled specular reflectance magnitudes, ranging from almost matte to almost mirror-like, along with ratings from 20 human observers.

\subsection{Models and data}\label{supp:gloss_models}

In general, and to make direct comparisons possible, we followed Storrs et al. ~\cite{storrs2021unsupervised} as a blueprint for our gloss experiments. We used the rendered bumpy-surface images described in the previous section. Unless otherwise noted, all images were resized to \(128\times128\) pixels and normalized to the range \([-1,1]\).

Because the original dataset did not include a validation split, we introduced one for model selection while keeping the test set held out for final evaluation. Specifically, we partitioned the 9,998 images into 3 subsets: 8,000 training images, 1,000 validation images, and 998 test images.

In the original study, the authors compared an unsupervised PixelVAE model~\cite{zhao2017towards} with a supervised ResNet-18 model~\cite{he2016deep}. For direct comparison, we trained the same two model classes using the public implementation provided in the Storrs et al. ~\cite{storrs2021unsupervised} repository and evaluated them with the same downstream procedures, including the human-alignment analysis. For PixelVAE, we followed the architecture and hyperparameters reported in the original study, including six residual blocks with three layers of 64 convolutional feature maps. For ResNet-18, we followed the setup consisting of three residual blocks each made up of three layers of 56 convolutional feature maps, with a fully connected latent-code layer, and a 2-unit softmax classification head. In both models, the bottleneck representation was used for the downstream probing analyses. For JEMs, we inserted a small projection head preceded by a concat-ELU nonlinearity, motivated by its use in the PixelVAE architecture from the original study. All reported JEM results use this parameterization.

\subsection{Training protocol}\label{supp:gloss_training}

We first reproduced the original findings using their reported hyperparameters and data split, obtaining nearly identical metrics. We then retrained both PixelVAE and ResNet18 on the same data split used for the JEMs to provide fair baselines for comparison. For JEMs, PixelVAE, and ResNet18 alike, we considered latent dimensionalities of 10, 100, 500, and 2000. PixelVAE baselines were trained with 10 random seeds. In contrast, the ResNet18 baselines used one seed, and each of the eleven JEM variants were trained with two seeds per condition. Additionally, we used mild label smoothing of 0.05 for the JEMs, which we found helpful for stabilizing training in the binary classification setting.

The next step was to evaluate each model through a fixed-dimensional latent representation. In their analysis ~\cite{storrs2021unsupervised}, the bottleneck layer of the PixelVAE was treated as the model's latent feature space and used for downstream analyses of gloss, depth, and human alignment. For our experiments, aditionally, we needed to assess how the balance bewteen discriminative and generative learning affected both alignment with human judgements and the recoverability of task-relevant scene properties. To do this and, analog to what was done in the reference material, we extracted a fixed-dimensional representation from each JEM and applied the same downstream probes to that representation. 

Following the general analysis protocol of Storrs et al.~\cite{storrs2021unsupervised}, we extracted a fixed-dimensional latent representation and used it to assess the recoverability of gloss, illumination, and surface relief. All downstream probes were trained on the non-test portion of the data, (training  and validation images), and were evaluated on the test set of images.

\subsection{Evaluation metrics}\label{supp:gloss_probes}

We report three scene-property metrics: \textbf{(i)} gloss classification accuracy, \textbf{(ii)} six-way illumination classification accuracy, and \textbf{(iii)} surface-relief prediction accuracy measured by \(R^2\).

\textbf{Gloss accuracy} was measured directly from the models' classifier head. For fully generative models, we instead trained a linear SVM on the extracted representation, again using five-fold cross-validation on the non-test portion of the data to select \(C\). The linear SVM additionally provides decision values, which were used in the human-alignment analysis described ahead.

For \textbf{light field} classification, we trained a linear SVM, and the regularization parameter \(C\) was selected by using five-fold cross-validation within the non-test portion of the data. 

For \textbf{surface-relief prediction}, we trained a linear regression model and report test-set \(R^2\) for predicting bump height.

In all cases, latent features were standardized using a \texttt{StandardScaler} before fitting the downstream model.

\subsection{Human alignment}\label{supp:gloss_human_alignment}

To evaluate alignment with human perception, we compared model-derived gloss predictions with  the human gloss ratings on the 50 surfaces rendered by Storrs et al.~\cite{storrs2021unsupervised}. These were rated by 20 human observers on a six-point scale from 1 (matte) to 6 (glossy).

As done in our reference study, we derived a continuous gloss prediction for each image from the decision value of a glossy-versus-matte linear SVM, that is, the signed distance of the image representation from the gloss-separating hyperplane. For model families with multiple training instances, predictions were computed separately for each instance and then averaged across instances for each of the 50 stimuli.

We quantified human alignment by computing the correlation between these continuous model predictions and the mean human gloss rating for each of the 50 stimuli. Higher correlation indicates that the learned representation gives rise to gloss predictions that are more consistent with human perceptual judgments.

The same evaluation procedure was applied to JEMs, PixelVAE seeds, and ResNet18 baselines, so that human-alignment comparisons were based on a matched protocol across model families.

\subsection{Further discussion}

\begin{figure}[h]
\centering
\begin{tikzpicture}
    % ===== Top row =====
    \draw [anchor=north west] (0.0\linewidth, 0.95\linewidth)
    node {\includegraphics[width=0.48\linewidth]{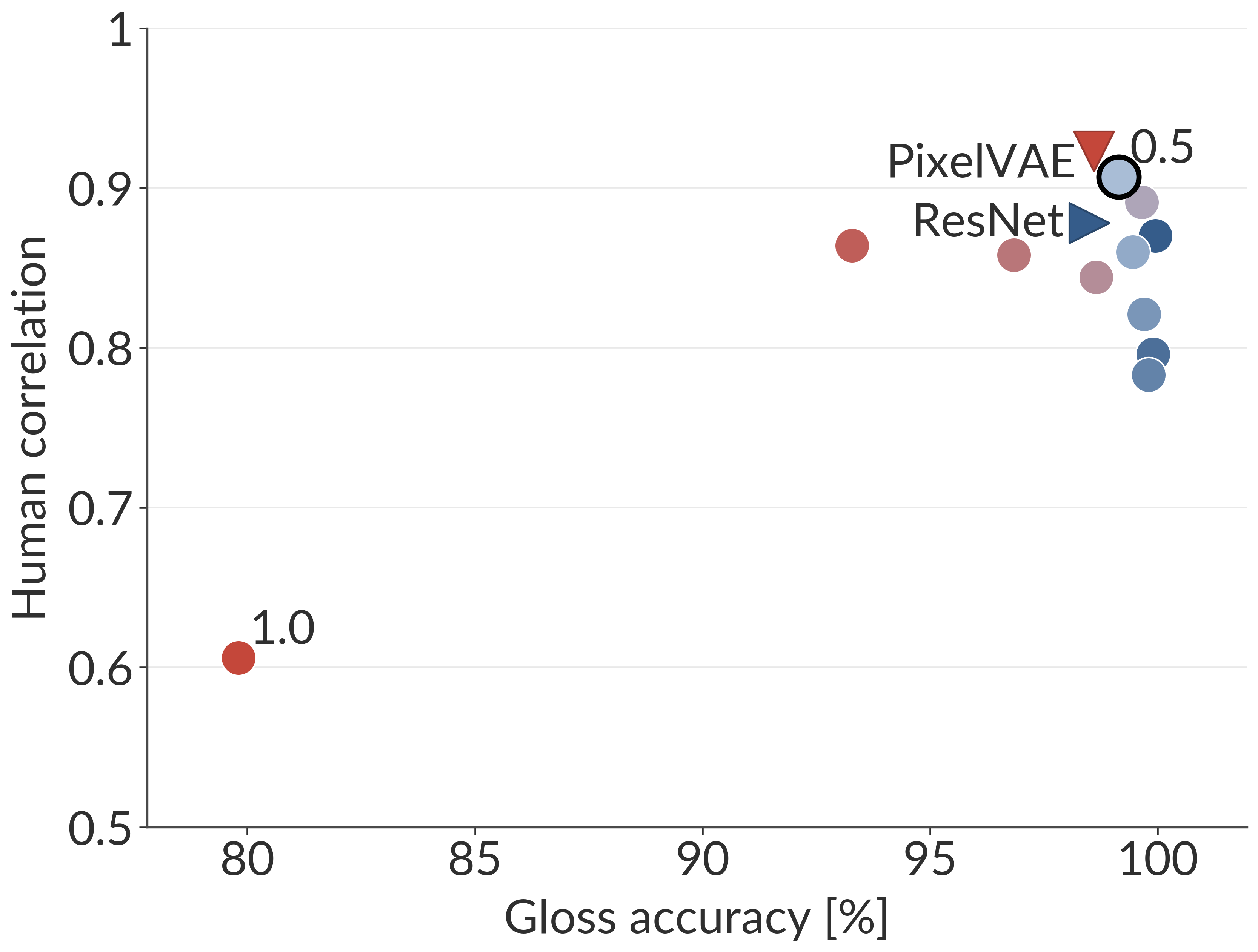}};
    \draw [anchor=north west] (0.5\linewidth, 0.95\linewidth)
    node {\includegraphics[width=0.48\linewidth]{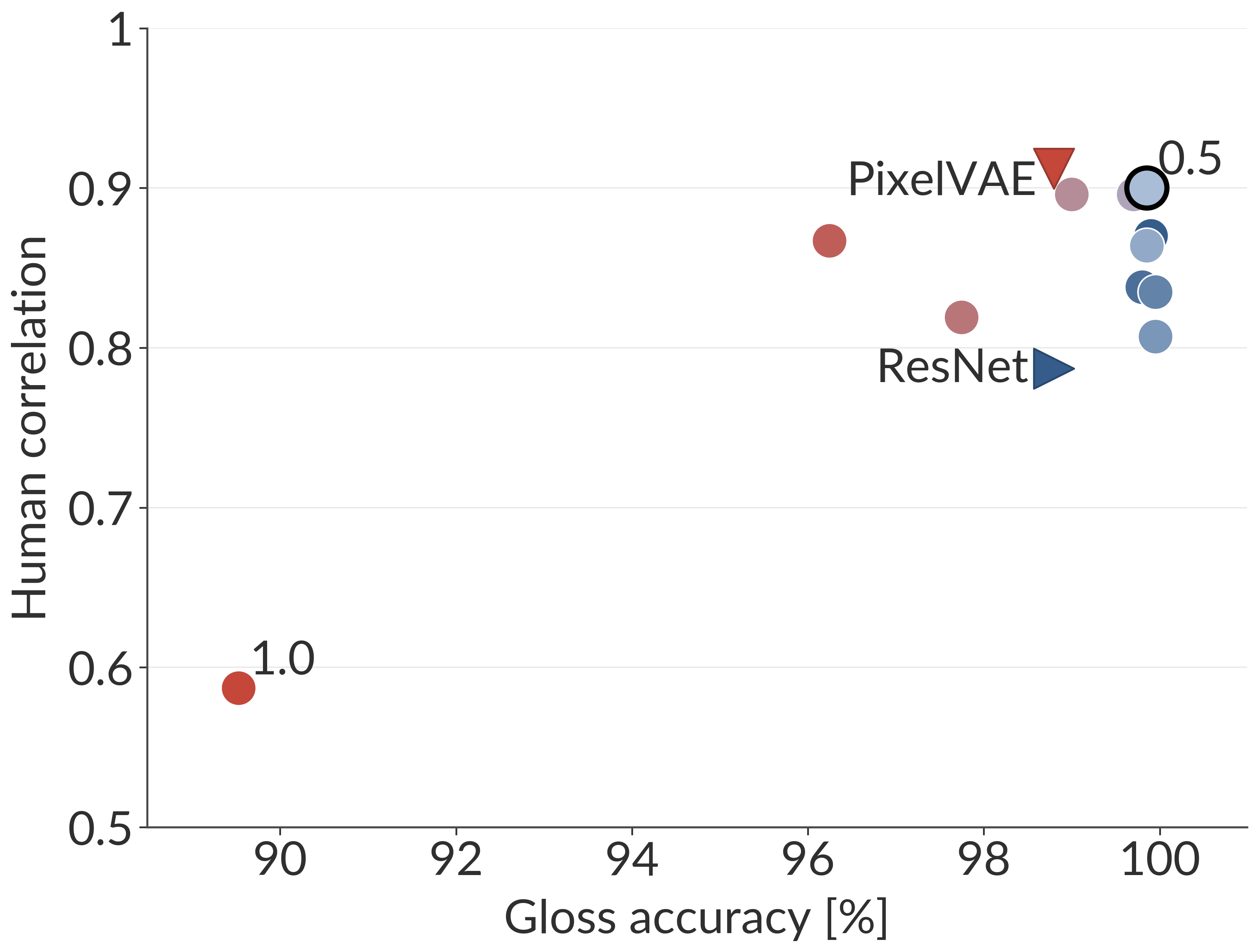}};

    % ===== Bottom row (pushed lower for more separation) =====
    \draw [anchor=north west] (0.0\linewidth, 0.50\linewidth)
    node {\includegraphics[width=0.48\linewidth]{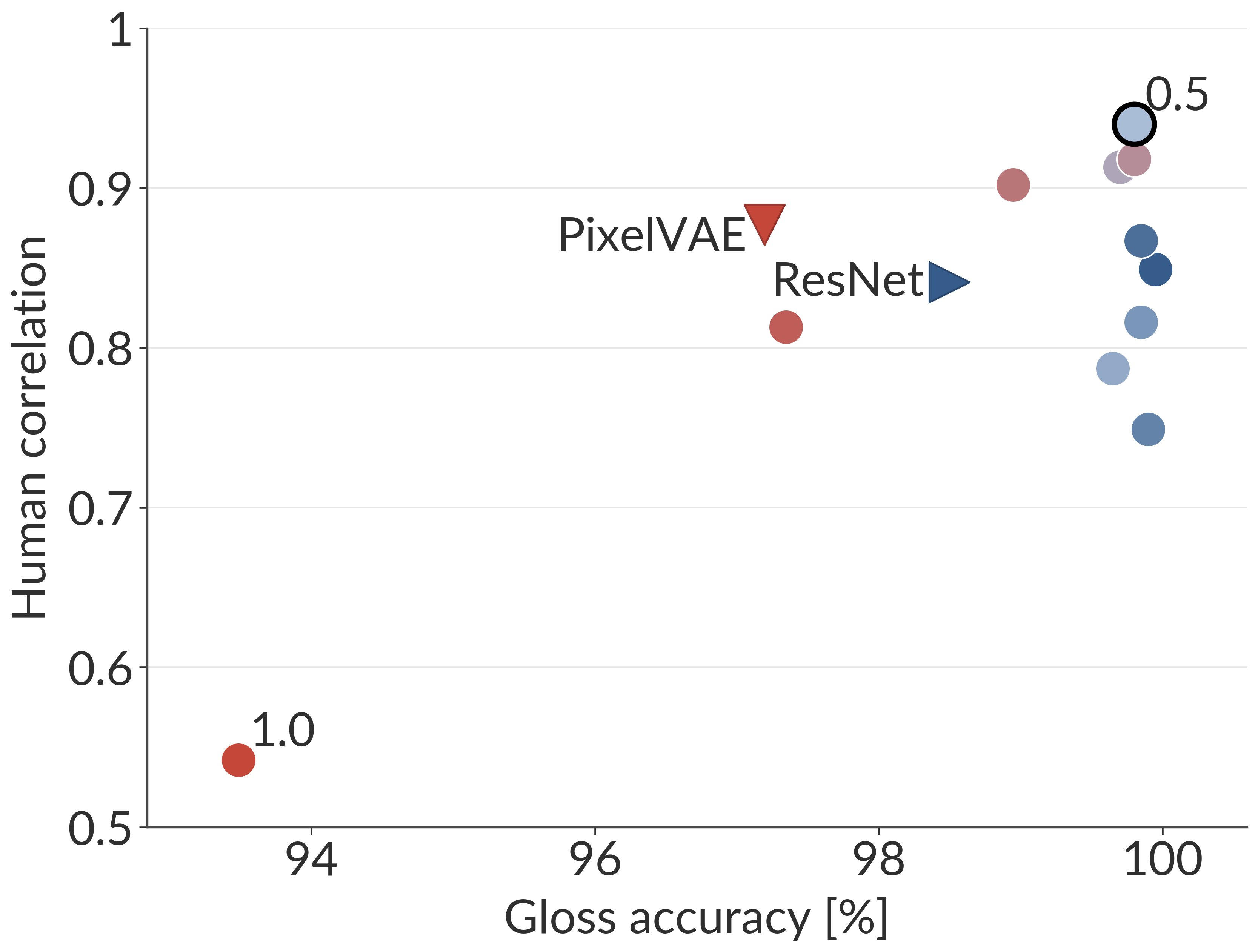}};
    \draw [anchor=north west] (0.5\linewidth, 0.50\linewidth)
    node {\includegraphics[width=0.48\linewidth]{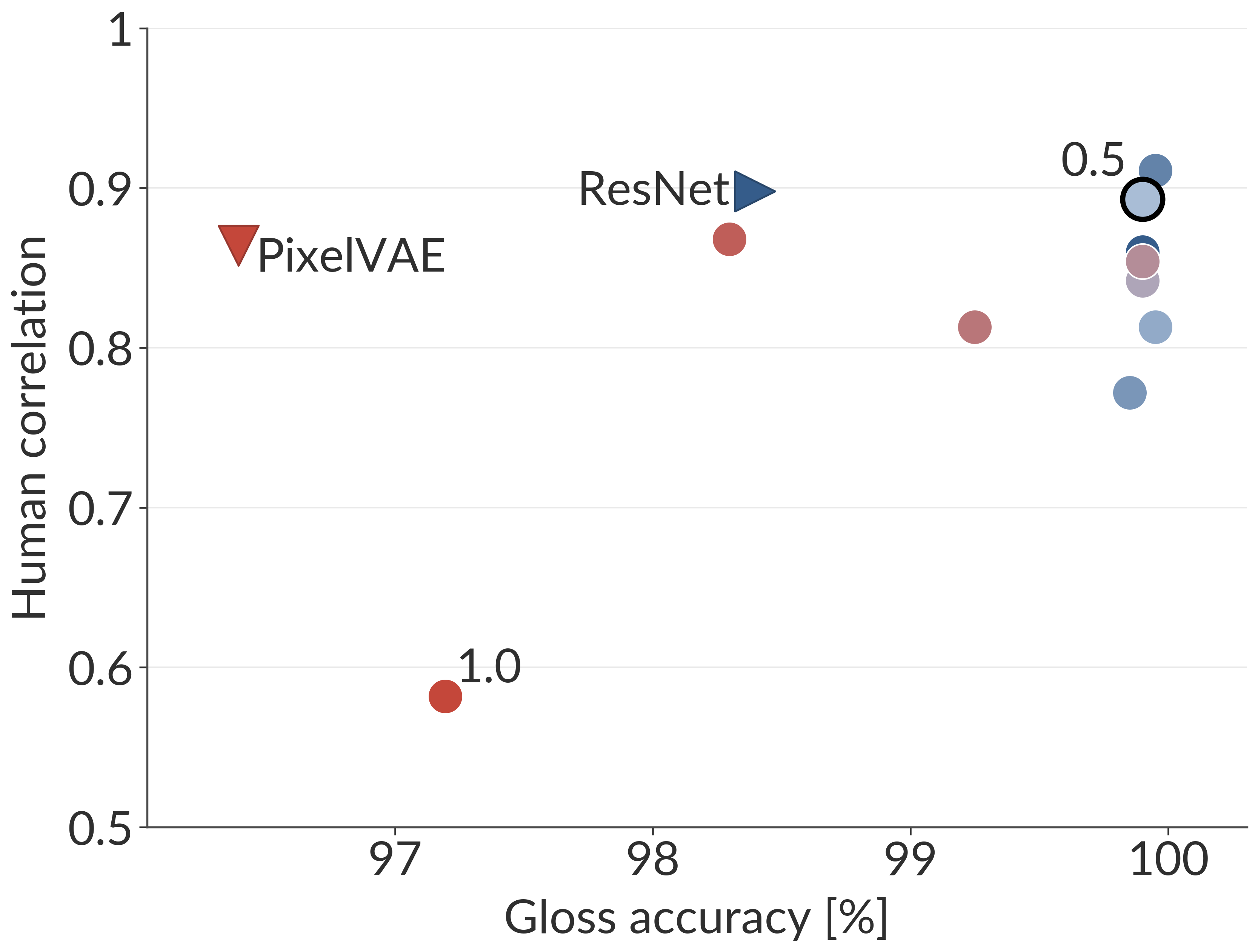}};

    % ===== Legend =====
    \node[anchor=north, inner sep=0pt] at (0.5\linewidth, 0.13\linewidth)
    {\includegraphics[width=0.8\linewidth]{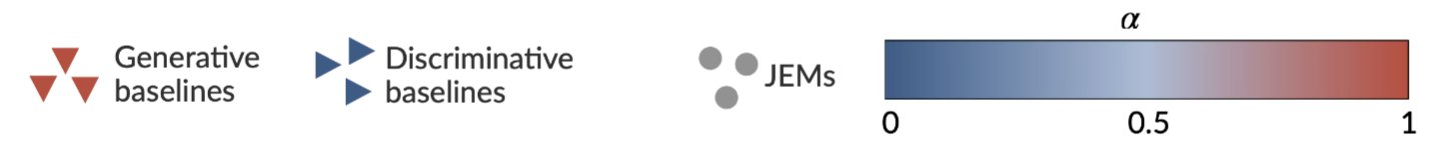}};

    % ===== Panel labels =====
    \begin{scope}
        \draw [anchor=north west, fill=white, align=left] (0.0\linewidth, 1.00\linewidth)
        node {{\bf a)} 10 dimensions};
        \draw [anchor=north west, fill=white, align=left] (0.5\linewidth, 1.00\linewidth)
        node {{\bf b)} 100 dimensions};
        \draw [anchor=north west, fill=white, align=left] (0.0\linewidth, 0.55\linewidth)
        node {{\bf c)} 500 dimensions};
        \draw [anchor=north west, fill=white, align=left] (0.5\linewidth, 0.55\linewidth)
        node {{\bf d)} 2000 dimensions};
    \end{scope}
\end{tikzpicture}
\caption{Gloss accuracy vs. human correlation in experiments with different latent sizes.}
\label{fig:gloss_dims_figure}
\end{figure}
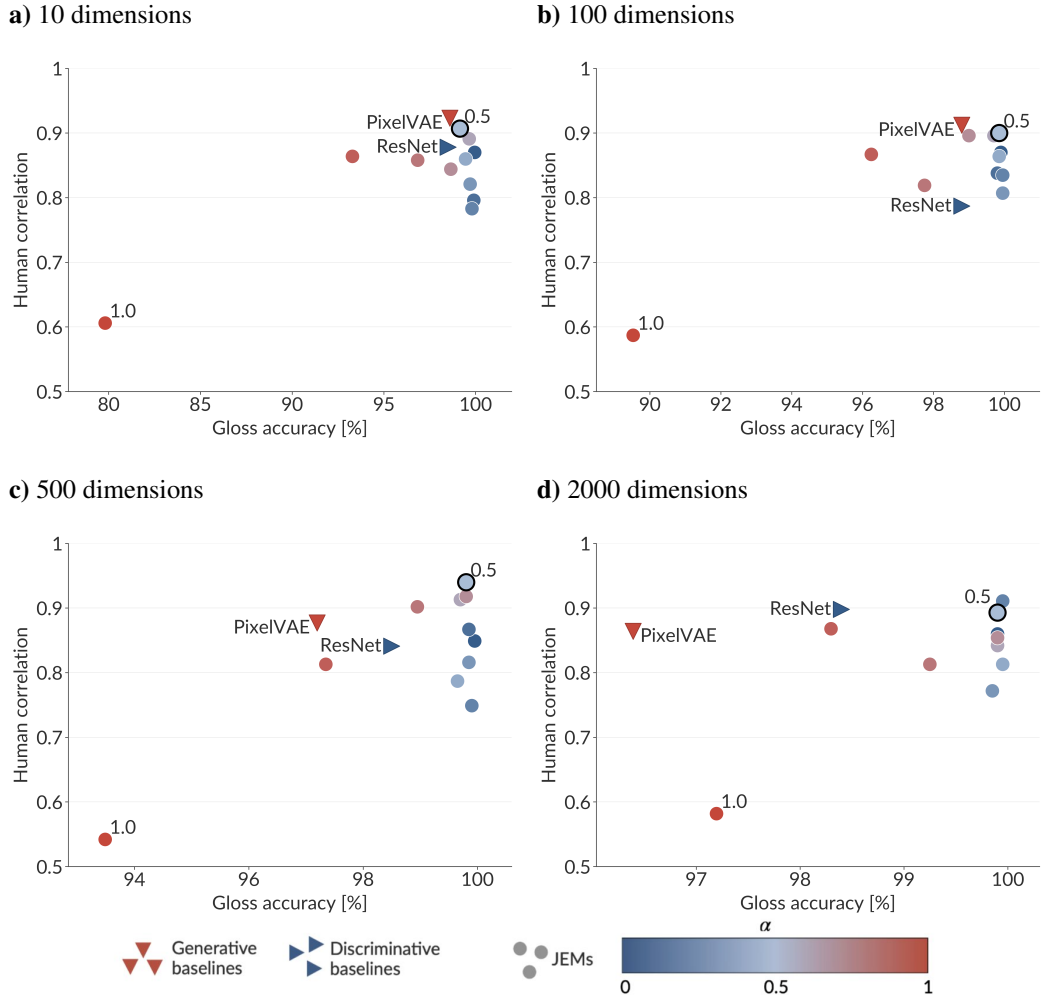

Human alignment peaks in the hybrid regime. Across latent dimensionalities, intermediate JEMs---especially around \(\alpha=0.5\)---consistently achieve the strongest or near-strongest human correlation, whereas the purely generative endpoint (\(\alpha=1\)) performs noticeably worse. Because the JEM comparisons keep the architecture fixed, this pattern provides stronger evidence than prior cross-model comparisons that the learning objective itself plays a central role in shaping human-aligned gloss representations. This conclusion is especially clear in the 500-dimensional setting, where the JEM is approximately matched in parameter count to the PixelVAE (roughly 37M versus 40M parameters). Even in this more comparable regime, the strongest human alignment is obtained by a hybrid JEM rather than by the purely generative baseline (Fig.~\ref{fig:gloss_dims_figure}).

\newpage

\begin{figure}[p]
\centering
\begin{tikzpicture}
    % ===== Row 1: latents 10 and 100 =====
    \draw [anchor=north west] (0.0\linewidth, 0.95\linewidth)
    node {\includegraphics[width=0.48\linewidth]{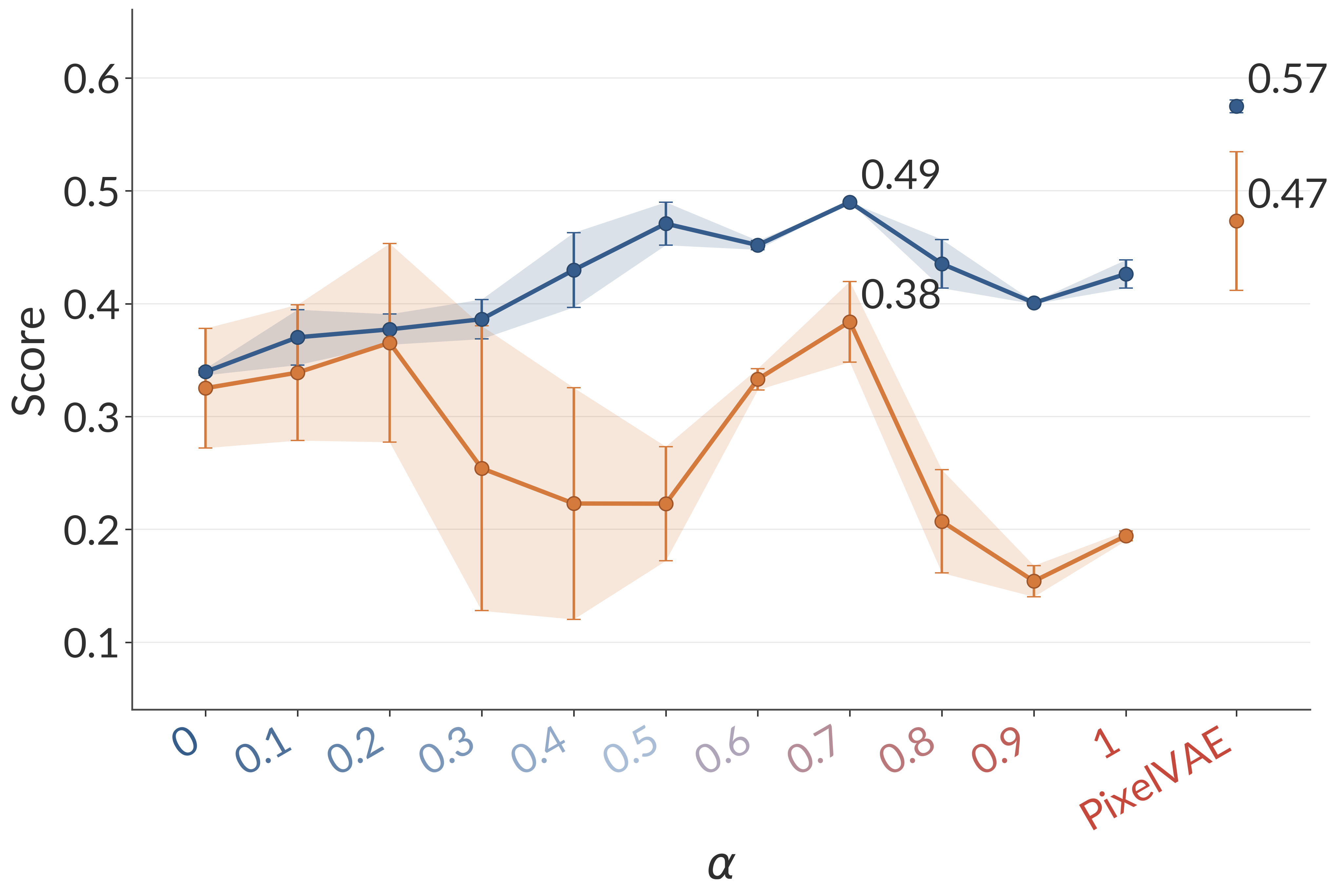}};
    \draw [anchor=north west] (0.5\linewidth, 0.95\linewidth)
    node {\includegraphics[width=0.48\linewidth]{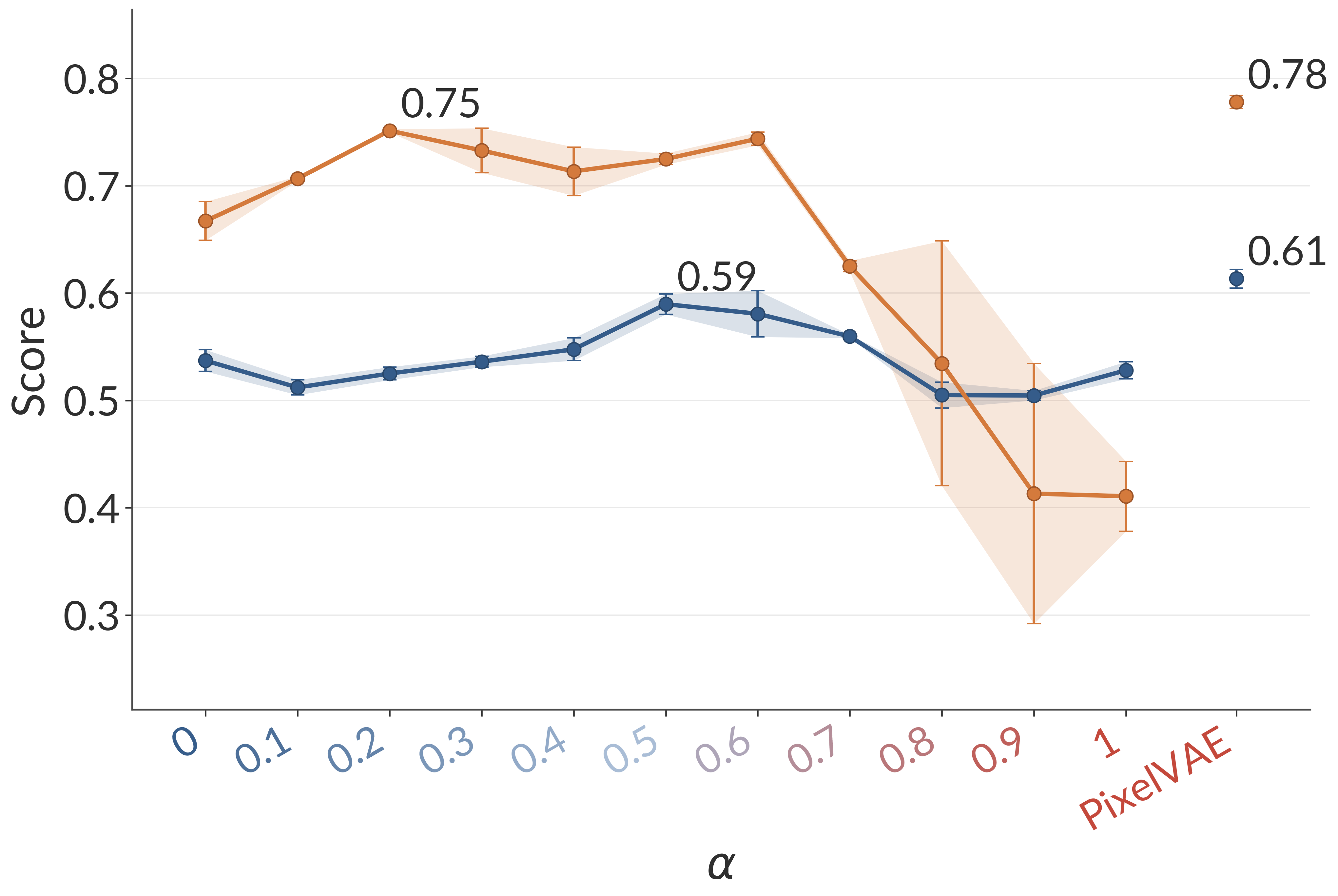}};

    % ===== Row 2: latents 500 and 2000 =====
    \draw [anchor=north west] (0.0\linewidth, 0.55\linewidth)
    node {\includegraphics[width=0.48\linewidth]{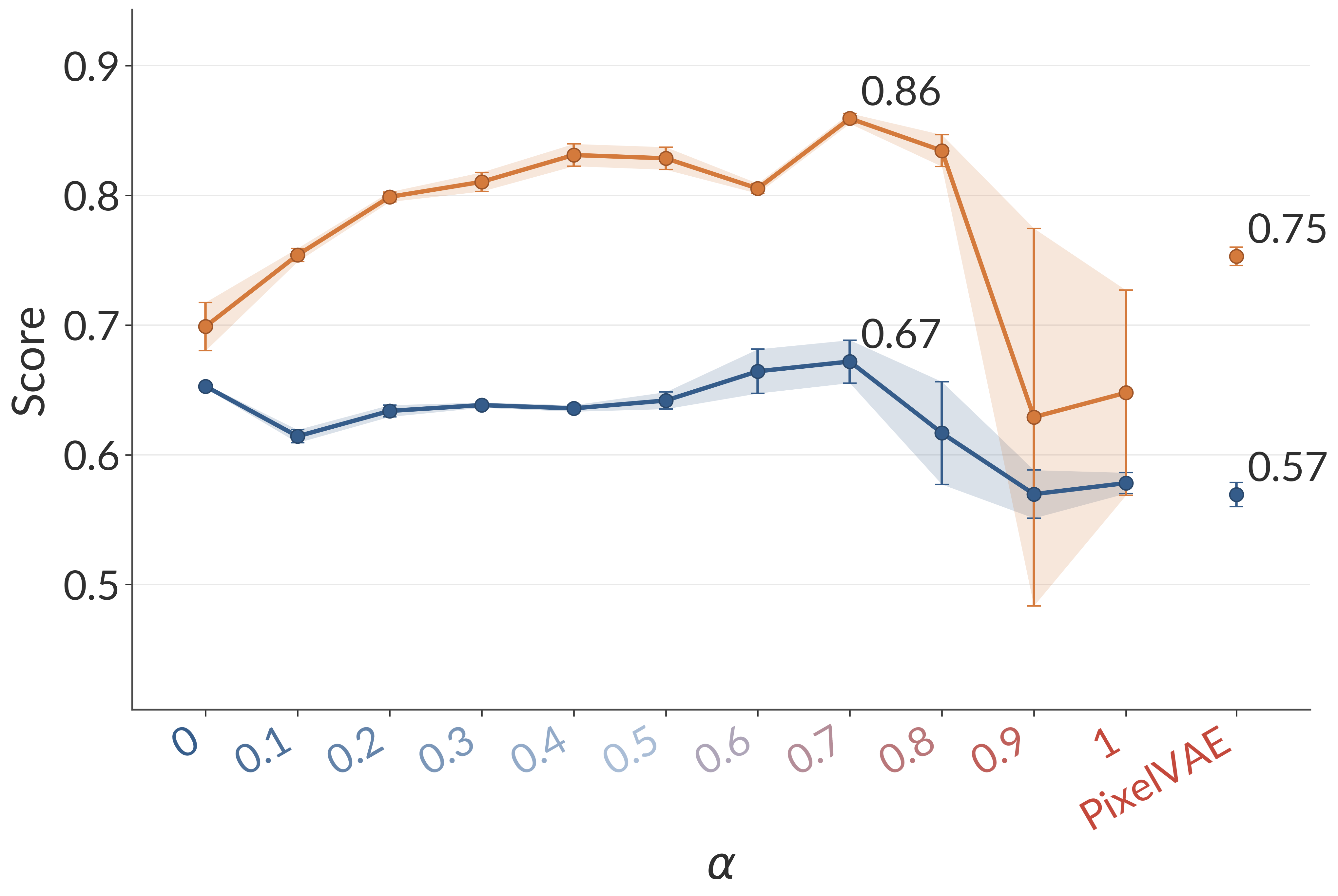}};
    \draw [anchor=north west] (0.5\linewidth, 0.55\linewidth)
    node {\includegraphics[width=0.48\linewidth]{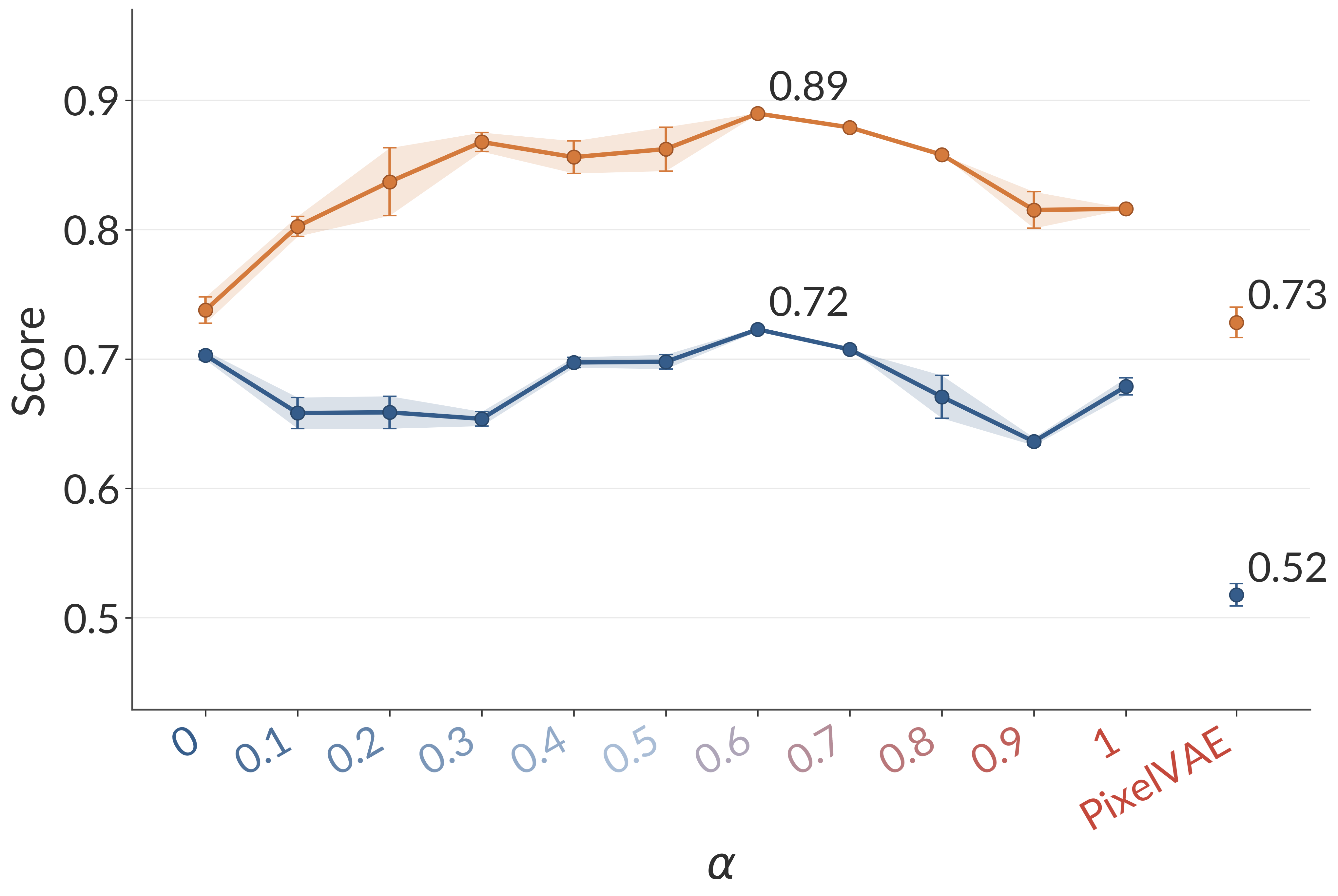}};

    % ===== Legend (relief + light field) =====
    \node[anchor=north, inner sep=0pt] at (0.5\linewidth, 0.18\linewidth)
    {\includegraphics[width=0.5\linewidth]
{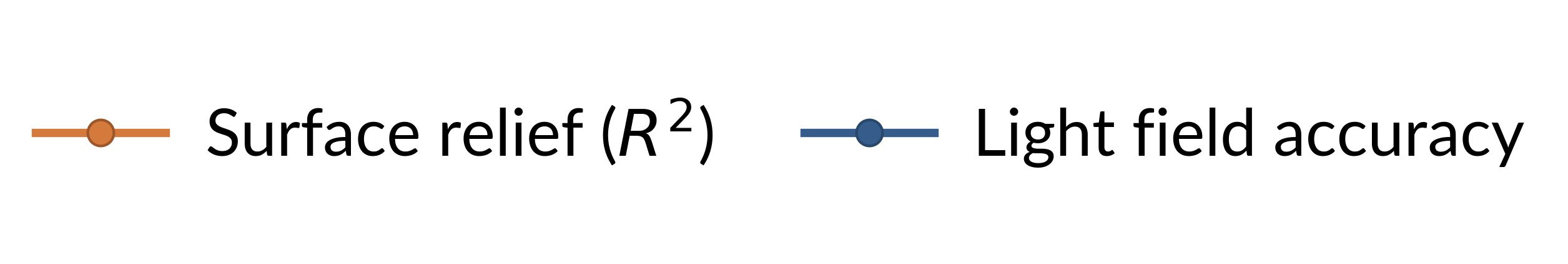}};

    % ===== Panel labels =====
    \begin{scope}
        \draw [anchor=north west, fill=white, align=left] (0.0\linewidth, 1.00\linewidth)
        node {{\bf a)} 10 dimensions};
        \draw [anchor=north west, fill=white, align=left] (0.5\linewidth, 1.00\linewidth)
        node {{\bf b)} 100 dimensions};
        \draw [anchor=north west, fill=white, align=left] (0.0\linewidth, 0.60\linewidth)
        node {{\bf c)} 500 dimensions};
        \draw [anchor=north west, fill=white, align=left] (0.5\linewidth, 0.60\linewidth)
        node {{\bf d)} 2000 dimensions};
    \end{scope}
\end{tikzpicture}
\caption{Surface relief ($R^2$) and light field accuracy across JEM $\alpha$ values for different latent dimensionalities. Shaded regions indicate the standard error of the mean (SEM) across two seeds. The disconnected point on the right of each panel shows the corresponding PixelVAE baseline.}
\label{fig:gloss_performance_relief_lightfield}
\end{figure}

PixelVAE nevertheless remains a strong baseline, reinforcing the original insight of Storrs et al.~\cite{storrs2021unsupervised}: that generative learning captures important structure for gloss perception. However, because the original PixelVAE--ResNet comparison also changed model family and parameterization, it did not isolate the contribution of the learning objective itself. Our results refine that conclusion: once architecture is controlled, the best human alignment does not arise from purely generative training, but from an intermediate balance between generative and discriminative pressure. This hybrid advantage is not limited to human correlation. The same intermediate JEMs also yield the strongest downstream decoding of scene properties, including surface relief and light field, which indicates that the hybrid regime preserves the latent physical variables underlying gloss more effectively than either endpoint of the continuum (Fig.~\ref{fig:gloss_performance_relief_lightfield}).

\begin{figure}[p]
\centering
\begin{tikzpicture}
    % ===== Row: gloss-human correlation and gloss accuracy =====
    \draw [anchor=north west] (0.0\linewidth, 0.55\linewidth)
    node {\includegraphics[width=0.48\linewidth]{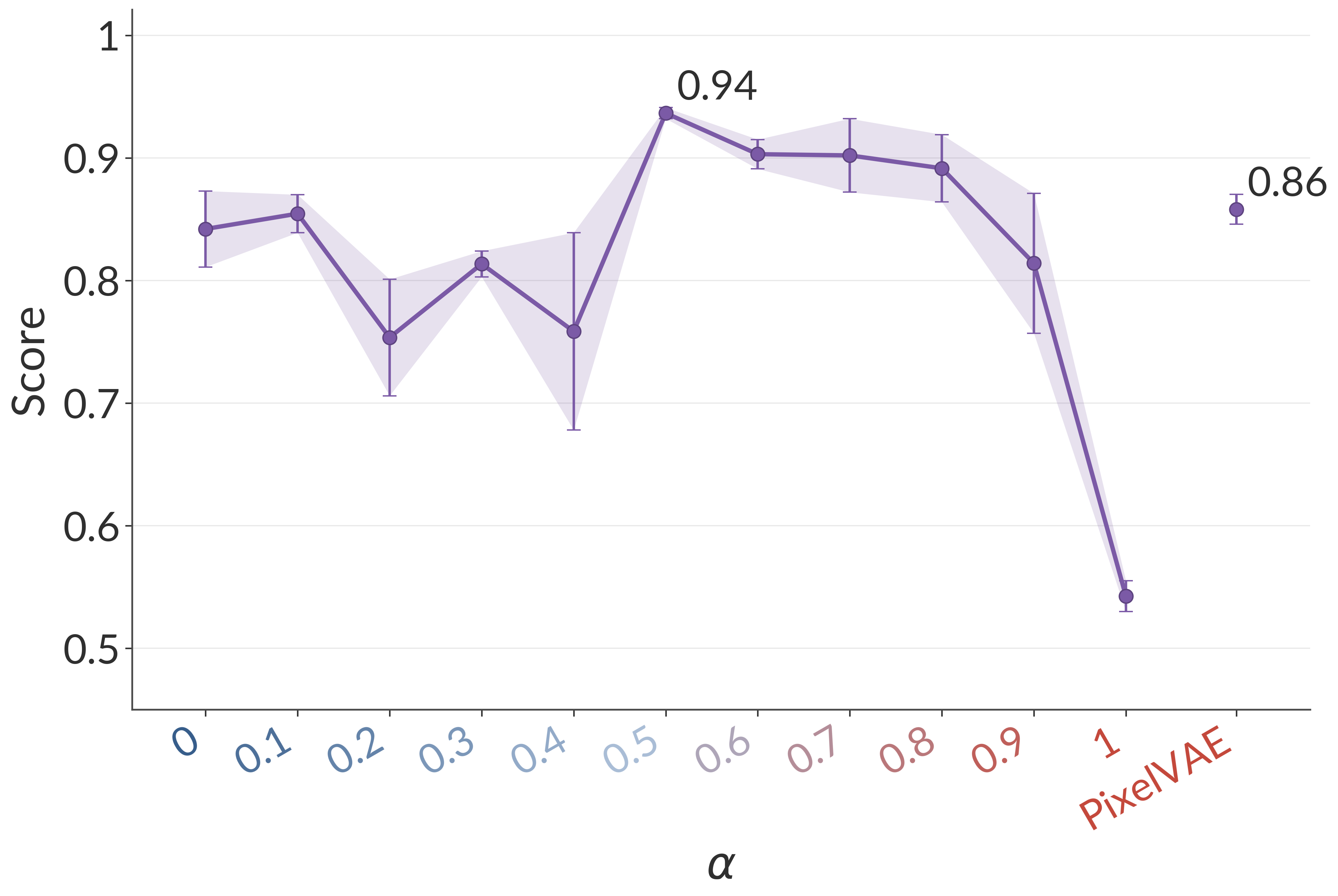}};
    \draw [anchor=north west] (0.5\linewidth, 0.55\linewidth)
    node {\includegraphics[width=0.48\linewidth]{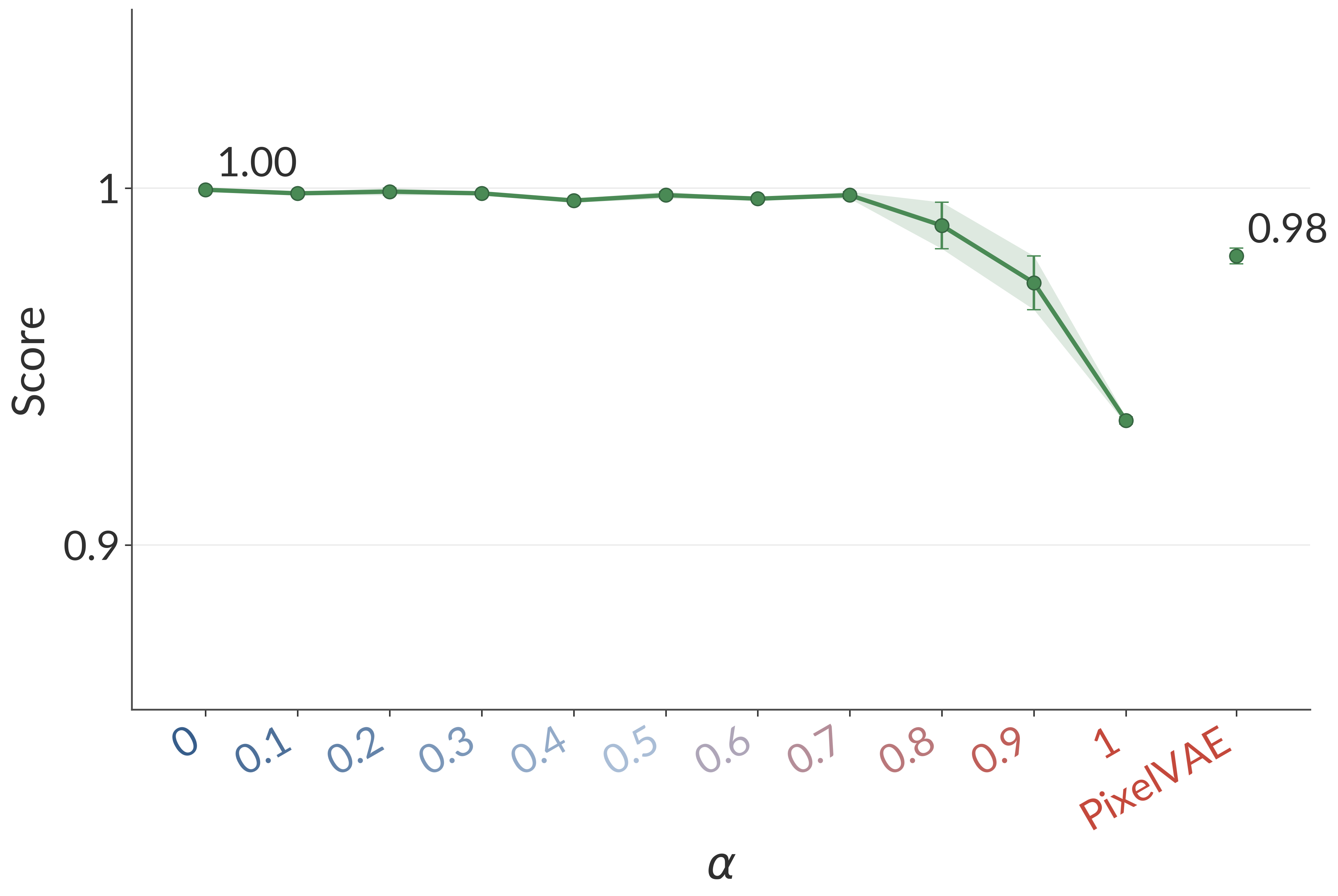}};

    % ===== Legend (centered) =====
    \node[anchor=north, inner sep=0pt] at (0.5\linewidth, 0.18\linewidth)
    {\includegraphics[width=0.5\linewidth]{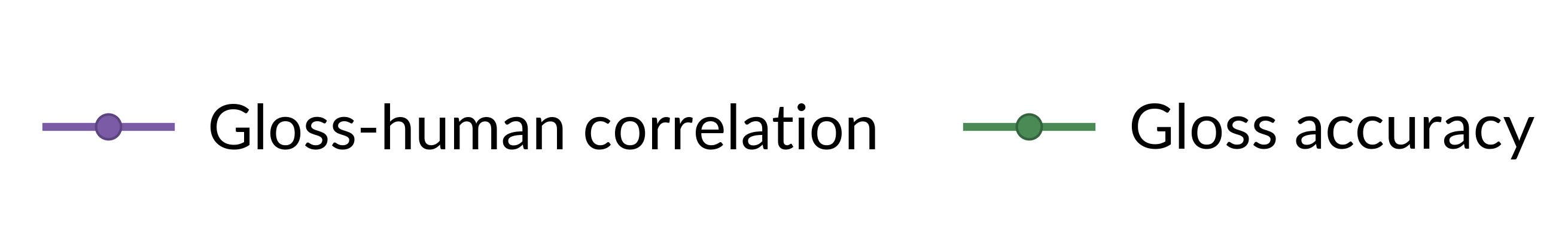}};

    % ===== Panel labels =====
    \begin{scope}
        \draw [anchor=north west, fill=white, align=left] (0.0\linewidth, 0.60\linewidth)
        node {{\bf a)} Human correlation};
        \draw [anchor=north west, fill=white, align=left] (0.5\linewidth, 0.60\linewidth)
        node {{\bf b)} Gloss accuracy};
    \end{scope}
\end{tikzpicture}
\caption{Gloss-human correlation and gloss accuracy across JEM $\alpha$ values at 500 latent dimensions. Shaded regions indicate the standard error of the mean (SEM) across two seeds. The disconnected point on the right of each panel shows the corresponding PixelVAE baseline.}
\label{fig:gloss_performance_correlation_accuracy}
\end{figure}

\begin{figure}[h]
    \centering
    
    \includegraphics[width=1.0\textwidth]{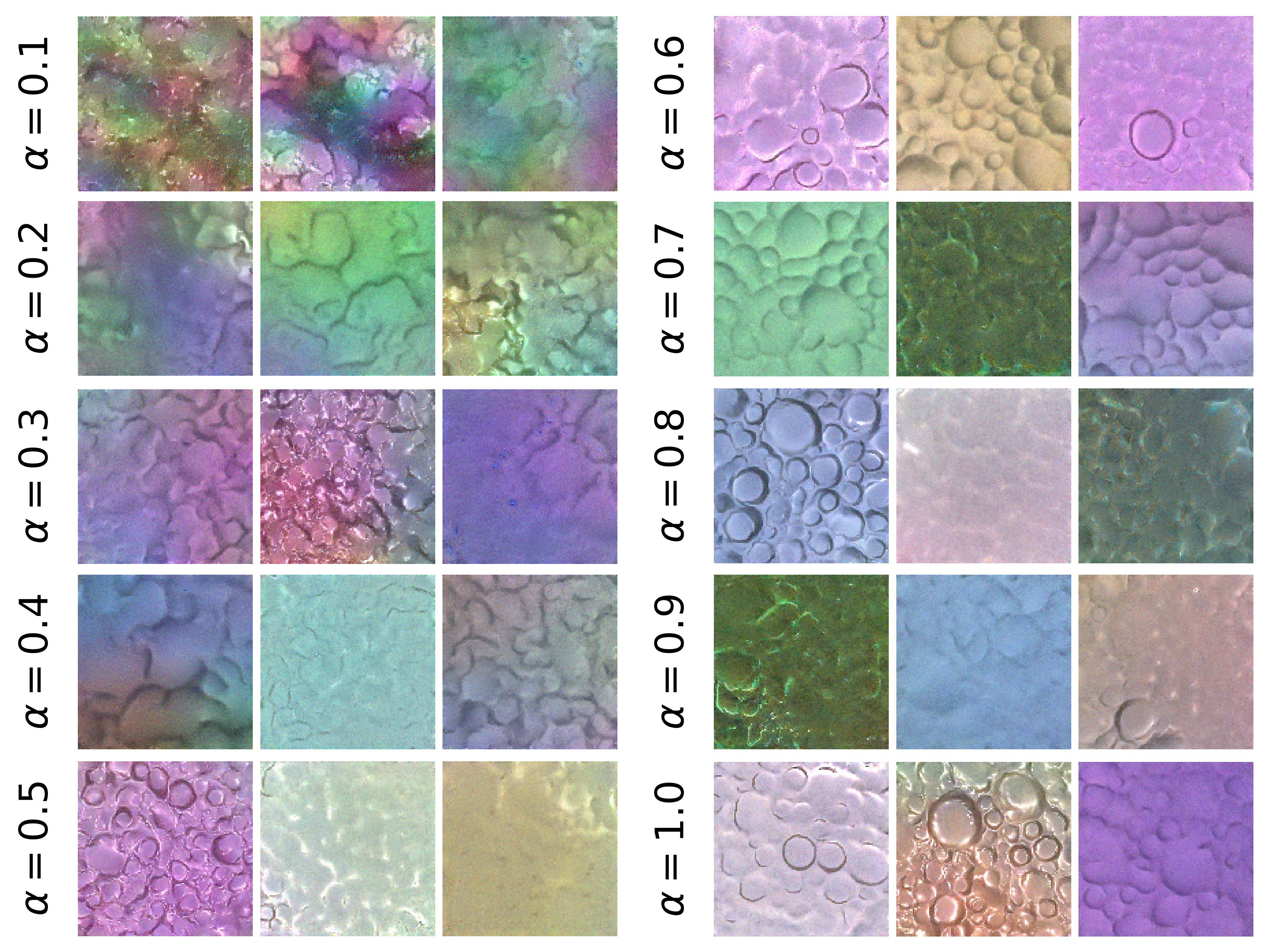}
    \caption{Qualitative gloss generations as a function of \(\alpha\).}
    \label{fig:gloss_gens}
\end{figure}

For completeness, we also report the two-seed results for the 500-dimensional setting (Fig.~\ref{fig:gloss_performance_correlation_accuracy}) together with qualitative generations across \(\alpha\) (Fig.~\ref{fig:gloss_gens}). While increasing \(\alpha\) generally improves the visual plausibility of the generated surfaces, the best alignment with human gloss judgments is achieved in the hybrid regime rather than at the purely generative endpoint.

% =====================================================
% Figure 1: latents 10, 100, 500, 2000 with relief + light field legend
% =====================================================

% =====================================================
% Figure 2: latent 500 — gloss-human correlation and gloss accuracy
% =====================================================

\newpage

\newpage
\section{CIFAR10H}\label{supp:cifar10h}

CIFAR-10H extends standard object classification to \textbf{visual categorization under human uncertainty}. It provides full distributions of human labels for the entire 10,000-image CIFAR-10 test set, covering the same ten object categories and based on 511{,}000+ crowdsourced human judgments, corresponding to 51.1 judgments per image on average (range: 47--63). Rather than reducing each image to a single hard label, CIFAR-10H captures the full distribution of human categorizations, allowing evaluation of whether a model reflects not only the most likely class but also whether the model's uncertainty across categories is aligned with human categorization behavior.

\subsection{Models and training protocol}\label{supp:cifar10h_training}

In the original CIFAR-10H paper, the authors considered an explicit human-supervision setting in which pretrained models were fine-tuned on CIFAR-10H soft labels. Our goal here is different: we ask whether alignment with human uncertainty arises naturally from the representations learned by JEMs, rather than being directly imposed through supervision on CIFAR-10H. Therefore, all models in our study were trained on standard CIFAR-10~\cite{krizhevsky2009learning} and evaluated on CIFAR-10H only at test time.

We trained three JEM instances with different seeds for each value of \(\alpha\), using the same general procedure described in Appendix~\ref{supp:ebm_training}. For the discriminative baselines, we trained VGG, ResNet, and ResNeXt models using the \texttt{pytorch\_image\_classification} codebase, matching the repository used by Peterson et al.~\cite{peterson2019human}. Additionally, we included a ConvNeXt-Tiny baseline trained using \texttt{timm}, which was not part of the original CIFAR-10H study. For model selection, we held out 5\% of the CIFAR-10 training set as a validation split. Moreover, we used random cropping and random horizontal flipping as the only data augmentations. We did not use label smoothing or any fine-tuning on CIFAR-10H in order to keep the comparison with the JEMs as fair as possible. 

Across the discriminative baselines, we used Adam with learning rate \(10^{-3}\). For JEMs, we used a smaller learning rate of \(10^{-4}\), since larger values tended to destabilize the hybrid objective. Aside from this difference, baseline models were trained until convergence on the same CIFAR-10 setup used for the JEM experiments and were then evaluated on the CIFAR-10H test set using the metrics described below. 

\subsection{Evaluation metrics}\label{supp:cifar10h_metrics}

Following Peterson et al.~\cite{peterson2019human}, we use \textbf{cross-entropy} with respect to the human label distribution as our primary metric on CIFAR-10H. When the target is a human soft-label distribution \(p_{\mathrm{hum}}(y \mid x)\), cross-entropy provides a principled measure of how well the model prediction \(p_{\vth}(y \mid x)\) matches human categorization behavior under uncertainty. It is also more informative than top-1 accuracy in this setting, since accuracy alone does not capture how probability mass is distributed across incorrect classes.

We additionally report \textbf{KL divergence} from the human distribution to the model distribution as a complementary diagnostic of distributional mismatch. Because cross-entropy and KL divergence differ only by the entropy of the human label distribution, the two metrics are closely related. We use cross-entropy as the primary metric for consistency with Peterson et al. ~\cite{peterson2019human}, and KL divergence as a secondary summary of alignment with human uncertainty.

\begin{figure}[h]
\centering

\begin{subfigure}{0.49\linewidth}
    \centering
    \caption{CIFAR-10 top-1 accuracy}
    \label{fig:cifar10h-acc}
    \includegraphics[width=\linewidth]{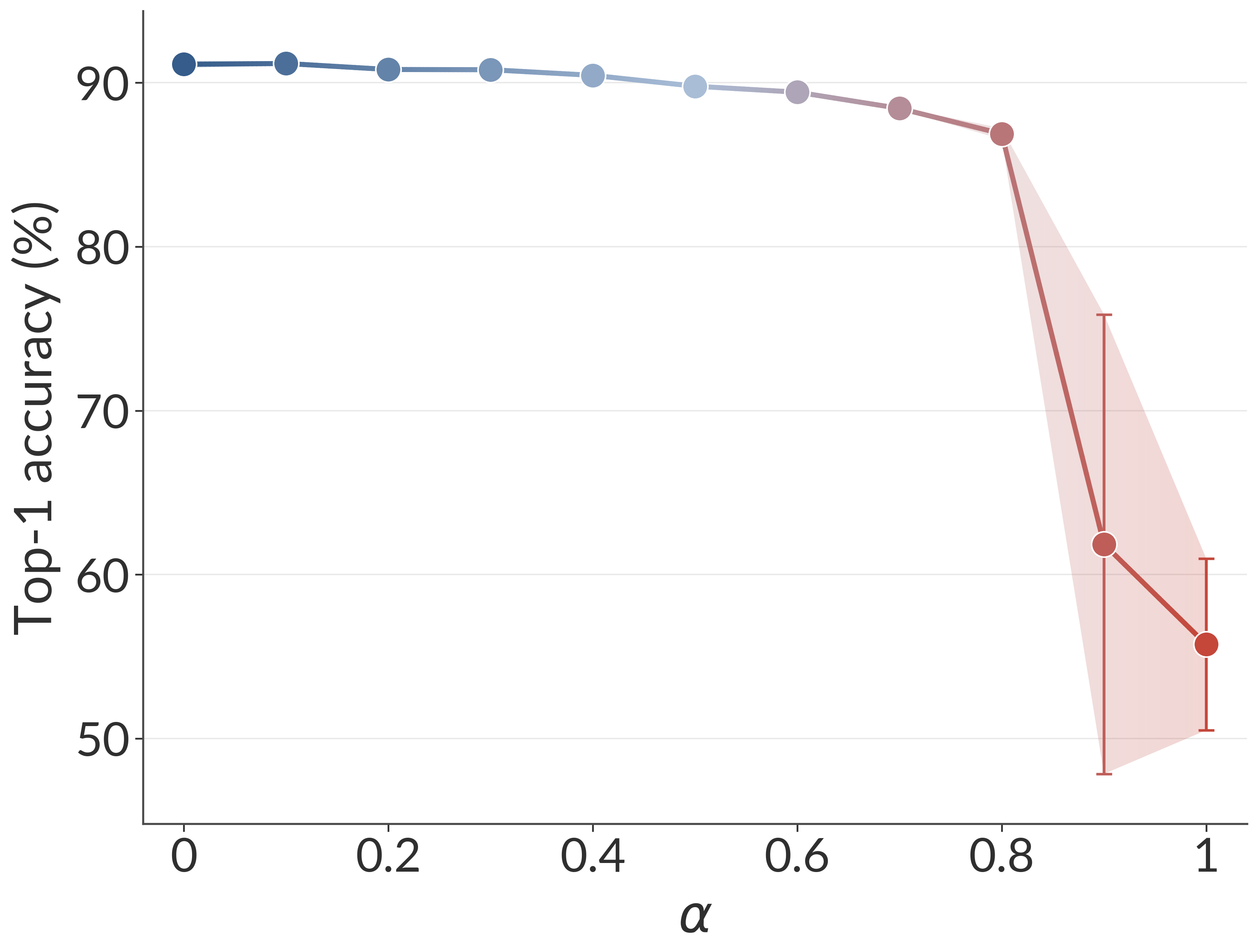}
\end{subfigure}
\hfill
\begin{subfigure}{0.49\linewidth}
    \centering
    \caption{CIFAR-10H cross-entropy}
    \label{fig:cifar10h-ce}
    \includegraphics[width=\linewidth]{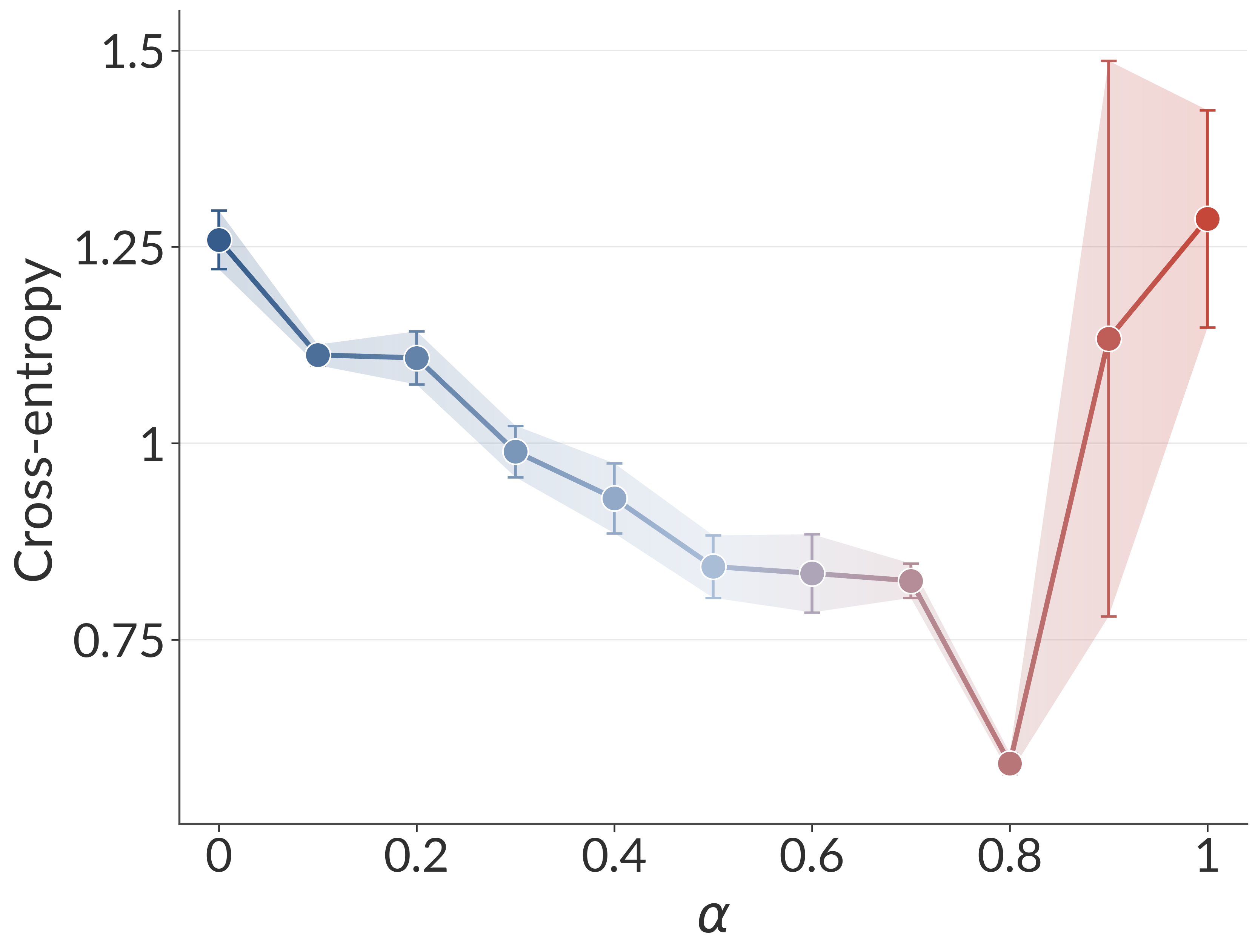}
\end{subfigure}
\hfill
\begin{subfigure}{0.49\linewidth}
    \centering
    \caption{CIFAR-10H KL divergence}
    \label{fig:cifar10h-kl}
    \includegraphics[width=\linewidth]{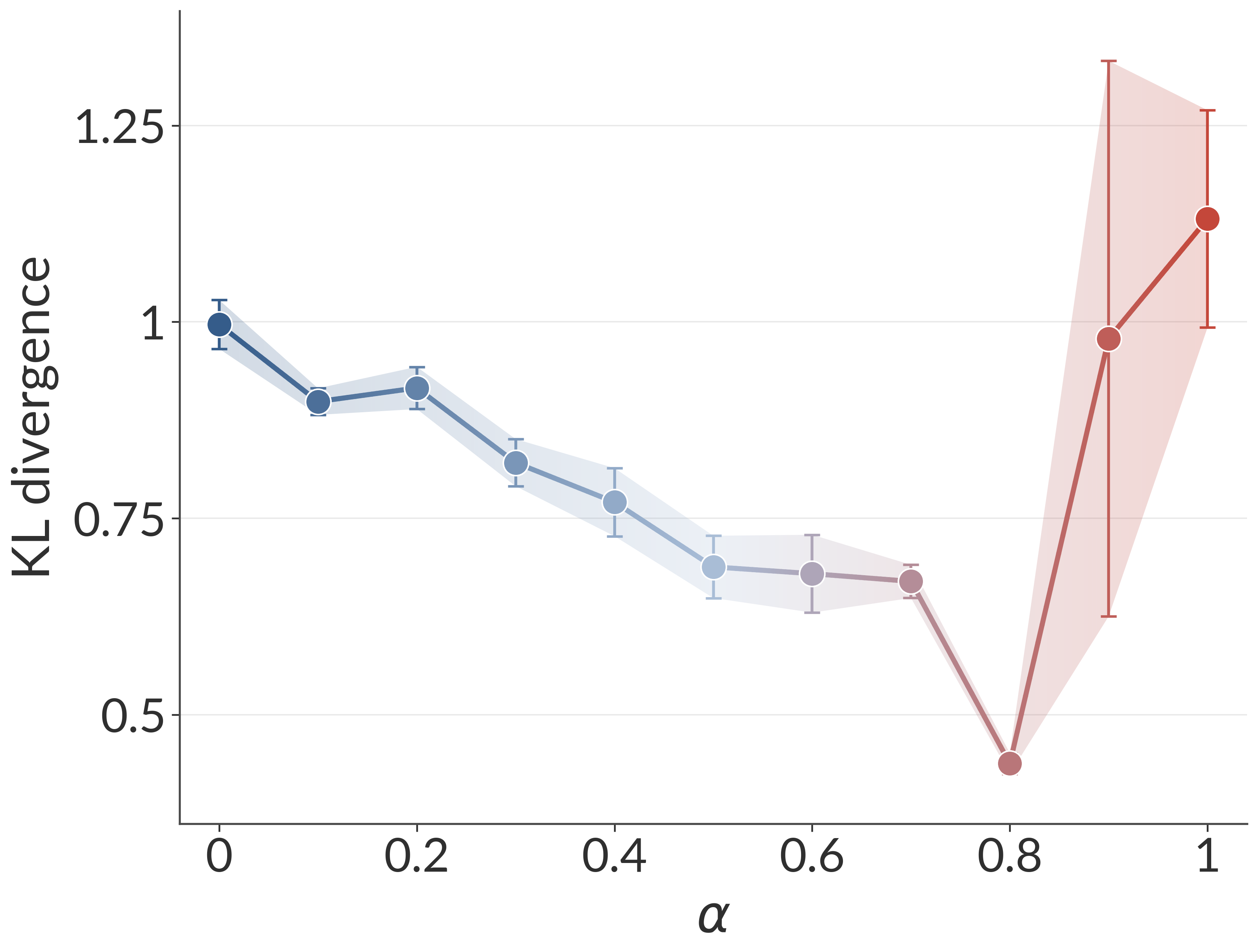}
\end{subfigure}

\caption{CIFAR-10 and CIFAR-10H evaluation across $\alpha$.}
\label{fig:cifar10h}
\caption{
CIFAR-10 classification performance and alignment with CIFAR-10H human soft labels as a function of the generative--discriminative trade-off parameter \(\alpha\). Each point shows the mean across three seeds, and error bars denote the standard error of the mean (SEM).
}
\label{fig:cifar10h-ce-kl}
\end{figure}

For completeness, we also report \textbf{top-1 accuracy} on the same 10,000 test images. This allows us to distinguish conventional classification performance from alignment with human uncertainty: a model may achieve high top-1 accuracy while still assigning probability mass across categories in a way that differs substantially from human judgments. In the corresponding plots (Fig.~\ref{fig:cifar10h}), each point shows the mean across three seeds, and error bars denote standard error of the mean.

\newpage

%\begin{figure}[h]
%    \centering
%    \includegraphics[width=0.57\textwidth]{figures/cifarh/samples.png}
%    \caption{Your caption here.}
%    \label{fig:your_label}
%\end{figure}
\newpage

\section{Generalization benchmark}\label{supp:bench_generalization}

\begin{figure}[ht]
    \centering
    \includegraphics[trim={30pt 0pt 0pt 0pt}, clip, width=.9\linewidth]{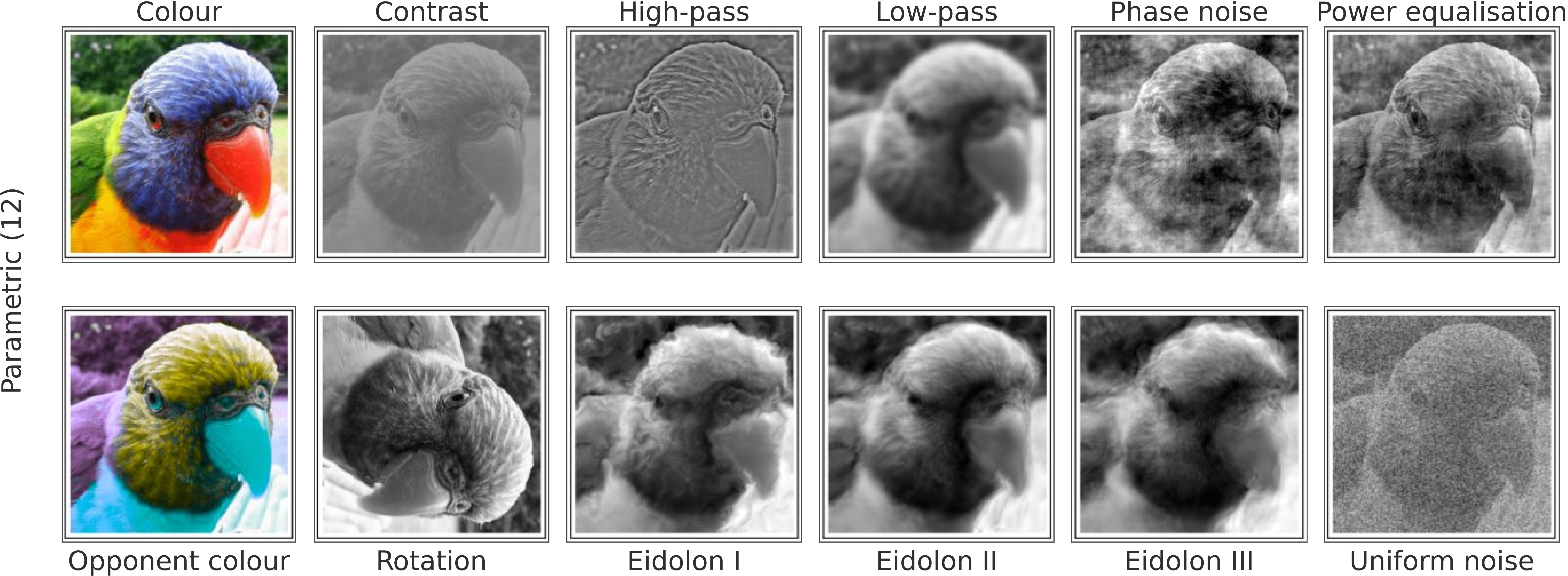}
    \caption{Parametric transformations used in the Model-vs-Human benchmark: colour, contrast, frequency filtering, phase noise, power equalisation, opponent colour, rotation, eidolons, and uniform noise.}
    \label{fig:parametric_transforms}
\end{figure}

\begin{figure}[ht]
    \centering
    \includegraphics[trim={30pt 0pt 0pt 0pt}, clip, width=\linewidth]{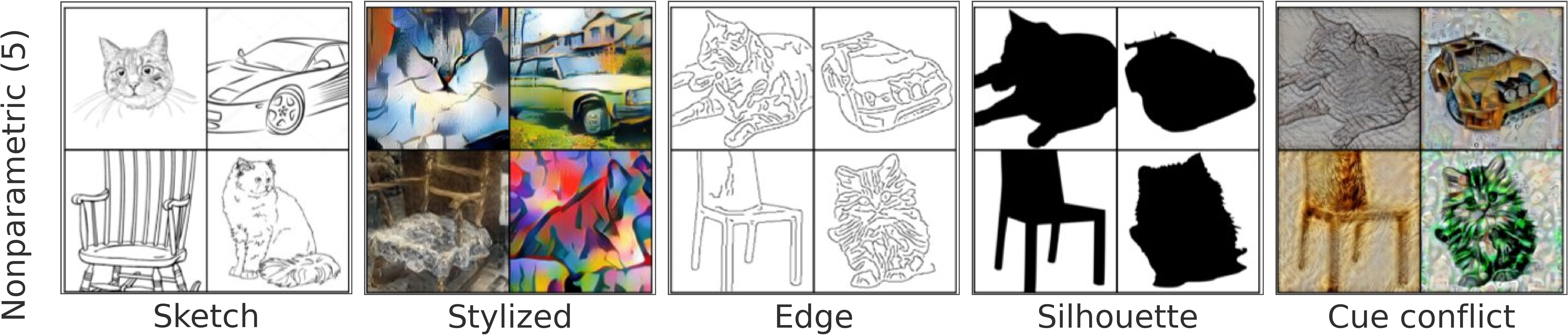}
    \caption{Nonparametric transformations used in the Model-vs-Human benchmark: sketch, stylized images, edge maps, silhouettes, and cue-conflict images.}
    \label{fig:nonparametric_transforms}
\end{figure}

To evaluate model generalization, we use the \href{https://github.com/bethgelab/model-vs-human}{Model-vs-Human} benchmark suite introduced by ~\citet{geirhos2021partial} . This suite consists of 17 out-of-distribution (OOD) datasets designed for ImageNet-trained models and evaluated on 16 categories derived from the same dataset. In addition, the benchmark provides human psychophysical data collected under controlled laboratory conditions. We describe further each dataset in Tables ~\ref{noise_gen_table},~\ref{shape_texture_table}, and show examples in  Figures ~\ref{fig:parametric_transforms},~\ref{fig:nonparametric_transforms}.

The human dataset comprises 85,120 trials with 90 participants, who were asked to classify the transformed images. ~\citet{geirhos2021partial} also evaluated a broad range of state-of-the-art ImageNet models on these benchmarks, including standard supervised classifiers, self-supervised models, vision transformers, adversarially trained models, and models trained on substantially larger datasets. This makes the benchmark effective for assessing not only OOD accuracy, but also the extent to which different models generalize in a human-like way.

\begin{table}[ht]
  \caption{\textbf{Noise generalization benchmarks.} Parametric image manipulations from ~\citet{geirhos2018generalisation}.}
  \label{tab:model_vs_human_parametric}
  \centering
  \small
  \begin{tabular}{ll}
    \toprule
    \textbf{Benchmark} & \textbf{Levels / description} \\
    \midrule
    Greyscale & Colour vs. greyscale \\
    Contrast & 100, 50, 30, 15, 10, 5, 3, 1\% contrast \\
    High-pass & $\sigma=\infty$, 3.0, 1.5, 1.0, 0.7, 0.55, 0.45, 0.4 \\
    Low-pass & $\sigma=0$, 1, 3, 5, 7, 10, 15, 40 \\
    Phase scrambling & $0^\circ$, $30^\circ$, $60^\circ$, $90^\circ$, $120^\circ$, $150^\circ$, $180^\circ$ phase noise \\
    Power equalisation & Original vs. power-spectrum-equalised images \\
    False colour & True-colour vs. opponent-colour images \\
    Rotation & $0^\circ$, $90^\circ$, $180^\circ$, $270^\circ$ rotations \\
    Eidolon I & Coherence 10; reach varies with $\log_2$ reach 0--7 \\
    Eidolon II & Coherence 3 \\
    Eidolon III & Coherence 0 \\
    Uniform noise & Width 0.0, .03, .05, .1, .2, .35, .6, .9 \\
    \bottomrule
  \end{tabular}
  \label{noise_gen_table}
\end{table}

\begin{table}[ht]
  \caption{\textbf{Texture--shape benchmarks.} Nonparametric datasets from~\citet{geirhos2018imagenet} and ~\citet{wang2019learningrobustglobalrepresentations}.}
  \label{tab:model_vs_human_nonparametric}
  \centering
  \small
  \begin{tabular}{ll}
    \toprule
    \textbf{Benchmark} & \textbf{Levels / description} \\
    \midrule
    Original & Clean reference photographs \\
    Greyscale & Desaturated originals \\
    Edge & Canny-edge line drawings \\
    Silhouette & Black-on-white object silhouettes \\
    Texture & Texture-only patches \\
    Cue conflict & Stylized images with conflicting shape and texture cues \\
    Stylized & Stylized-ImageNet images \\
    Sketch & ImageNet-Sketch hand-drawn sketches \\
    \bottomrule
  \end{tabular}
  \label{shape_texture_table}
\end{table}

%\iffalse
%\begin{figure}[h]
%    \centering
%    \includegraphics[width=1.0\textwidth]{figures/all_transformations.png}
%    \caption{Example of transformations used in model-vs-human benchmark from Geirhos et al.}
%    \label{fig:gloss_dataset}
%\end{figure}
%\fi

\paragraph{Evaluation metrics}

\begin{figure}[!htbp]
    \centering
    \begin{tikzpicture}
        \node [anchor=south west, inner sep=0pt] (img) at (0,0)
            {\includegraphics[width=0.46\linewidth]{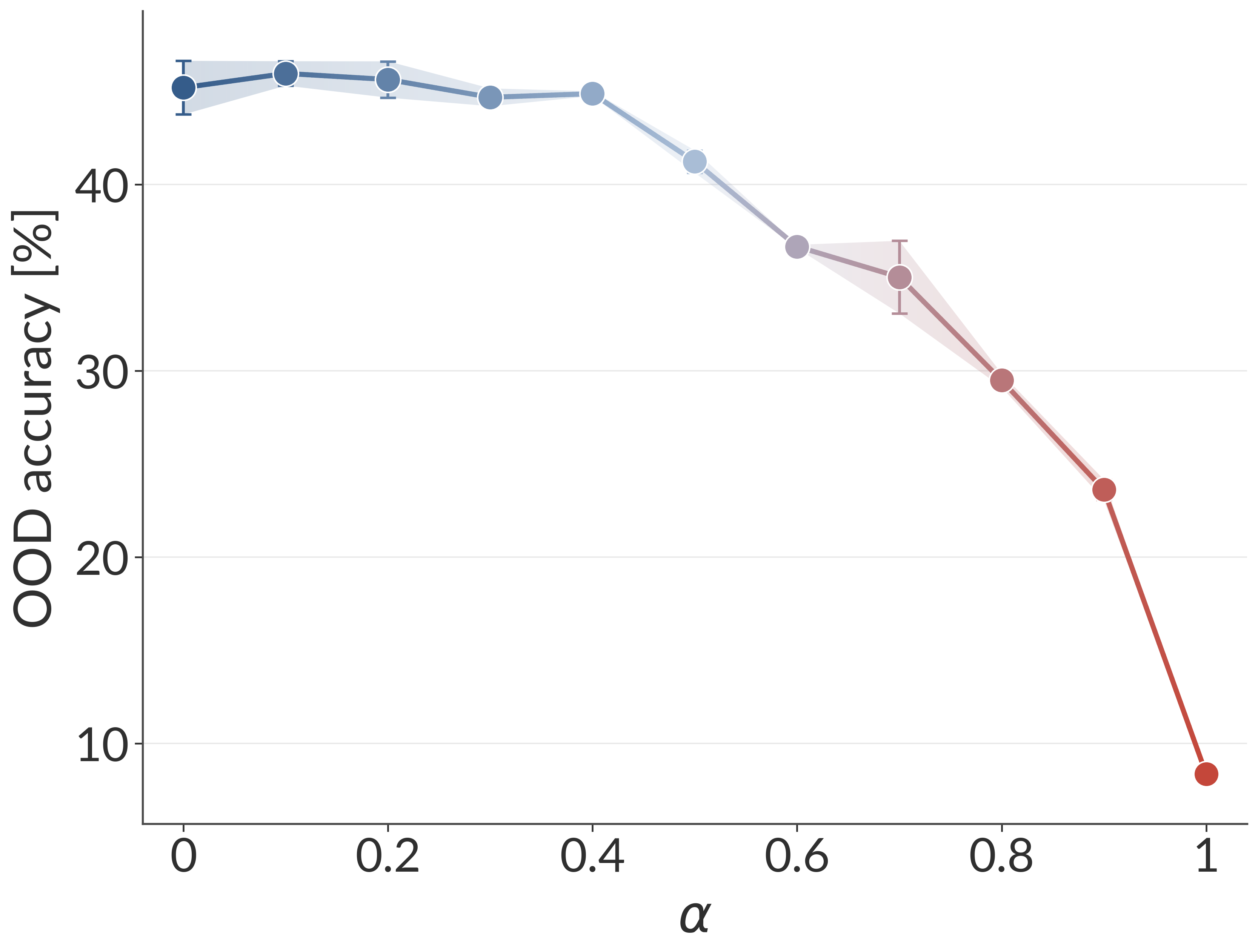}%
             \includegraphics[width=0.46\linewidth]{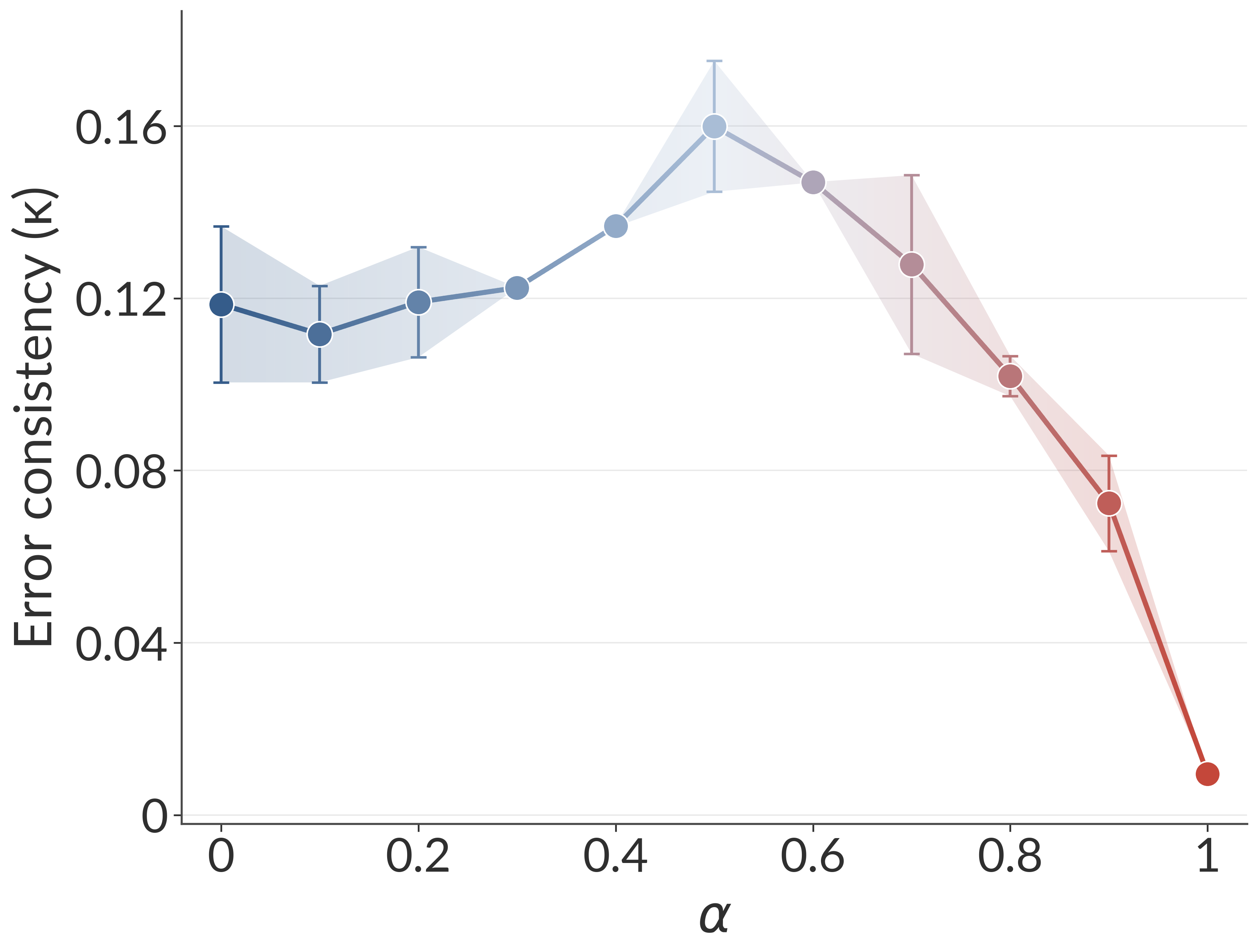}};

    \end{tikzpicture}
    \caption{OOD accuracy and error consistency metrics, across JEM $\alpha$ values. Shaded regions indicate the standard error of the mean (SEM) across two seeds.}
    \label{fig:ood_ec_seeds}
\end{figure}

\begin{figure}[h]
    \centering
    \makebox[\textwidth][c]{%
        \includegraphics[trim={20 220pt 0 20},clip,width=1.0\textwidth]{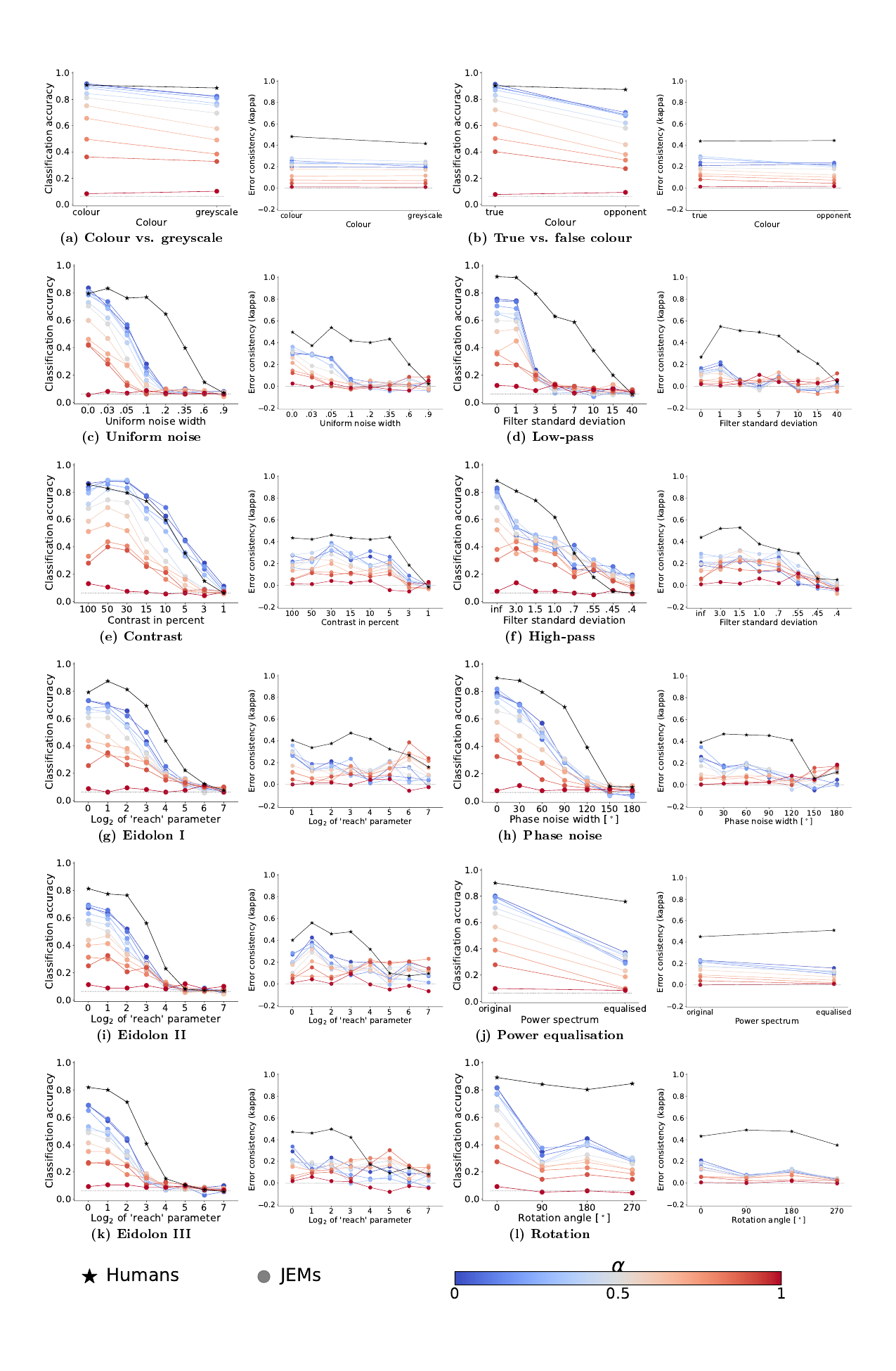}%
    }
    \caption{OOD accuracy and error consistency for each benchmark individually.}
    \label{fig:app_generalization_repor}
\end{figure}

We evaluated 11 ImageNet-trained JEMs, corresponding to values of \(\alpha\) ranging from 0 to 1 in increments of 0.1, i.e., from purely discriminative to purely generative training. Following ~\citet{geirhos2021partial}, we assessed these models on the Model-vs-Human benchmark suite and report two summary metrics: OOD accuracy and error consistency (Figs.~\ref{fig:ood_ec_seeds} \& \ref{fig:app_generalization_repor}). 

\paragraph{OOD accuracy.}
For each benchmark condition, we compute the model's classification accuracy on the corresponding out-of-distribution images. Because the model is not trained on these image manipulations, this value measures OOD accuracy. In the main results, we report OOD accuracy averaged across benchmark conditions and datasets.

\paragraph{Error consistency.}
In addition to accuracy, we measure whether the model tends to make mistakes on the same images as human observers.  We use error consistency, a chance-corrected agreement measure closely related to Cohen's \(\kappa\). This correction is important because two decision makers with high accuracy can show high raw agreement simply because both classify most images correctly. Error consistency, instead, asks whether the observed overlap in errors exceeds what would be expected from two independent decision makers with matched accuracies.

Formally, let \(D\) denote the set of datasets, \(C_d\) the set of conditions for dataset \(d\), \(H_d\) the set of human participants evaluated on dataset \(d\), and \(S_{d,c}\) the set of stimuli in condition \(c\) of dataset \(d\). Let \(b_{h,m}(s)=1\) indicate when human observer \(h\) and model \(m\) give the same correctness outcome on stimulus \(s\) (that is, both are correct or both are incorrect), and \(b_{h,m}(s)=0\) otherwise. Let \(\hat{o}_{h,m}(S_{d,c})\) denote the expected agreement under independent binomial decision makers with matched accuracies. The average error consistency is then
\begin{equation}
    E(m) : \mathbb{R} \to [-1,1], \quad
    m \mapsto
    \frac{1}{|D|}
    \sum_{d \in D}
    \frac{1}{|H_d|}
    \sum_{h \in H_d}
    \frac{1}{|C_d|}
    \sum_{c \in C_d}
    \frac{
        \dfrac{1}{|S_{d,c}|}\displaystyle\sum_{s \in S_{d,c}} b_{h,m}(s)
        - \hat{o}_{h,m}(S_{d,c})
    }{
        1 - \hat{o}_{h,m}(S_{d,c})
    }
\end{equation}
where higher values indicate that the model's pattern of errors is more similar to that of human observers than would be expected by chance.

\newpage

\newpage
\section{Shape---texture cue conflict benchmark}\label{supp:bench_shape_texture}

\begin{figure}[ht]
    \centering
    \includegraphics[width=1\textwidth]{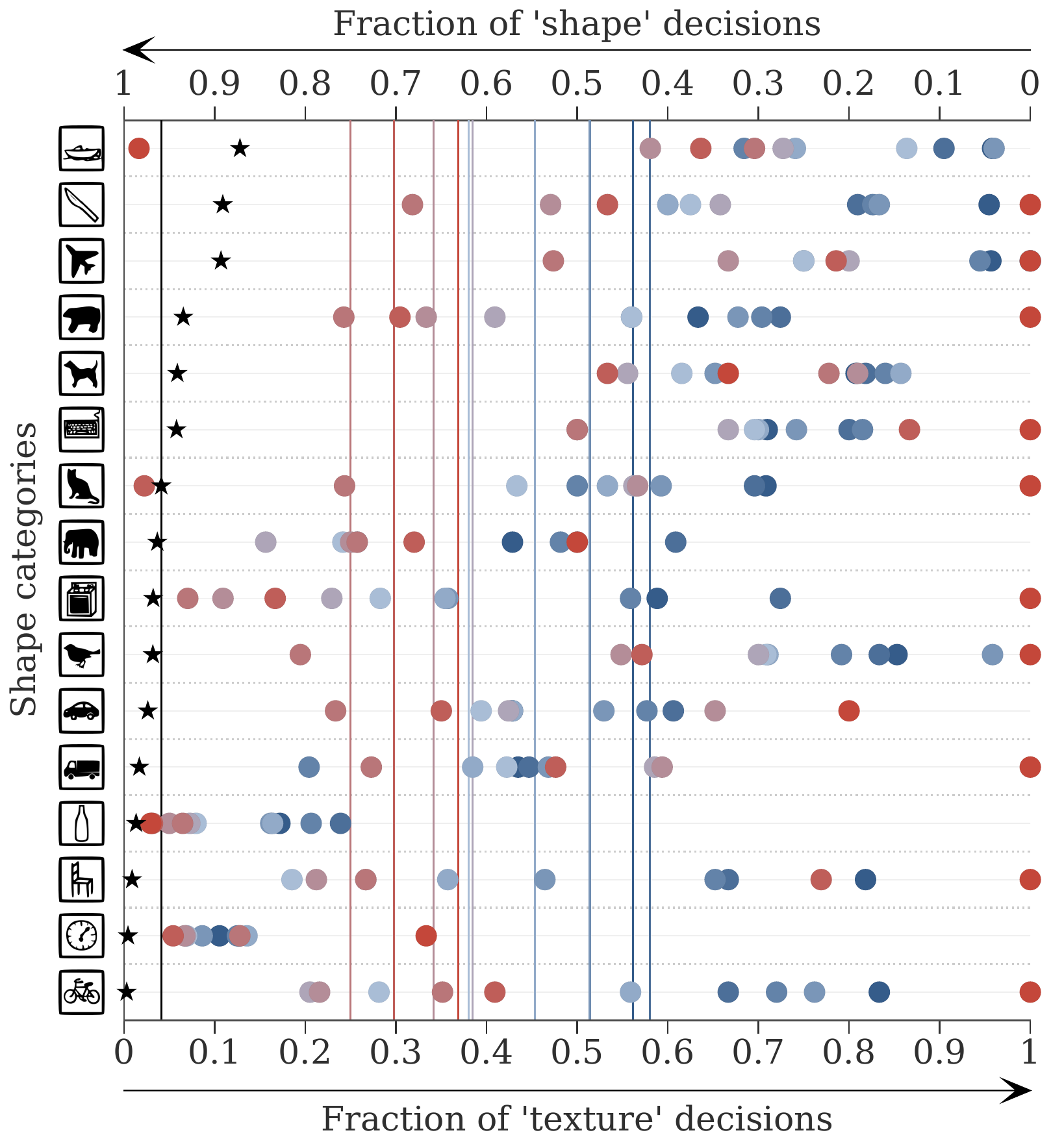}
    \caption{Percentage of shape or texture choice made per category for each model .}
    \label{fig:category_shape_texture}
\end{figure}

 \citet{geirhos2018imagenet} created a cue-conflicting images dataset as shown in  Fig.\ref{fig:shape-bias}, where the outline is from one Imagenet category, and the texture is from another one. This allowed us to quantify across humans and models, which was the most important factor in making a classification decision.
Traditional CNNs have been shown to be biased toward texture, whereas generative models are much more shape-leaning. Humans evaluated on this task are heavily biased towards shape/contours.
We can also clearly see from Fig.~\ref{fig:category_shape_texture} that this task is object-dependent and varies significantly across objects. Another thing to note is that the purely generative JEM seems all over the place; this is because the classification head was never trained and is randomly initialized.

%Geirhos et al.~\cite{geirhos2018imagenet} introduced the shape--texture cue-conflict benchmark, in which each image combines the shape of one ImageNet class with the texture of another. This makes it possible to quantify whether humans and models rely more strongly on shape or texture when making object-recognition decisions. Geirhos et al. showed that standard ImageNet-trained CNNs are strongly biased toward texture, whereas human observers are strongly biased toward shape~\cite{geirhos2018imagenet}.

%In our results (Fig.~\ref{fig:main_alpha_curves}e and Fig.~15), this benchmark reveals a clear progression across the JEM continuum. As the generative contribution increases, shape bias increases as well, moving the models away from the texture-dominated regime of standard CNNs and toward more human-like behavior. We also observe substantial variation across object categories, consistent with prior work showing that cue-conflict responses depend strongly on the object involved~\cite{geirhos2018imagenet}. Finally, the purely generative JEM appears unstable on this benchmark, which is expected in our setting because its classification head is not trained directly and therefore does not provide a meaningful discriminative readout.

\begin{figure}[h]
    \centering
    \includegraphics[width=1\textwidth]{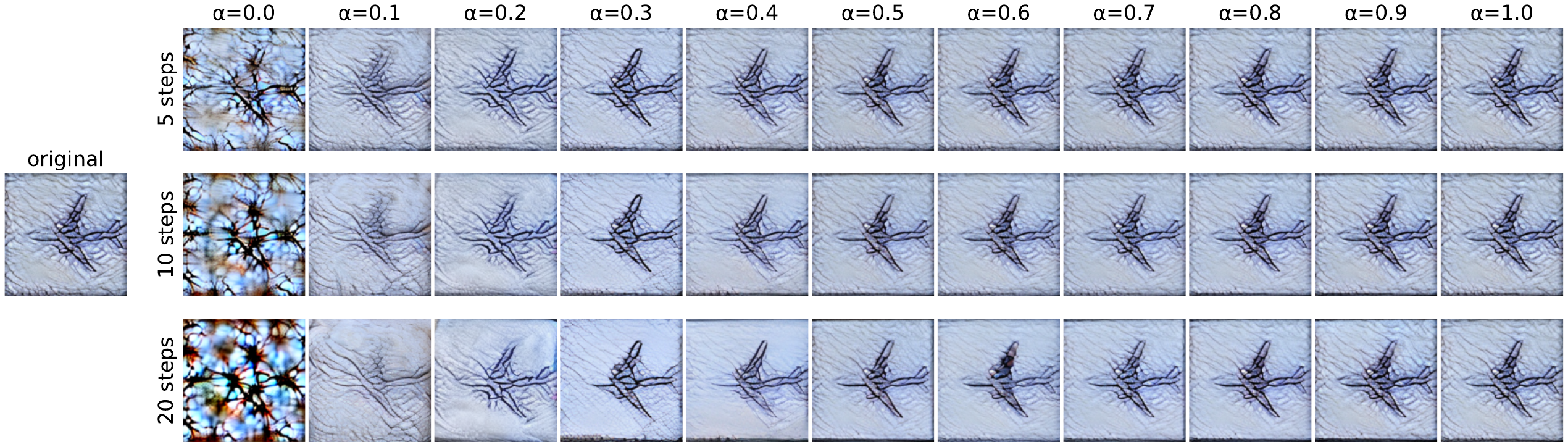}
    \caption{Evolution of an image dependent on the alpha and the number of MCMC steps.}
    \label{fig:evolution_cue_opposition}
\end{figure}

One advantage of JEMs is that they permit refinement of the input through MCMC during testing, pushing the image toward regions of higher probability. This can be interpreted as allowing the model to ``think longer'' and move away from uncertain or ambiguous states. As a result, models with small \(\alpha\) refine images toward textures whereas in models with larger \(\alpha\), shape bias becomes even stronger after a few MCMC steps. To illustrate this effect, we visualize the evolution of a representative cue-conflict image at 0, 5, 10, and 20 MCMC steps (Fig. \ref{fig:evolution_cue_opposition}). 

\begin{figure}[h]
    \centering
    \includegraphics[width=.6\textwidth]{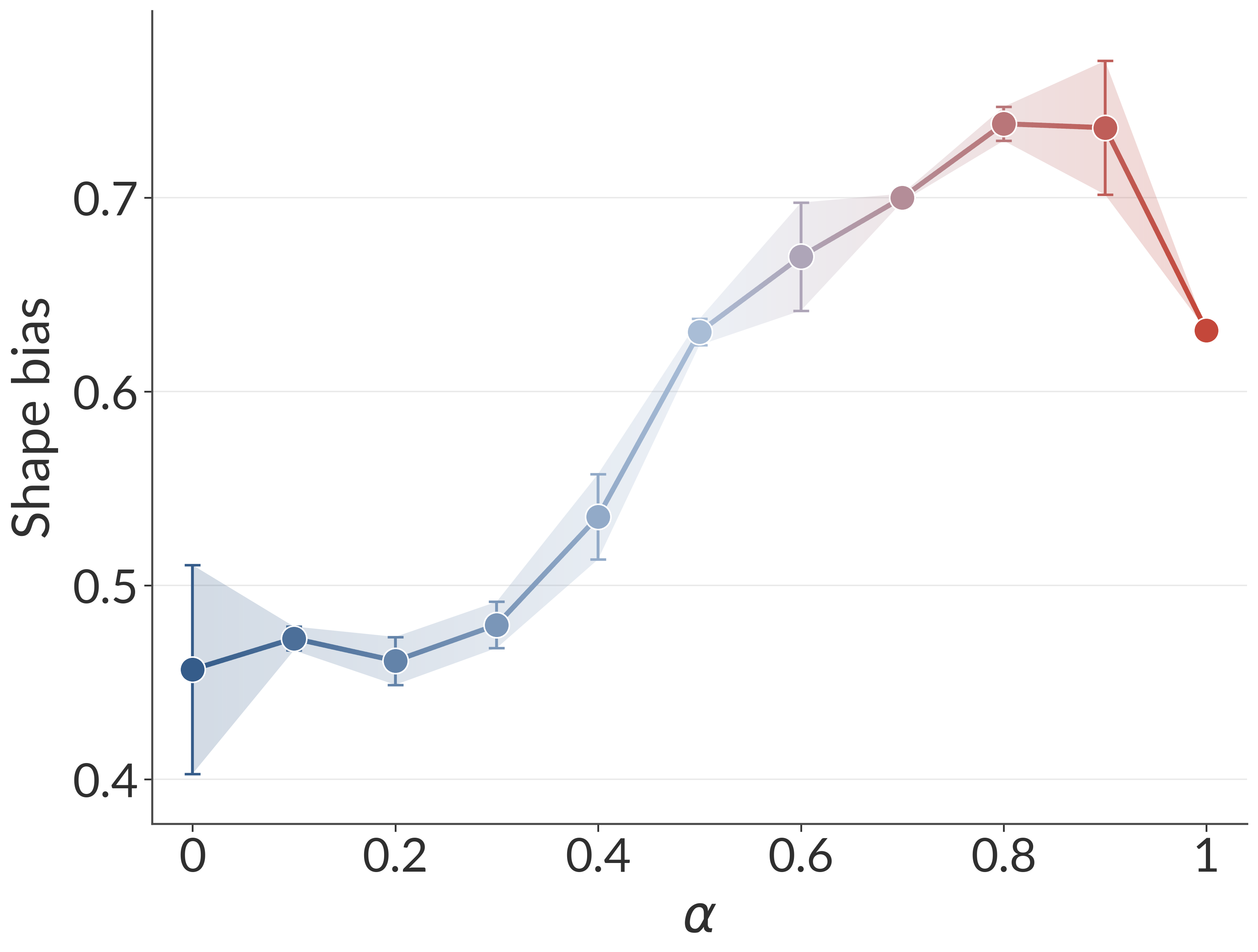}
    \caption{Shape bias across JEM $\alpha$ values. Shaded regions indicate the standard error of the mean (SEM) across two seeds.}
    \label{fig:shape_seeds}
\end{figure}
\newpage
\section{The Click-Me benchmark}\label{supp:click_me}

\begin{figure}[h]
    \centering
    \includegraphics[width=.9\textwidth, trim={0 510pt 0 510pt}, clip]{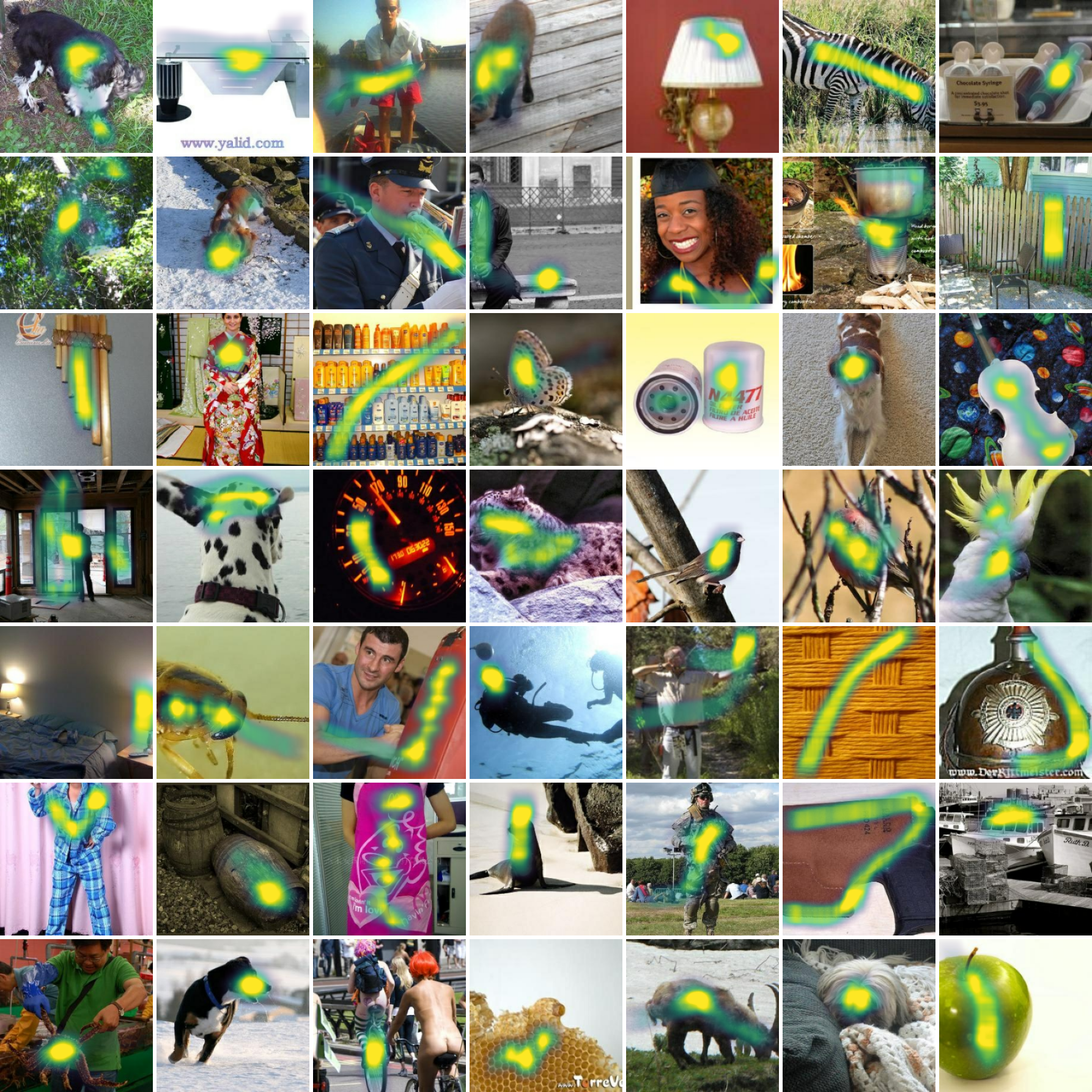}
    \caption{Examples of human feature importance maps.}
    \label{fig:Clickme_human}
\end{figure}

Modern CNNs achieve high performance on object-recognition benchmarks, but they are also known to rely on shortcut cues that can diverge from the diagnostic features used by human observers. To assess this aspect of alignment, we use the ClickMe dataset introduced by~\citet{linsley2018learning}, which provides large-scale human feature-importance maps for ImageNet images (Fig. \ref{fig:Clickme_human}). In this task, participants highlight the image regions they consider most informative for determining the object category, yielding a direct measure of the visual evidence humans rely on for recognition.

\newpage

\begin{figure}[h]
    \centering
    \includegraphics[trim={0 25pt 0 15pt}, clip, width=.6\textwidth]{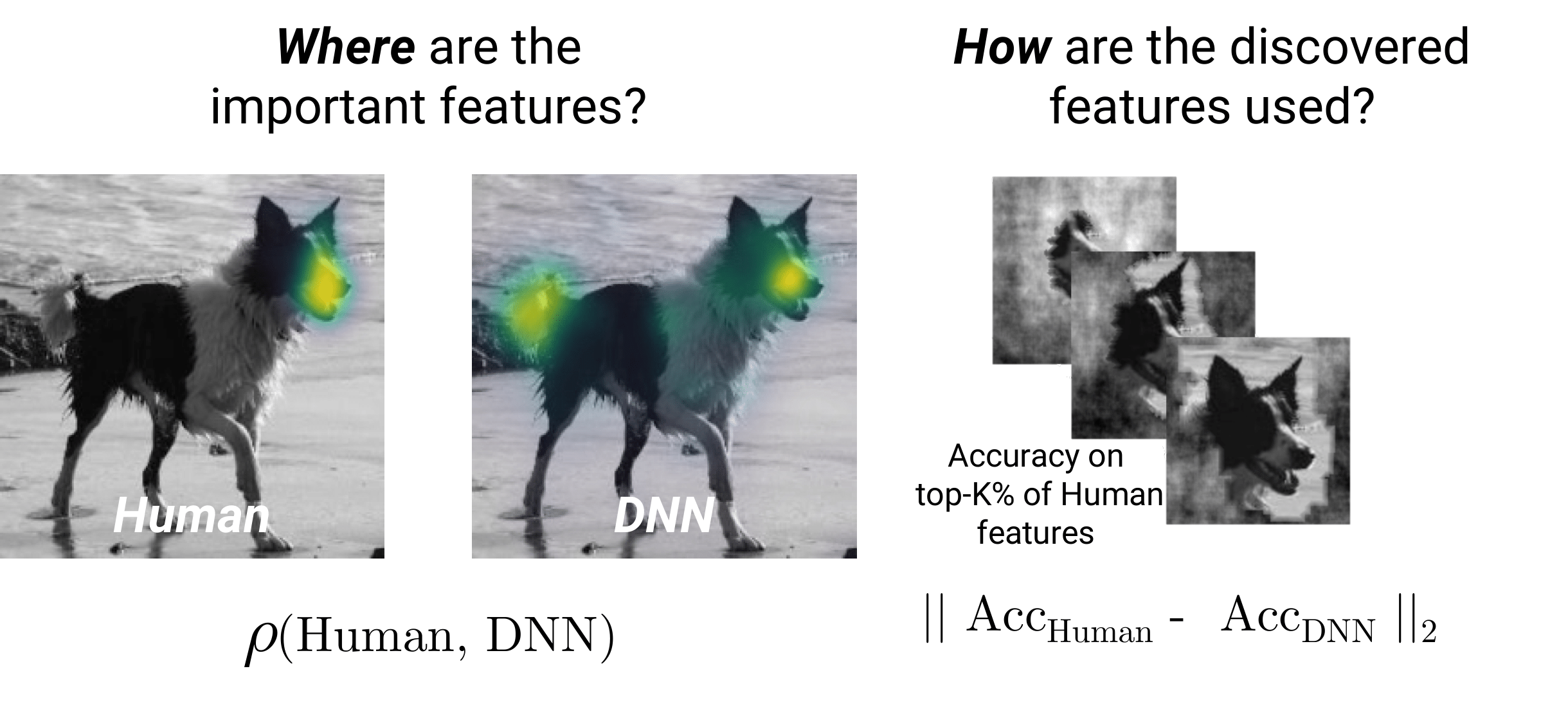}
    \caption{Visual strategy of object recognition.}
    \label{fig:Clickme_human_vs_model_attention}
\end{figure}

\paragraph{Evaluation metrics}.
To compare models with humans, we follow the evaluation protocol of~\citet{fel2022b}. For each model, saliency maps are computed on the ClickMe images and compared with the corresponding human feature-importance maps, yielding a quantitative measure of feature alignment (Fig. \ref{fig:Clickme_human_vs_model_attention}). In our case, however, the classifier is built on top of a pretrained VAE, and we do not want gradients through the VAE to dominate the attribution signal. We therefore adopt the approach of~\citet{boutin2024latent}, which uses the VAE decoder's Jacobian to propagate saliency from the latent space back to pixel space.

We call  $p_\psi(x|z)$ the decoder of the VAE, and $p_\theta(e|z)$ the energy density learnt by the JEM. To make the mathematical derivations more concise, we define the following functions:

\begin{equation}
    p_\psi : \mathbb{R}^d \longrightarrow \mathbb{R}^D \quad \text{and} \quad p_\theta : \mathbb{R}^d \longrightarrow \mathbb{R}
\end{equation}

\begin{equation}
    z \longmapsto x = \log p_\psi (\cdot|z) \quad z \longmapsto e = \log p_\theta (\cdot|z) 
\end{equation}

To project each energy value $e$ into the pixel space, we feed them into the decoder. The resulting projection is $x = p_{\psi,\theta} (z) = p_\psi \circ p_\theta (z)$.

For each energy value, the importance feature map quantifies how the absolute value of $p_{\psi,\theta}(e)$ changes as $z$ varies. $\phi(x)$ describes the accumulation, of these ``local feature maps'':

\begin{align}
    \phi(x) &= \frac{\partial p_{\psi,\theta} (e)}{\partial x}  \\
    &=  \frac{\partial p_\psi \circ p_\theta (e)}{\partial x} \\
    &= \frac{\partial p_\psi}{\partial x} (p_\theta (z)) \frac{\partial p_\theta}{\partial z} (e)  \\
    &= J_{p_\psi} (x) \nabla_{z} p_\theta (z) 
\end{align}

with $J_{p_\psi} (x)$ being the Jacobian of the function $p_\psi$ w.r.t. $x$ computed in $p_\theta (z)$. 

%It was collected in the form of a 2 participant game where one participant gradually revealed parts of the image and had to get the second participant to guess what the category is. The second participant's role is not majorly important and has in later version been replaced by a CNN.
%CNNs have had great success in achieving high accuracy on benchmarks, but they have been shown to base themselves a lot on shortcuts.The Clickme dataset colected by \cite{linsley2018learning} comprises nearly 200,000 human feature importances maps. In these participants revealed what part of the image plays a role to select the category and what visual strategy is used.
%To evaluate models against human we use the method form Fel et al.\cite{fel2022b}. To evaluate what pixel the model pays most attention to, they used feature visualisation methods such as saliency on all images from the clickme dataset. The saliency maps can then be correlated with the human feature attention maps, and allows to get a quantitative measure of feature alignment use.
%In our case however, the model has been built on top of a pretrained VAE, and we do not want the gradient of the VAE to pollute the information from the JEM. For this, we use a method from Boutin et al. \cite{boutin2024latent} using the jacobian of the VAE decoder to propagate the saliency information to the pixel space.

\begin{figure}[h]
    \centering
    \includegraphics[trim={0 0 0 40pt}, clip, width=.47\textwidth]{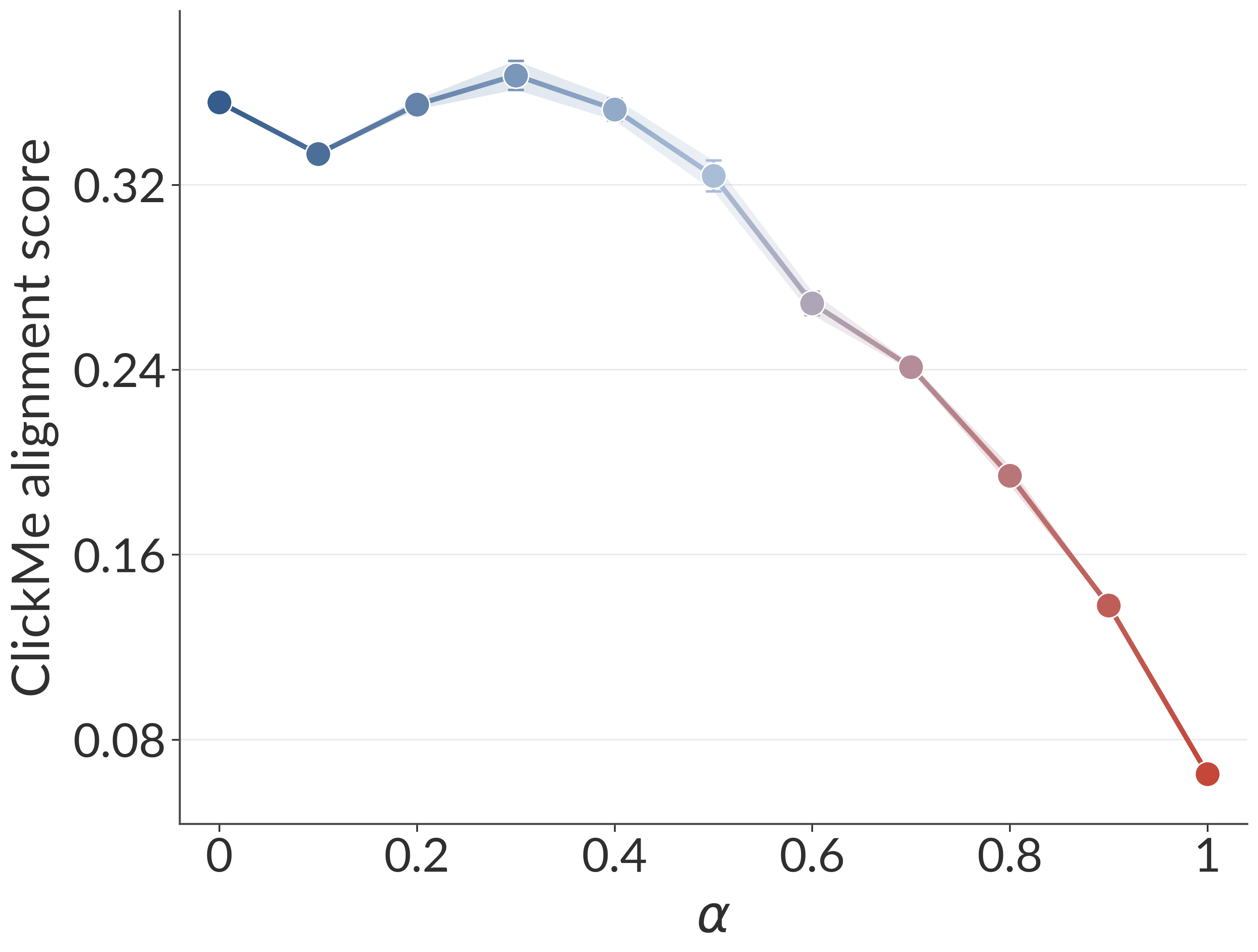}
    \caption{ClickMe alignment score across JEM $\alpha$ values. Shaded regions indicate the standard error of the mean (SEM) across two seeds.}
    \label{fig:clickme_seeds}
\end{figure}

\newpage
\end{document}